\pdfoutput=1
\documentclass{article}

    \usepackage[final]{neurips_2021}

\usepackage[utf8]{inputenc} %
\usepackage[T1]{fontenc}    %
\usepackage{hyperref}       %
\usepackage{url}            %
\usepackage{booktabs}       %
\usepackage{amsfonts}       %
\usepackage{nicefrac}       %
\usepackage{microtype}      %
\usepackage{xcolor}         %

\title{
    Beware of the Simulated DAG! \\ 
    Causal Discovery Benchmarks May Be Easy To Game
}
\author{%
Alexander G. Reisach\textsuperscript{1,2}
\hfill
Christof Seiler\textsuperscript{2,3}
\hfill
Sebastian Weichwald\textsuperscript{1}
\\\\  %
\textsuperscript{1}Department of Mathematical Sciences,
University of Copenhagen,
Denmark\hfill\-\\
\textsuperscript{2}Department of Data Science and Knowledge Engineering,
Maastricht University,
The Netherlands\hfill\-\\
\textsuperscript{3}Mathematics Centre Maastricht,
Maastricht University,
The Netherlands\hfill\-
}

\usepackage[toc,page,header]{appendix}
\usepackage{minitoc}
\usepackage{float}
\usepackage{graphicx}
\usepackage{grffile}
\usepackage{subcaption}
\usepackage{tabularx}
\usepackage{csquotes}
\usepackage{longtable}
\usepackage{todonotes}
\usepackage{xfrac}
\usepackage{mathtools}
\mathtoolsset{centercolon}

\usepackage{amsmath,amsthm,amsfonts,amssymb}
\usepackage[ruled,vlined]{algorithm2e}
\usepackage[capitalise,noabbrev]{cleveref}
\usepackage{centernot}

\renewcommand{\phi}{\varphi}

\renewcommand{\hat}{\widehat}

\newcommand{\MSE}[2]{\operatorname{MSE}_{#2}\left(#1\right)}
\newcommand{\PA}[1]{\operatorname{PA}\left(#1\right)}
\newcommand{\tr}{\operatorname{tr}}
\newcommand{\X}{\mathbf{X}}
\newcommand{\R}{\mathbf{R}}
\newcommand{\D}{\mathbf{D}}
\newcommand{\diag}[1]{\operatorname{diag}\left(#1\right)}
\DeclareMathOperator*{\argmin}{\arg\!\min}

\newcommand{\includegraphicsmaybe}[2][]{\IfFileExists{#2}{\includegraphics[#1]{#2}}{\includegraphics[#1]{figures/default/spaceTerminator.png}}}

\newcommand{\var}[1]{\ensuremath\operatorname{Var}(#1)}
\newcommand{\cor}{\ensuremath\operatorname{Corr}}
\newcommand{\covar}[2]{\ensuremath\operatorname{Cov}(#1,#2)}

\usepackage{tikz}

\usepackage{listings}

\DeclareSymbolFont{matha}{OML}{txmi}{m}{it}
\DeclareMathSymbol{\varv}{\mathord}{matha}{118}

\usepackage{enumitem}

\usepackage{wrapfig}

\definecolor{linkcolor}{HTML}{338040}
\definecolor{citecolor}{HTML}{660022}
\definecolor{urlcolor}{HTML}{006666}

\hypersetup{
    colorlinks,
    linkcolor={linkcolor},
    citecolor={citecolor},
    urlcolor={urlcolor}
}

\begin{document}

\maketitle

\renewcommand \thepart{}
\renewcommand \partname{}
\doparttoc %
\faketableofcontents %

\begin{abstract}
Simulated DAG models may exhibit properties that, perhaps inadvertently, render their structure identifiable and unexpectedly affect structure learning algorithms.
Here, we show that marginal variance tends to increase along the causal order for generically sampled additive noise models.
We introduce \emph{varsortability} as a measure of the agreement between the order of increasing marginal variance and the causal order.
For commonly sampled graphs and model parameters, we show that the remarkable performance of some continuous structure learning algorithms can be explained by high varsortability and matched by a simple baseline method.
Yet, this performance may not transfer to real-world data where varsortability may be moderate or dependent on the choice of measurement scales.
On standardized data, the same algorithms fail to identify the ground-truth DAG or its Markov equivalence class.
While standardization removes the pattern in marginal variance, we show that data generating processes that incur high varsortability also leave a distinct covariance pattern that may be exploited even after standardization.
Our findings challenge the significance of generic benchmarks with independently drawn parameters. The code is available at \url{https://github.com/Scriddie/Varsortability}.
\end{abstract}

\section{Introduction}\label{sec:introduction}

\paragraph{Causal structure learning} aims to infer a causal model from data.
Academic disciplines anywhere from biology, medicine, finance,
to machine learning are interested in causal models
\citep{rothman2008modern,imbens2015causal,sanford2012bayesian,scholkopf2019causality}.
Causal models not only describe the observational joint distribution of variables but also formalize predictions under interventions and
counterfactuals \citep{spirtes2000causation,pearl2009causal,peters2017elements}.
Directed acyclic graphs (DAGs) are common to represent causal structure:
nodes represent variables and directed edges point from cause to effect representing the causal relationships.
This graphical representation rests on assumptions which have been critically questioned, for example by \cite{dawid2010beware}.
Inferring causal structure from observational data is difficult: 
Often we can only identify the DAG up to its Markov equivalence class (MEC) and finding high-scoring DAGs is NP-hard~\citep{chickering1996learning,chickering2004large}.
Here, we focus on learning the DAG of linear additive noise models (ANM). 

\paragraph{Data scale and marginal variance} may carry information about the data generating process.
This information can dominate
benchmarking results,
such as, for example, the outcome of the NeurIPS Causality 4 Climate competition~\citep{runge2020causality}.
Here, the magnitude of regression coefficients was informative
about the existence of causal links
such that ordinary regression-based methods on raw data
outperformed causal discovery algorithms~\citep{weichwald2020causal}.
Multiple prior works 
state the importance of data scale for structure learning
either
implicitly or explicitly.
Structure identification by \textit{ICA-LiNGAM}~\citep{shimizu2006linear},
for example, is susceptible to
rescaling of the variables.
This motivated the development of \textit{DirectLiNGAM}~\citep{shimizu2011directlingam}, a scale-invariant causal discovery algorithm for linear non-Gaussian models.
The causal structure of ANMs
is proven to be identifiable given the noise scale
(cf.~\cref{sec:identifiability}).
Yet, such identifiability results require knowledge about the ground-truth data scale.

\paragraph{Simulated DAGs} may be identifiable from marginal variances under generic parameter distributions.
An instructive example is the causal graph $A \to B$ with structural equations $A = N_A$ and $B = wA + N_B$
with $w\neq 0$ and independent zero-centered noise variables $N_A, N_B$.
The mean squared error (MSE) of a model $X\to Y$ is given by $\MSE{X\to Y}{} = \var{X} + \var{Y|X}$.
It holds that
$    \MSE{A \to B}{} < \MSE{B \to A}{}
    \iff \var{A} < \var{B}
    \iff (1 - w^2) \var{N_A} < \var{N_B}$
(see \cref{app:two_node}).
Deciding the directionality of the edge between $A$ and $B$ based on the MSE amounts to inferring an edge from the lower-variance variable to the higher-variance variable.
For error variances $\var{N_A} \leq \var{N_B}$ 
and any non-zero edge weight $w$, the MSE-based inference is correct.
This resembles known scale-based identifiability results
based on equal or monotonically increasing error variances~\citep{peters2014identifiability,park2020identifiability}.
However, if the observations of $A$ were multiplied by a sufficiently large constant, the MSE-based inference would wrongfully conclude that $A \gets B$.
This is problematic since simply choosing our units of measurement differently may change the scale and variance of $A$.
Arguably, this is often the case for observations from real-world systems:
There is no canonical choice as to whether
we should pick meters or yards for distances, 
gram or kilogram for weights,
or
yuan or dollar as currency.
A researcher cannot rely on obtaining the same results for different measurement scales or after re-scaling the data when applying any method that leverages the data scale (examples include \cite{peters2014identifiability,park2020identifiability}, or \cite{zheng2018dags}, who employ the least squares loss studied by \cite{loh2014high}).
\paragraph{Continuous causal structure learning algorithms}  
optimize model fit under a differentiable acyclicity constraint~\citep{zheng2018dags}. 
This allows for the use of continuous optimization and avoids 
the explicit combinatorial traversal of possible causal structures.
This idea has found numerous applications and extensions
\citep{
    lachapelle2019gradient,
    lee2019scaling,
    ng2020role,
    yu2019dag,
    brouillard2020differentiable,
    pamfil2020dynotears,
    wei2020dags,
    zheng2020learning,
    bhattacharya2021differentiable};
\cite{vowels2021dya} provide a review.
\textit{NOTEARS}~\citep{zheng2018dags} uses the MSE with reference to \cite{loh2014high}, while \textit{GOLEM}~\citep{ng2020role} assesses model fit by the penalized likelihood assuming a jointly Gaussian model.
On simulated data and across noise distributions,
both methods recover graphs that are remarkably close to the ground-truth causal graph in structural intervention distance (SID) and structural hamming distance (SHD).
We agree with the original authors that these empirical findings,
especially under model misspecification and given the non-convex loss landscape, may seem surprising at first.
Here, we investigate the performance under data standardization
and explain how the causal order is (partially) identifiable from the raw data scale
alone in common generically simulated benchmarking data.

\paragraph{Contribution.} \label{subsec:contribution}

We show that causal structure drives the marginal variances of nodes in an ANM and can lead to (partial) identifiability. 
The pattern in marginal variances is dominant in ANM benchmark simulations with edge coefficients drawn identically and independently.
We introduce varsortability as a measure of the information the data scale carries about the causal structure.
We argue that high varsortability affects
the optimization procedures
of continuous structure learning algorithms.
Our experiments demonstrate that varsortability dominates the optimization and helps achieve state-of-the-art performance
provided the ground-truth data scale.
Data standardization or an unknown data scale remove this information and the same algorithms
fail to
recover the ground-truth DAG. %
Even methods using a score-equivalent likelihood criterion (\textit{GOLEM})
recover neither ground-truth DAG nor its MEC on standardized data.
To illustrate that recent benchmark results depend heavily on high varsortability,
we provide a simple baseline method that exploits increasing marginal variances
to achieve state-of-the-art results on these benchmarks.
We thereby provide an explanation for the unexpected
performance of recent continuous structure learning algorithms in identifying the true DAG.
{
Neither algorithm dominates on raw or standardized observations of the analyzed real-world data.
We show how, even if data is standardized and even in non-linear ANMs, a causal discovery benchmark may be gamed
due to covariance patterns.
Consequently, recent benchmark results may not transfer to (real-world) settings 
where the correct data scale is unknown or where edge weights are not drawn independent and identically distributed (iid).
We conclude that structure learning benchmarks on ANMs with generically sampled parameters
may be distorted due to unexpected and perhaps unintended regularity patterns in the data.

\section{Background} \label{section:related_lit}

\subsection{Model Class}

We consider acyclic linear additive noise models.
Single observations are denoted by
$x^{(i)} \in \mathbb{R}^d$
where $x^{(i)}_j$ denotes 
the $j^\text{th}$ dimension
of
the $i^\text{th}$
iid\ observation
of random vector $X = [X_1,...,X_d]^\top$.
All observations are stacked as $\X = [x^{(1)},...,x^{(n)}]^\top \in \mathbb{R}^{n \times d}$
and $x_j \in \mathbb{R}^n$ refers to the $j^\text{th}$ column of $\X$.
Analogously,
$n^{(i)}$ denotes the corresponding $i^\text{th}$ iid\ observation
of the random noise variable $N = [N_1,...,N_d]^\top$
with independent zero-centred components.
The linear effect of variable $X_k$ on $X_j$ is
denoted by $w_{k\to j} = w_{kj}$.
The causal structure corresponding to the adjacency matrix $W = [w_{kj}]_{k,j=1,...,d}$ 
with columns $w_j = [w_{k\to j}]_{k=1,...,d}\in\mathbb{R}^d$
can be represented by
a directed acyclic graph 
$G = (V_G, E_G)$ with vertices $V_G=\{1,...,d\}$
and edges $E_G = \{(k,j) : w_{k\to j} \neq 0\}$.
Edges can be represented by an adjacency matrix $E$
such that the $(k,j)^\text{th}$ entry of $E^l$ is non-zero if and only if a directed path of length $l$
from $k$ to $j$ exists in $G$.
For a given graph, the parents of $j$ are denoted by $\PA{j}$.
The structural causal model is
$X = W^\top X + N$.

\subsection{Identifiability of Additive Noise Models}\label{sec:identifiability}
Identifiability of the causal structure or its MEC
requires causal assumptions. %
Under causal faithfulness and Markov assumptions,
the causal graph can be recovered up to its MEC~\citep{chickering1995bayesian,spirtes2000causation}.
Faithfulness, however, is untestable~\citep{zhang2008detection}.
\cite{shimizu2006linear} show that under the assumptions of no unobserved confounders, faithfulness, linearity,
and non-Gaussian additive noise, the causal graph can be recovered from data.
\cite{hoyer2009nonlinear} show that this holds for any noise distribution under the assumption of strict non-linearity.
This finding is generalized to post-nonlinear functions by \cite{zhang2012identifiability}.
\cite{peters2014identifiability} prove that the causal structure of a linear causal model with
Gaussian noise is identifiable if the error variances are equal or known.
Any unknown re-scaling of the data breaks this condition.
For the case of linear structural causal models,
\cite{loh2014high} provide a framework for DAG estimation
based on a noise variance-weighted least squares score function.
For ANMs, they give conditions under which the general Gaussian case can be identified via approximating it by the equal noise-variance case given knowledge of the (approximate) noise scale.
Finally, subsuming further prior results on (linear)
ANMs~\citep{hoyer2009nonlinear,ghoshal2017learning,ghoshal2018learning,chen2019on},
\cite{park2020identifiability} shows that the causal structure is identifiable under regularity conditions on the conditional variances along the causal order.
In particular, identifiability holds if the error variances of nodes
are weakly monotonically increasing along the causal order.

\subsection{Structure Learning Algorithms} \label{subsec:algorithms}

\paragraph{Combinatorial structure learning algorithms (such as \textit{PC, FGES, DirectLiNGAM})}
separately
solve the combinatorial problem of searching over structures and finding the optimal parameters for each structure.
To remain computationally feasible, the search space of potential structures is often
restricted or traversed according to a heuristic.
One can, for example, carefully choose which conditional independence statements to evaluate
in constraint-based algorithms,
or employ greedy (equivalence class) search in score-based algorithms.
In our experiments,
we consider \textit{PC}~\citep{spirtes1991algorithm},
\textit{FGES}~\citep{meek1997graphical, chickering2002optimal}, \textit{DirectLiNGAM}~\citep{shimizu2011directlingam},
{and a greedy DAG search (GDS) algorithm \textit{MSE-GDS}
that greedily includes those edges that reduce the MSE the most.}
For details see \cref{app:algorithms}.
\paragraph{Continuous structure learning algorithms (such as \textit{NOTEARS} and \textit{GOLEM})} employ continuous optimization to simultaneously optimize over structures and parameters.
As a first step towards expressing causal structure learning as a continuous optimization problem,
\cite{aragam2015concave} propose $l^1$-regularization instead of the conventional $l^0$-penalty for model selection.
\citet{zheng2018dags} propose a differentiable acyclicity constraint,
allowing for end-to-end optimization of score functions over graph adjacency matrices. 
We examine and compare the
continuous structure learning algorithms \textit{NOTEARS}~\citep{zheng2018dags} and \textit{GOLEM}~\citep{ng2020role}.
For details see \cref{app:algorithms}.
\textit{NOTEARS} \citep{zheng2018dags} minimizes the MSE between observations 
and model predictions subject to a hard acyclicity constraint.
The MSE with respect to $W$ on observations $\X$ is defined as
    \(
    \MSE{W}{\X} = \frac{1}{n}\|\X - \X W\|_2^2
    \)
where 
\(
\|\cdot \|_2 = \|\cdot\|_F
\) denotes the Frobenius norm.

\textit{GOLEM}~\citep{ng2020role} performs
maximum likelihood estimation (MLE) under the assumption of a Gaussian distribution with equal (EV) or non-equal (NV) noise variances. 
There are soft acyclicity and sparsity constraints.
The unnormalized negative likelihood-parts of the objective function are
\(
    \mathcal{L}_{EV}(W, \X) = \log(\MSE{W}{\X})%
\)
and
\(
    \mathcal{L}_{NV}(W, \X) = \sum_{j=1}^{d}\log\big(\frac{1}{n}\|x_j-\X w_j\|_2^2\big)%
\),
respectively,
omitting a $-\log(|\det(I-W)|)$ term that vanishes when $W$ represents a DAG \citep{ng2020role}.

To ease notation,
we sometimes drop the explicit reference to $\X$ when referring to
$\operatorname{MSE},\mathcal{L}_{EV},\mathcal{L}_{NV}$.

\section{Varsortability} \label{section:methodology}
The data generating process 
may leave information about the causal order in the data scale. 
We introduce varsortability as a measure of such information.
When varsortability is maximal, the causal order is identifiable.
Varsortability is high in common simulation schemes used for benchmarking causal structure learning algorithms.
We describe how continuous structure learning algorithms are affected by marginal variances and how they may leverage high varsortability.
This elucidates the results of continuous methods
reported by \cite{zheng2018dags}, \cite{ng2020role}, and others on raw data
and predicts impaired performance on standardized data
as confirmed in \cref{sec:simulations}.
We introduce \textit{sortnregress}
as simple baseline method that sorts variables by marginal variance followed by parent selection.
The performance of \textit{sortnregress}
reflects the degree of varsortability
in a given setting
and establishes a reference baseline
to benchmark structure learning algorithms against.

\subsection{Definition of Varsortability}\label{sec:varsortability}

We propose varsortability
as a measure of agreement between
the order of increasing marginal variance and the causal order.
For any causal model
over variables $\{X_1, ..., X_d\}$
with (non-degenerate) DAG adjacency matrix $E$
we define varsortability as
the fraction of directed paths that start from a node with strictly lower variance than the node they end in, that is,

\(
\varv := \frac{
\sum_{k=1}^{d-1} \sum_{i\to j\in E^k}
\operatorname{increasing}\left(\var{X_i},\var{X_j}\right)
}{
\sum_{k=1}^{d-1} \sum_{i\to j\in E^k} 1
} \in [0, 1]
\)
where
\(
\operatorname{increasing}(a,b) = 
\begin{cases}
1 & a < b \\
\sfrac{1}{2} & a = b \\
0 & a > b 
\end{cases}
\)

\begingroup

\definecolor{colA}{HTML}{386641}
\definecolor{colB}{HTML}{6a994e}
\definecolor{colC}{HTML}{bc4749}
\definecolor{colD}{HTML}{a7c957}
For example, we calculate the varsortability as
$\varv = \frac{
{\color{colA}1} +
{\color{colB}1} +
{\color{colD}1} \phantom{+ 1}}{
{\color{colA}1} +
{\color{colB}1} +
{\color{colD}1} +
{\color{colC}1}
} = \frac{3}{4}$
given the causal graph below.

\endgroup

\begin{wrapfigure}{r}{0.4\textwidth}
\begin{center}
\begin{tikzpicture}
\definecolor{colA}{HTML}{386641}
\definecolor{colB}{HTML}{6a994e}
\definecolor{colC}{HTML}{bc4749}
\definecolor{colD}{HTML}{a7c957}
\node (vara) at(0,0.5) {$\var{A}=2$}; 
\node (a) at(0,0) {$A$}; 
\node (b) at(1.75,-1.25) {$B$}; 
\node (varb) at(1.75,-1.75) {$\var{B}=1$}; 
\node (varc) at(3,1.0) {$\var{C}=3$}; 
\node (c) at(3,0.5) {$C$}; 
\draw[->,thick,colC] (a) -- (b);
\draw[->,thick,colA] (a) -- (c);
\draw[->,thick,colB] (b) -- (c);

\draw[thick,colD] (.4,-.1) -- (1.75,-1.05);
\draw[->,thick,colD] (1.75,-1.05) -- (2.75,0.35);
\end{tikzpicture}
\end{center}
\end{wrapfigure}

Varsortability equals one if the marginal variance of each node is strictly greater than that of its causal ancestors.
Varsortability equals zero if the marginal variance of each node is strictly greater than that of its 
descendants.
Varsortability does not depend on choosing one of the possibly multiple causal orders and captures the overall agreement
between the partial order induced by the marginal variances
and all pathwise descendant relations implied by the causal structure.
In the two-node introductory example (cf.~\cref{sec:introduction}), varsortability $\varv = 1$ is equivalent to
\begin{align*}
    \var{A} < \var{B} \iff \var{N_A} < w^2 \var{N_A} + \var{N_B} %
\end{align*}
where $A$ and $B$ are nodes in the causal graph $A \overset{w}{\to} B$
with noise variances
$\var{N_A}$ and $\var{N_B}$.

We can also understand varsortability as a property of the distribution 
of graphs and parameters that we sample from
for benchmarks on synthetic data.
The distribution of weights and noise variances determines whether the causal order of any two connected nodes in the graph agrees with the order of increasing marginal variance and
in turn determines the varsortability of the simulated causal models.
We observe that even for modest probabilities of any two neighboring nodes being correctly ordered by their marginal variances, the variance of connected nodes
tends to increase quickly
along the causal order
for many ANM instantiations
(cf.\ \cref{app:causal_order_vsb}).
For a heuristic explanation, recall that
we obtain the marginal variance of a node by adding the variance contribution of all its ancestors to the node's own noise variance; to obtain the variance contribution of an ancestor, we take the product of the edge weights along each directed path from ancestor to node, sum these path coefficient products, square, and multiply with the ancestor's noise variance.
While the sum of path coefficient products may vanish or be small
such that the variance contribution of an ancestor cancels out or is damped across the different connecting paths,
we find it is unlikely if edge weights are drawn independently
(cf.\ \cite{meek1995strong} for why exact cancellation and faithfulness violations are unlikely). 
Furthermore, the further apart a connected pair of nodes,
the more variance may be accumulated in the descendant node along all incoming paths
in addition to one ancestor's (possibly damped) variance contribution
further fueling the tendency for descendant nodes to have higher variance than their ancestors.
In practice we indeed find that
an ordering of nodes by increasing marginal variance
closely aligns with the causal order
for commonly simulated linear and non-linear ANMs
(cf.~\cref{appendix:varsortability}).

\subsection{Varsortability and Identifiability}

If varsortability $\varv = 1$,
the causal structure is identifiable.
It can be recovered by ordering the nodes by increasing marginal variance and regressing each node onto its predecessors using conventional sparse regression approaches. 
The causal structure learning problem is commonly decoupled into causal order estimation and parent selection~\citep{shimizu2006linear,shojaie2010penalized,buhlmann2014cam,chen2019on,park2020identifiability}.
This decoupling is further warranted,
since we only need the causal ordering to consistently estimate interventional distributions \cite[Section 2.6]{buhlmann2014cam}. 
At $\varv=1$, an ordering by marginal variance is a valid causal ordering.
Given a causal ordering, one can construct a fully connected DAG and use parent selection to prune edges and reconstruct the graph in the sample limit under mild assumptions. 
\citet[Section 2.5]{buhlmann2014cam} discuss parent selection and
\cite{shojaie2010penalized} establish the consistency of an adaptive lasso approach for edge selection given a valid causal order.
The identifiability conditions
by \cite{park2020identifiability} are closely related to varsortability, though not equivalent as we prove in~\cref{sec:varsort_park}.
Identifiability of the causal order is immediate if varsortability $\varv=1$,
though, this shares severe drawbacks with other identifiability conditions
that rely on data scale by \cite{peters2014identifiability,loh2014high}, and \cite{park2020identifiability}.
First, it is difficult to verify or assess the plausibility of assumptions about the correctness or suitability of the data scale for any given dataset.
Second, any unknown rescaling may break previously met identifiability conditions.
Third, even if variables are on the same measurement scale the units may not correctly capture the ground-truth causal scale.
For example, a dartist's distance from the dartboard may affect the
precision of their throw measured by the distance between hit and target.
Here, the effect variable's marginal variance may be smaller than that of the cause (even) if both distances are measured in centimetres.
Nonetheless, it may be possible to exploit varsortability
if one can establish that certain assumptions on the data scale be met.

\subsection{Varsortability in Benchmarking Scenarios} \label{sec:bench}

For real-world data we cannot readily assess nor presume varsortability as we do not know the parameters and data scale of the data generating process.
When benchmarking causal structure learning algorithms, however, we can evaluate varsortability for the simulated DAGs and parameter settings.
We may acquire an intuition about the probability of varsortable cause-effect pairs in our simulation settings by considering two neighboring nodes $A \overset{w}{\to} B$ in the sampled graph without common ancestors and no other directed path from $A$ to $B$.
Under these assumptions, 
$\var{B} = w^2\var{A} + \var{\sum_{C\in\PA{B}\setminus\{A\}} w_{C\to B}^2 C} + \var{N_B}$
such that
$|w| > 1$ implies that the variable pair is varsortable.
To simulate an ANM, we need to sample a DAG, decide on the noise distributions,
and then sample edge weights and noise variances.
Across simulation instances in a given benchmark set-up,
the edge weights $w_{k\to j}$ and noise variances $s_k^2$ are iid
instantiations of independent random variables $W$ and $S^2$.
The distributions of $W$ and $S^2$ induce a distribution
of the marginal variance $V_Y$ of node $Y$ in the resulting ANM.
The probability for the variable pair $A \to B$ to be varsortable in a simulated ANM is then bounded from below by
\(
\operatorname{P}[(1-W_{A\to B}^2)V_A < S_{N_B}^2]
\)
(cf.~\cref{app:derivation_bound}).
If $A$ is a root node, $V_A = S_{N_A}^2$.
In the experimental settings used by,
for example, \cite{zheng2018dags,zheng2020learning,lachapelle2019gradient,ng2020role}, 
edge weights are independently drawn from a uniform distribution and noise standard deviations or variances are either fixed or also drawn independently from a uniform distribution.
For our parameters $W\overset{\text{iid}}{\sim}\operatorname{Unif}((-2, -0.5)\cup(0.5, 2))$ and $S\overset{\text{iid}}{\sim}\operatorname{Unif}((0.5, 2))$,
which resemble common choices in the literature, any pair is varsortable with probability at least $\sfrac{2}{3}$ due to $\operatorname{P}[|W| > 1] = \sfrac{2}{3}$,
and with probability $p > 0.93$ provided $A$ is a root node.
Empirically, we find that varsortability averages above $0.94$ in our simulated graphs and above $0.71$ in commonly considered non-linear 
ANMs (cf.~\cref{appendix:varsortability}).
This result indicates that in benchmark simulations the marginal variance of any two nodes in a graph tends
to increase along the causal order
and that we may game these benchmarks and perform well by exploiting this pattern.

If $A$ and $B$ have a common ancestor or mediator $C$,
the effect of $C$ on $B$ may either compound or
partially cancel out the effect of $A$ on $B$.
In practice, the effect commonly increases the variance of the effect node $B$, which may be attributed to the independent sampling
of path coefficients which also renders faithfulness violations improbable~\citep{meek1995strong}.
We find varsortability to increase with graph density and the lower bound presented above to be loose.
Motivated by the strong impact of different levels of varsortability on some structure learning algorithms as reported in \cref{sec:simulations,app:varsortability_sortnregress}, we advocate an empirical evaluation and reporting of varsortability (cf.~\cref{algo:varsortability} for the implementation) when simulating ANMs.
We emphasize that even for varsortability $<1$,
where the order of increasing variance does not perfectly agree with the causal order,
experimental results may still be largely driven by the overall
agreement between increasing marginal variance and causal order.
The extent to which varsortability may distort experimental
comparisons of structure learning algorithms on linear ANMs is demonstrated in~\cref{sec:simulations}.
\subsection{Marginal Variance yields Asymmetric Gradients for Causal and Anti-Causal Edges} \label{subsec:grads_under_var}
We explain how varsortability may dominate the performance of continuous structure learning algorithms.
We do not expect combinatorial structure learning algorithms that use a  
score-equivalent (see e.g.~\cite{yang2002comparison,chickering2002learning}) criterion or scale-independent (conditional) independence tests to be dependent on the data scale.
This includes 
\textit{PC}, as local constraint-based algorithm, 
\textit{FGES} as locally greedy score-based search using a score-equivalent criterion,
and \textit{DirectLiNGAM}, a procedure minimizing residual dependence.
By contrast, combinatorial algorithms with a criterion that is not score-equivalent (such as the MSE) depend on the data scale.
Due to the optimization procedure,
continuous structure learning algorithms may depend on the data scale irrespective of whether the employed score is score-equivalent (as, for example, \textit{GOLEM} for Gaussian models)
or not (as, for example, \textit{GOLEM} under likelihood misspecification or \textit{NOTEARS}).

We first establish how varsortability affects the gradients 
of MSE-based score functions (which are akin to assuming equal noise variances in the Gaussian setting) and
when initialising with the empty graph $0_{d\times d}$
(as is done in \textit{NOTEARS} and \textit{GOLEM}).
Full statements of objective functions and respective gradients are found in~\cref{app:gradients}.
Since
$\nabla\MSE{W}{} \propto \X^\top(\X - \X W)$
we have that
$\nabla\MSE{0_{d\times d}}{} \propto \X^\top \X$
and
$\nabla\mathcal{L}_{EV}(0_{d\times d}) \propto \sfrac{1}{\|\X\|_2^{2}} \X^\top \X$.
The initial gradient step of both
\textit{NOTEARS} and \textit{GOLEM-EV}
is symmetric. 
We have $\nabla\operatorname{MSE}(W) \propto [\X^\top(x_1-\X w_1),...,\X^\top(x_d-\X w_d)]$
where the $j^\text{th}$ column $\X^\top (x_j - \X w_j)$ reflects the 
vector of empirical covariances
of the $j^\text{th}$ residual vector $x_j - \X w_j$ with each $x_i$.
Provided a small identical step size is used across all entries of $W$ in the first step
(as, for example, in \textit{GOLEM-EV}),
we empirically find the residual variance after the first gradient step to be larger in those components that have higher marginal variance
(see \cref{app:residualvariance} for a heuristic argument). 
We observe that during the next optimization steps
$\nabla\operatorname{MSE}(W)$
tends to be larger magnitude for edges pointing in the direction of nodes with high-variance residuals
{(which tends to be those with high marginal variance)}
than for those pointing in the direction of nodes with low-variance residuals
{(which tends to be those with low marginal variance)}.
Intuitively,
when cycles are penalized,
the insertion of edges pointing to nodes with high residuals is favored as a larger reduction in MSE may be achieved than by including the opposing edge.
Given high varsortability,
this corresponds to favoring edges in the causal direction. 
This way, the global information about the causal order in case of high varsortability 
is effectively exploited.

Once we allow for unequal noise variances as in \textit{GOLEM-NV}, the marginal variances lead the gradients differently.
Letting $\MSE{w_j}{j} = \frac{1}{n}\|x_{j} - \X w_j\|_2^2$,
we have
\begin{align*}
\nabla\mathcal{L}_{NV}(W)
&\propto \left[\frac{\X^\top(x_j-\X w_j)}{\MSE{w_j}{j}}\right]_{j=1,...,d}
\end{align*}
such that the logarithmic derivative
breaks the symmetry of the first step for the non-equal variance formulation of \textit{GOLEM}
and we have
$\nabla\mathcal{L}_{NV}(0_{d\times d}) \propto
\X^\top \X
\operatorname{diag(\|x_1\|_2^{-2}...,\|x_d\|_2^{-2})}$.
While $\nabla_W\operatorname{MSE}(W) \propto [\X^\top(x_j - \X w_j)]_{j=1,...,d}$ tends to favor edges in causal direction (see above),
the column-wise inverse MSE scaling of $\X^\top(x_j - \X w_j)$ by $\MSE{w_j}{j}$ (the residual variance in the $j^\text{th}$ component)
leads to larger-magnitude gradient steps
for edges pointing in the direction of low-variance nodes rather than high-variance nodes. 
Given high varsortability,
this corresponds predominantly to the anti-causal direction.

We conjecture that the first gradient steps have a dominant role in determining the causal structure, even though afterwards the optimization is governed by a non-trivial interplay of optimizer, model fit, constraints, and penalties. 
For this reason we focus on the first optimization steps to explain
a)~why continuous structure learning algorithms that assume equal noise variance work remarkably well in the presence of high varsortability 
and b)~why performance changes once data is standardized and the marginal variances no longer hold information about the causal order.
Because of the acyclicity constraint, it may be enough for a weight $w_{i \to j}$ to be greater in magnitude than its counterpart $w_{j \to i}$ early on in the optimization for the smaller edge to be pruned from there on. For a discussion of the interplay between sparsity penalties and data scale see \cref{sec:landscape},
which indicates that the nodes need to be on a
comparable data scale for $l^1$-penalization to be well calibrated.
\cite{ng2020role} provide further discussion on sparsity and acyclicity constraints in continuous DAG learning.

\subsection{\textit{sortnregress}: A Diagnostic Tool to Reveal Varsortability} \label{sec:baselines}

We propose an algorithm \textit{sortnregress} performing the following two steps: 

\begin{description}[wide=0pt,topsep=0pt,itemsep=2pt]
 \item[order search] Sort nodes by increasing marginal variance.
 \item[parent search] Regress each node on all of its predecessors in that order, using a sparse regression technique to prune edges \citep{shojaie2010penalized}. We employ Lasso regression \citep{tibshirani1996regression} using the Bayesian Information Criterion \citep{schwarz1978estimating} for model selection.
\end{description}

As a baseline, \textit{sortnregress} is easy to implement (cf.~\cref{sec:sortnregress_code})
and
highlights and evaluates to which extent
the data scale is informative of the causal structure
in different benchmark scenarios.
An extension for non-linear additive noise models
is obtained by using an appropriate
non-linear regression technique in the parent search step,
possibly paired with cross-validated recursive feature elimination.
It facilitates a clear and contextualized assessment of different structure learning algorithms
in different benchmark scenarios. 
The relationship between varsortability and the performance of sortnregress in a linear setting is shown in \cref{app:varsortability_sortnregress}.
Varying degrees of varsortability and performance of \textit{sortnregress} add an important dimension
which current benchmarks do not consider.

\section{Simulations}\label{sec:simulations}  %

We compare the performance of the algorithms introduced in \cref{subsec:algorithms} on raw and standardized synthetic data.
In our comparison, we distinguish between settings with different noise distributions, graph types, and graph sizes. 
Our experimental set-up follows
those in \cite{zheng2018dags, ng2020role}
and we contribute results obtained repeating their experiments
in \cref{appendix:detailed_results}.
We complement our and previous DAG-recovery results
by additionally evaluating how well the DAG output by continous structure learning
algorithms identifies the MEC of the ground-truth DAG.

\subsection{Data Generation}
We sample Erdös-Rényi (ER) \citep{erdHos1960evolution} and Scale-Free (SF) \citep{barabasi1999emergence} graphs and the parameters for
ANMs
according to the simulation details in \cref{table:data_params}. 
For a graph specified as ER-$k$ or SF-$k$ with $d$ nodes, we simulate $dk$ edges.
For every combination of parameters,
we create a raw data instance
and a standardized version that is de-meaned and re-scaled to unit variance.
On standardized data, we have varsortability $v=\frac{1}{2}$
and the marginal variances hold no information about the causal ordering of the nodes.
In all our experimental settings, varsortability averages above $0.94$ on the raw data scale (cf.\ \cref{appendix:varsortability_linear}).

\begin{table}[H]
    \caption{Parameters for synthetic data generation.}
    \label{table:data_params}
    \small
    \centering
    \begin{tabular}{lc|lc}
    \toprule
        Repetitions & 10 &  
        Edge weights & iid\  $\operatorname{Unif}((-2,-.5)\cup(.5,2))$\\ 
        Graphs & ER-2, SF-2, SF-4 &
        Noise distributions & Exponential, Gaussian, Gumbel \\  
        Nodes & $d\in\{10, 30, 50\}$ &
        Noise 
        standard deviations &
        $1$ (Gaussian-EV);
        iid\  $\operatorname{Unif}(.5,2)$ (others) \\
        Samples & $n=1000$ \\ 
        \bottomrule
    \end{tabular}
\end{table}

\subsection{Evaluation}

We evaluate performance using structural intervention distance (SID) 
\citep{peters2015structural} and structural Hamming distance (SHD)
between the recovered graph and ground truth.
Additionally, we contribute a performance assessments of continuous structure learning algorithms
in terms of the SID and SHD between the ground-truth MEC and the recovered MEC (PC, FGES) or the MEC identified by the recovered graph (NOTEARS, GOLEM).
SID assesses in how far the recovered structure enables to correctly predict the effect of interventions.
SHD measures the distance
between the true and recovered graph
by counting how many edge insertions, deletions, and reversals
are required to turn the former into the latter.
Since interventional distributions can consistently be estimated
given only the causal ordering
\citep{buhlmann2014cam}, SID is less susceptible to arbitrary choices of edge pruning procedures and thresholds than SHD.
Intuitively, SID prioritizes the causal order,
while SHD prioritizes the correctness of individual edges.
We follow common practices for edge thresholding and scoring (see \cref{appendix:detailed_results}).

\subsection{Performance on Raw Versus Standardized Data}
We group our algorithms into combinatorial and continuous algorithms. 
We propose a novel baseline algorithm termed \textit{sortnregress} which serves as a reference marking the performance achievable by directly exploiting marginal variances. We indicate its performance on standardized data as \textit{randomregress}, since it amounts to choosing a random order and regressing each node on its predecessors.
We use boxplots aggregating the performance achieved in the $10$ repetitions on raw and standardized data by each algorithm
and create separate plots per
noise type to account for identifiability and MLE specification differences. 
We show the results obtained for ER-2 graphs 
and Gaussian noise with non-equal variances
in \cref{fig:main_norm_sid}.
These results are representative of
the results obtained for different graphs and noise distributions
(cf.\ \cref{appendix:detailed_results}).
For the simulated settings,
varsortability is high ($>0.94$) on the raw data scale
(cf.\ \cref{appendix:varsortability_linear}).

\begin{figure}[t]
    \centering
    \textit{Recovery of ground-truth DAG}\\
    \begin{subfigure}[b]{.49\linewidth}
        \centering
        \includegraphicsmaybe[width=\linewidth]
        {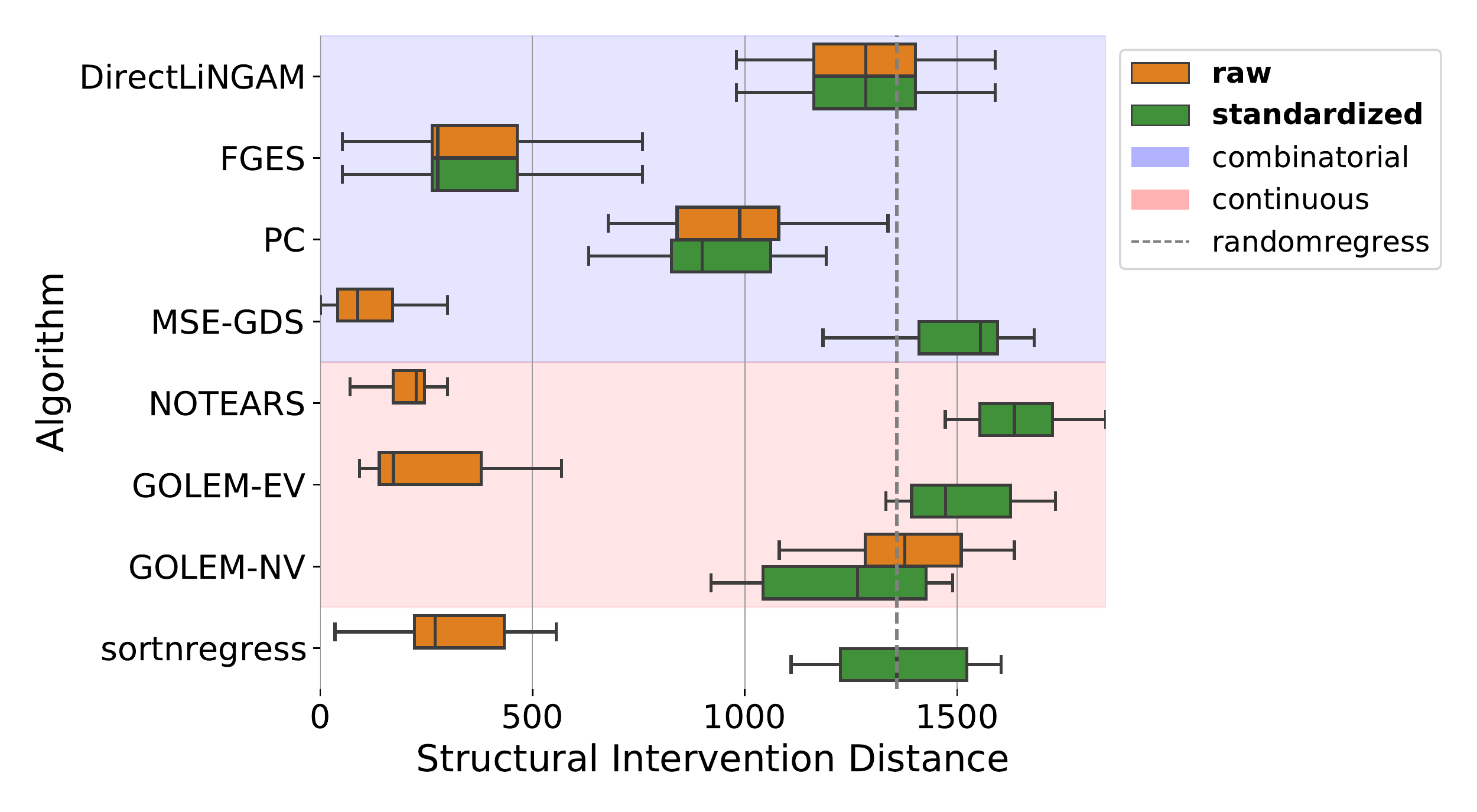}
    \end{subfigure}
    \hfill
    \begin{subfigure}[b]{.49\linewidth}
        \centering
        \includegraphicsmaybe[width=\linewidth]
        {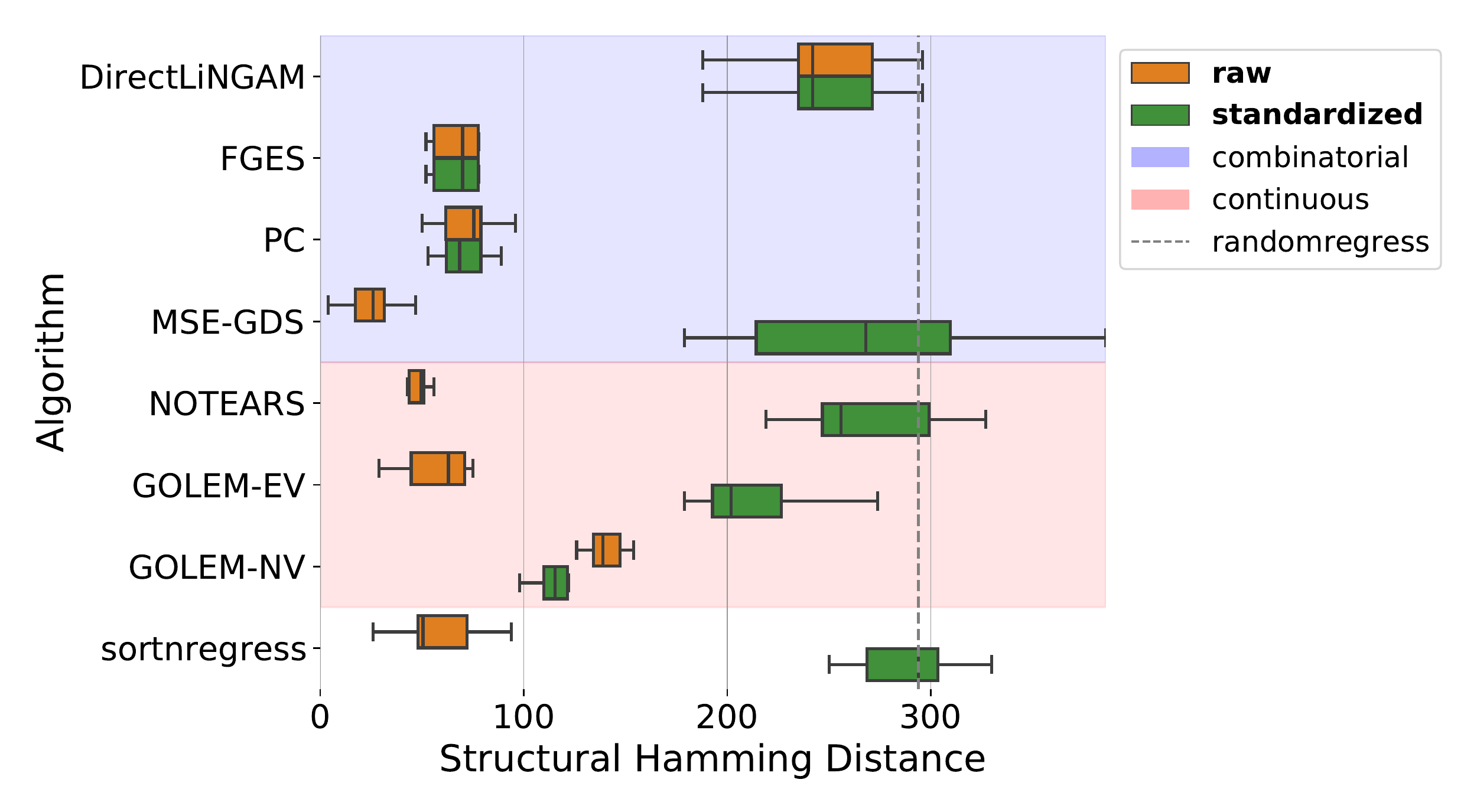}
    \end{subfigure}
    \textit{Recovery of ground-truth DAG's Markov equivalence class}\\
    \begin{subfigure}[b]{.49\linewidth}
        \centering
        \includegraphicsmaybe[width=\linewidth]{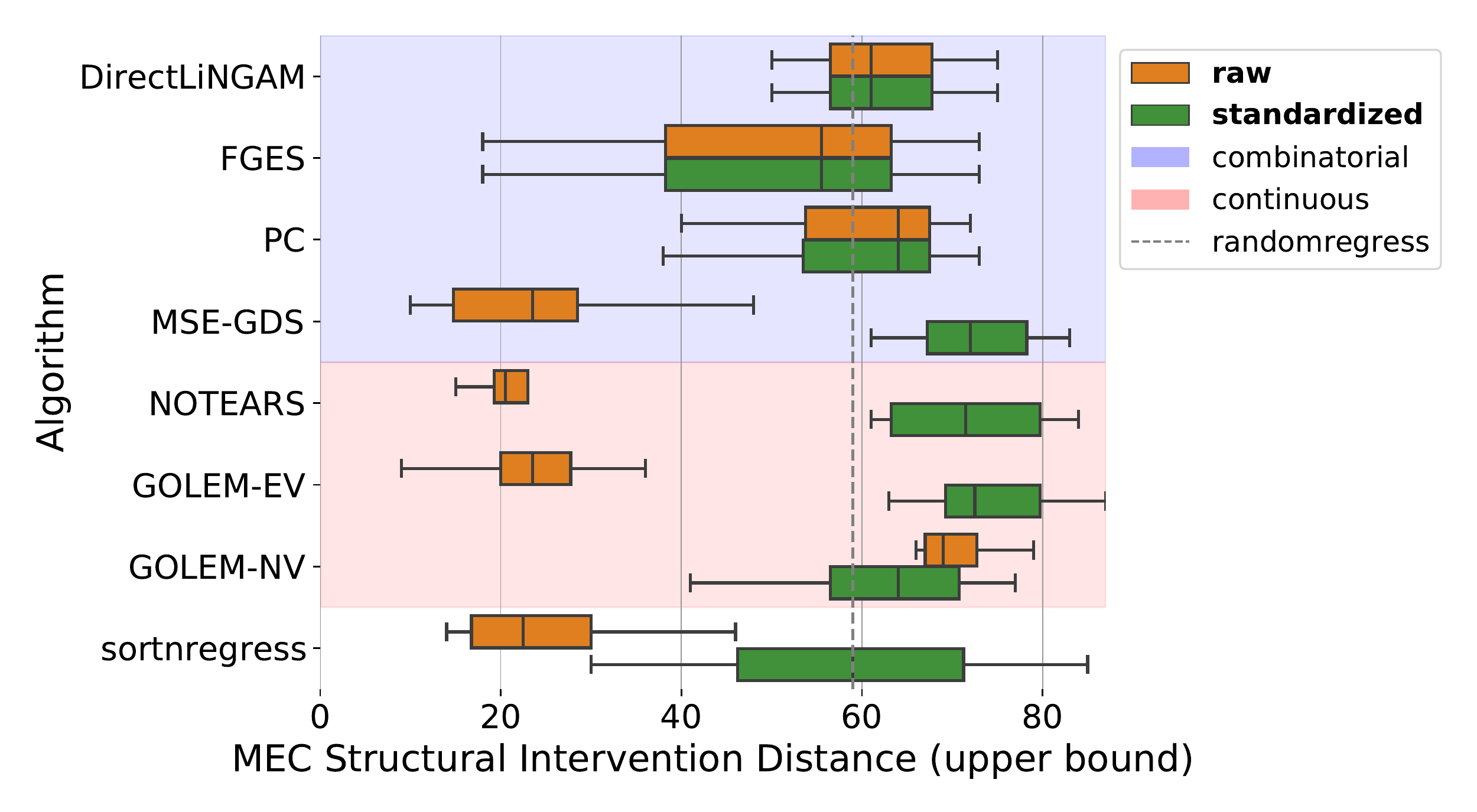}
    \end{subfigure}
    \hfill
    \begin{subfigure}[b]{.49\linewidth}
        \centering
        \includegraphicsmaybe[width=\linewidth]{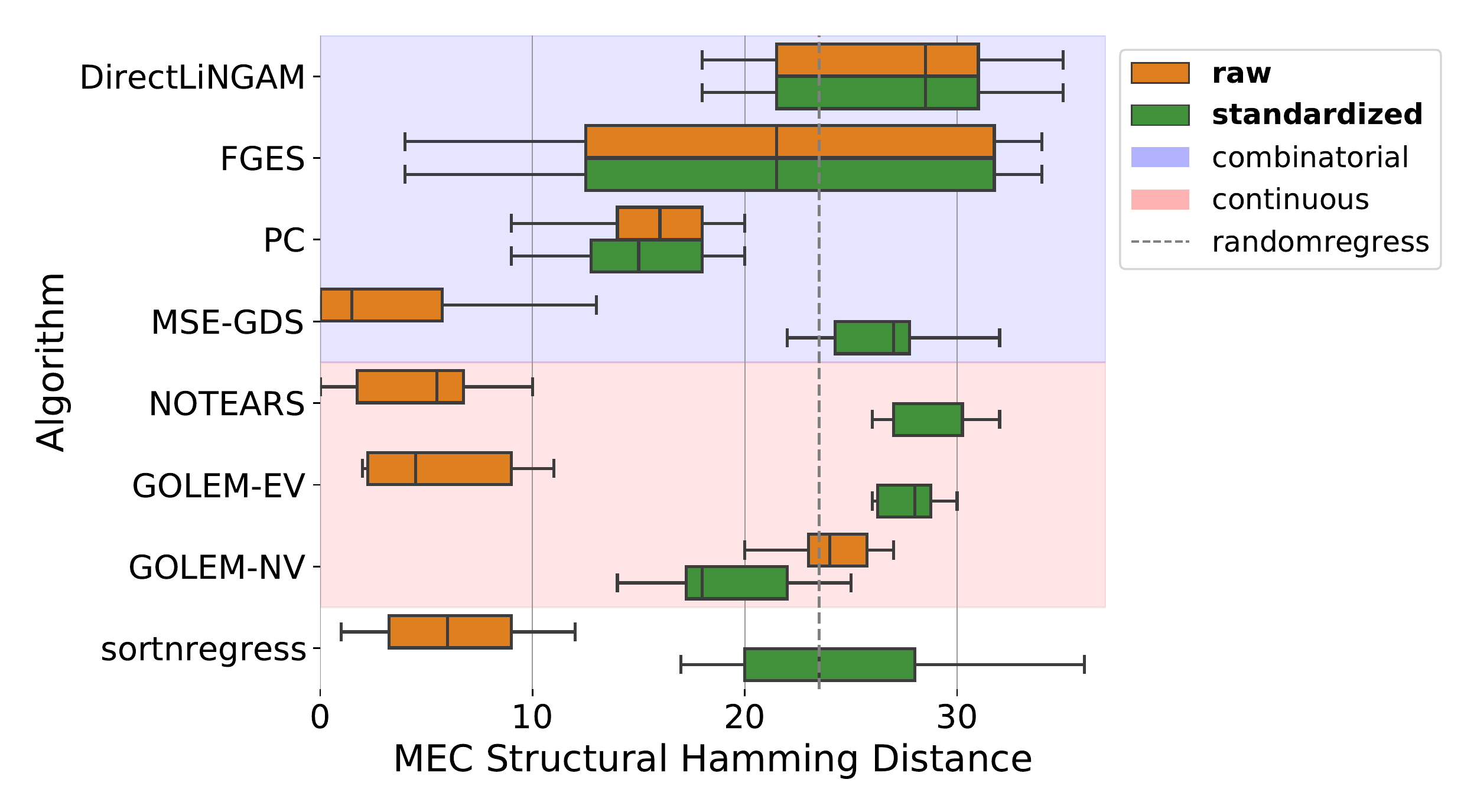}
    \end{subfigure}
    \caption{SID (left, lower is better) and SHD (right, lower is better) between recovered and ground-truth graphs (top) or Markov equivalence classes (bottom) for ER-2 graphs with 50 (top) or 10 (bottom) nodes and Gaussian-NV noise.
    The performance of \textit{sortnregress}, which only exploits varsortability,
    matches that of the continuous methods \textit{NOTEARS} and \textit{GOLEM}.
    }
    \label{fig:main_norm_sid}
\end{figure}

We observe that some algorithms are highly
scale-sensitive and perform vastly different on raw and standardized data. 
The algorithms \textit{NOTEARS}, \textit{MSE-GDS}, \textit{GOLEM-EV} are most affected -- their performance is excellent on raw data but far worse on standardized data.
Note that all of these rely on a loss function that revolves around the MSE.
The performance of \textit{GOLEM-NV} is also scale-sensitive but improves upon standardization.
The direction of the effect of standardization is in line with the predictions by our gradient analysis in \cref{subsec:grads_under_var}.
Note that we initialize all algorithms with the empty graph since we are primarily interested in comparing the impact of standardization given equal starting conditions. On standardized data, an initialization of \textit{GOLEM-NV} with the results of \textit{GOLEM-EV}, as recommended by \cite{ng2020role}, does not improve performance and may fail to converge. 
\textit{sortnregress}  achieves competitive performance on raw, 
and baseline performance on standardized data.
It thus qualifies as diagnostic tool to
highlight how much of a given causal structure learning task
can be resolved by exploiting data scale and sorting nodes by their marginal variance.

In summary, the evidence corroborates our claim that
the remarkable performance on raw data and the overall behavior upon standardization of the continuous structure learning algorithms may be driven primarily by high varsortability.
On a real-world causal protein signaling dataset \citep{sachs2005causal} we measure a mean varsortability of $0.57$ (which is close to chance level at 0.5) with a standard deviation of $0.01$ across our bootstrapped samples and do not observe the consistent performance pattern described for synthetic data with high varsortability (cf. \cref{app:sachs}).

\section{Gaming Further Benchmarks}

\subsection{Orienting Causal Chains on Standardized Data}

In order to design a causal discovery benchmark
that does not favor methods that explicitly exploit
marginal variances we may
standardize the data or employ
coefficient re-scaling schemes.
\cite{mooij2020joint}, for example,
propose a scale-harmonization by dividing each column $w_j = [w_{k\to j}]_{j=1,...,d}\in\mathbb{R}^d$
of the drawn adjacency matrices
by $\sqrt{\|w_j\|^2 + 1}$
such that each variable would have comparable
scale if all its direct parents were independently standard-normally distributed.
However, this does not avoid the problem of potentially inadvertent patterns in simulated ANMs. Even after standardization or scale-harmonization, DAGs with previously high varsortability generate data with distinct covariance patterns that may be exploited.

In~\cref{app:standardization} we present an instructive example of a decision rule
that can infer the orientation of a causal chain from raw, standardized, and scale-harmonized data with accuracy strictly greater than 50\%.
For a causal chain
$X_1 \to X_2 \to ... \to X_d$
where edge weights and noise terms are drawn iid
we can decide between 
the two Markov-equivalent graphs
$X_1 \to X_2 \to ... \to X_d$
and
$X_1 \gets X_2 \gets ... \gets X_d$
whith above-chance accuracy.
The empirical results for varying
chain-lengths
and various edge-weight distribution are deferred to the appendix where we discuss the 3-variable chain in detail and illustrate that the phenomenon extends from finite-sample to the population setting.

The intuition is as follows.
Consider data
generated by $X_1 \to X_2 \to ... \to X_d$ and the aim is to infer
from standardized data whether $X_1 \to ... \to X_d$ or $X_1 \gets ... \gets X_d$.
For data with high varsortability and comparable noise variance on the raw data scale it holds that the further downstream a node $X_i$ is in the causal chain,
the stronger the variance of its parent $\var{X_{i-1}}$
contributes to its marginal variance $\var{X_i} = \var{X_{i-1}} + \var{N_i}$ relative
to its noise variance $\var{N_i}$,
and the stronger is it correlated with its parent.
Thus, the sequence of regression coefficients,
which in the standardized case amounts to $(\cor(X_i, X_{i+1}))_{i=1,...,d-1}$,
tends to increase in magnitude along the causal order
and decrease in the anti-causal direction.
The proposed decision rule predicts the causal direction as the one
in which the absolute values of the regression coefficients tend to increase.
This chain orientation rule achieves above-chance performance on raw, standardized, and scale-harmonized data (cf.~\cref{app:standardization}).

\subsection{Sorting by Variance in Non-Linear Settings}\label{app:nonlinear_sorting}

Varsortability may also be exploited in non-linear settings.
\cref{tab:nonlin_baseline} shows the results of sorting by marginal variance
and filling in all edges from lower-variance nodes to higher-variance nodes in a non-linear setting.
This variance sorting strategy is more naive than \textit{sortnregress} and places no assumption on the functional form.
The results are substantially better than random sorting and may therefore be a more informative baseline than commonly used random graphs.
We do not show performance in terms of SHD,
as our variance sorting baseline always yields a fully connected graph.
Although the data generating process is not identical, we note that the improvement of our crude variance sorting over random sorting compares favorably to some of the improvements gained by more involved methods over random graphs as shown in \citet[Table 1]{lachapelle2019gradient}. 
Our results indicate that exploiting varsortability may also deliver competitive results in non-linear settings.
{
\begin{table}[H]
    \centering
    \caption{SID of naive baselines on non-linear data. Results on $1000$ observations of additive Gaussian process ANMs with noise variance 1 simulated as by \cite{zheng2020learning} ($10$ repetitions each; average varsortability $\varv$ per graph type shown in parentheses).}
    \label{tab:nonlin_baseline}
    \vspace{.5em}
    \begin{tabular}{l|lllll} 
    \toprule
    Algorithm & Graph & ER-1 & ER-4 &  SF-1 &  SF-4 \\
              & (average $\varv$) & ($0.87$) & ($0.95$) & ($0.95$) & ($0.98$)\\
    \midrule
    variance sorting &&   $7.7\pm \phantom{0}5.72$   & $25.2\pm 12.36$& $\phantom{0}1.9\pm 2.28$& $\phantom{0}7.6\pm 3.37$ \\
    random sorting && $27.9\pm 11.44$ & $63.1\pm \phantom{0}8.10$& $22.3\pm 13.14$ & $59.5\pm 7.32$ \\
    \bottomrule
    \end{tabular}
\end{table}
}
We find similarly high levels of varsortability for many non-linear functional relationships and graph parameters (cf. \cref{appendix:varsortability_nonlinear}).
This begs the question how much other successful methods exploit varsortability,
how they compare to non-linear nonparametric methods that leverage assumptions on the residual variances~\citep{gao2020a},
and how they perform under data standardization.
We encourage such an exploration in future work and suggest that varsortability and \textit{sortnregress} or \textit{variance sorting}
should always be included in future benchmarks.
\section{Discussion and Conclusion}\label{sec:conclusion}

We find that continuous structure learning methods
are highly susceptible to data rescaling
and some do not perform well without access to the true data scale.
Therefore, scale-variant causal structure learning methods
should be applied and benchmarked with caution,
especially if the variables do not share a measurement scale 
or when the true scale of the data is unattainable.
It is important to declare whether
data is standardized prior to being fed to various
structure learning algorithms.

Following the first release of the present paper,
\cite{kaiser2021unsuitability} also independently reported
the drop in performance of \textit{NOTEARS} upon standardizing the data
and presented a low-dimensional exemplary case.
Beyond a reporting of impaired \textit{NOTEARS} performance,
we also analyze score-equivalent methods, 
provide exhaustive simulation experiments,
and explain the phenomenon.

Our aim is to raise awareness of the severity
with which scaling properties in data from simulated DAGs and causal additive models may distort algorithm performance. 
Increasing marginal variances can render scenarios identifiable, which may commonly not be expected to be so—for example
the Gaussian case with non-equal variances.
We therefore argue that varsortability should be taken into account for future benchmarking.
Yet, with any synthetic benchmark there remains
a risk that the results are not indicative of algorithm performance on real data. 
Our results indicate that current structure learning algorithms 
may perform within the range of naive baselines on real-world datasets.

The theoretical results of our paper are limited to the setting of linear ANMs. Additionally, our conjecture regarding the importance of the first gradient steps, and with it a rigorous causal explanation for the learning behavior of different continuous algorithms and corresponding score functions remain open and require further research to be settled.
Our empirical findings indicate that causal discovery benchmarks can be similarly gamed on standardized data and in non-linear settings, but further research is needed to confirm this. We focus on a specific subset of algrorithms, the impact of patterns in benchmarking data on a wider class of algorithms and score functions remains to be explored.

Varsortability arises in many ANMs
and the marginal variances increase drastically along the causal order,
at least in common simulation settings.
This begs the question what degree of varsortability can be observed or assumed in real-world data. If the marginal variances carry information about the causal order, our results suggest that it can and should be leveraged for structure learning. Otherwise, our contribution motivates future research into representative benchmarks and may put the practical applicability of the additive noise assumption into question.

\subsection*{Acknowledgements}
We thank
Jonas M.\ Kübler,
Jonas Peters,
and
Sorawit Saengkyongam
for helpful discussions and comments.
SW was supported by the Carlsberg Foundation.

\clearpage

\section*{References}

{\small
\bibliography{references}
}

\appendix

\clearpage

\addcontentsline{toc}{section}{Appendix} %
\part{Appendix} %
\parttoc %

\clearpage
\section{Varsortability in the Two-Node Case} \label{app:two_node}

Consider the following ground truth 
and two competing linear acyclic models:

\begin{minipage}[t]{.3\textwidth}
Ground-truth model:
\begin{align*}
    A &:= N_A \\
    B &:= wA +N_B
\end{align*}
\end{minipage}
\hfill
\begin{minipage}[t]{.3\textwidth}
Model \textit{M1)} ``$A \to B$'':
\begin{align*}
    \widehat{A} &= 0 \\
    \widehat{B} &= \widehat{w} A
\end{align*}
\end{minipage}
\hfill
\begin{minipage}[t]{.3\textwidth}
Model \textit{M2)} ``$A \gets B$'':
\begin{align*}
    \widehat{A} &= \widehat{v} B \\
    \widehat{B} &= 0 
\end{align*}
\end{minipage}\\[1em]
where $w\neq 0$,
$N_A$ and $N_B$
are independent zero-centred noise terms that follow some distributions
with non-vanishing corresponding variance $V_A$ and $V_B$.
The model parameters $\widehat{w} = \frac{\covar{A}{B}}{\var{A}}=w$
and $\widehat{v} = \frac{\covar{A}{B}}{\var{B}} = \frac{wV_A}{\var{B}}$
are the corresponding ordinary least-squares linear regression coefficients.

We evaluate in which cases the true model \textit{M1}
obtains a smaller MSE than the wrong model \textit{M2},
to decide if and under which conditions
a MSE-based orientation rule recovers the
ground-truth edge direction:

\begin{align*}
    &&\MSE{M1}{} &< \MSE{M2}{}  \\
    \iff &&
    \var{A} + \var{B - \widehat{w}A} &< \var{A - \widehat{v}B} + \var{B} \\
    \iff &&
    \var{A} + \operatorname{Var}\left((wN_A + N_B) - wN_A\right)
    &<
    \operatorname{Var}\left({N_A - \frac{wV_A}{\var{B}}(wN_A+N_B)}\right) + \var{B} \\
    \iff &&
    V_A + V_B
    &<
    \frac{V_AV_B}{\var{B}} + w^2V_A + V_B
    \\
    \iff &&
    0 &< 
    V_A\left(\frac{V_B}{\var{B}} - 1\right) + w^2 V_A \\
    \iff &&
    0 &< \frac{-w^2 V_A}{\var{B}} + w^2 \\
    \iff &&
    V_A &< \var{B} \\
    \iff &&
    (1-w^2)V_A &< V_B
\end{align*}

For error variances $V_A  \leq V_B$ 
and any non-zero edge weight $w$, the MSE-based inference is correct.
This resembles known scale-based identifiability results
based on equal or monotonically increasing error variances~\citep{peters2014identifiability,park2020identifiability}.

\section{Derivation of Lower Bound on Pairwise Varsortability}\label{app:derivation_bound}

Let $A$ and $B$ be any two nodes in the sampled graph
with edge $A\overset{w}{\to} B$, noise terms $N_A, N_B$, and without common ancestors and no other directed path from $A$ to $B$.
When sampling edge coefficients and noise variances randomly
for the simulation of ANMs, distributions are incurred over the variances of $A$ and $B$ across those simulated ANMs.
Let edge weights be sampled as
$[W_{x\to y}]_{x,y=1,...,d}\sim\mathbb{P}_W$,
and noise variances be sampled as
$[S^2_{N_y}]_{y=1,...,d}\sim\mathbb{P}_{S^2}$.
Across simulations, the marginal variances of $A$ and $B$ are transformations of $S$ and $W$ and themselves random variables denoted as $V_A$ and $V_B$.
The marginal variance $V_Y$ of any node $Y$ depends on its noise variance and the additional variance incurred by predecessor nodes given as $\sum_{X \in \PA{Y}}W_{X\to Y}^2V_X$. We can therefore bound the probability for the variable pair $(A,B)$ to be varsortable from below via
\begin{align*}
    \operatorname{P}_{}[{V_A < V_B}]
    = &\operatorname{P}\left[V_A < \left({W_{AB}^2 V_A + \sum_{X\in\PA{B}\setminus\{A\}}W_{XB}^2V_X + S_{N_B}^2}\right)\right] \\
    \geq &\operatorname{P}[V_A < W_{AB}^2 V_A + S_{N_B}^2] \\
\end{align*}  
where equality holds if $A$ is the only parent of $B$ contributing to $B$'s marginal variance.

In common benchmarks, edge weights are drawn independently according to
$\mathbb{P}_W \sim \otimes_{k,j=1,...,d}\operatorname{Unif}((-2,-.5)\cup(.5,2))$
and
noise standard deviations are drawn iid $S_{N_j} \sim \operatorname{Unif}(.5, 2)$.

\section{Varsortability and Identifiability by Conditional Noise Variances}\label{sec:varsort_park}

While closely related,
varsortability {is not equivalent to}
the identifiability conditions laid out in Theorem 4, \cite{park2020identifiability},
(henceforth referred to as ``Theorem 4'').
We prove this by providing two examples.
In \cref{subsec:park_A} part A) of the conditions in Theorem 4 is satisfied,
while varsortability does not hold.
In \cref{subsec:park_B} varsortability holds but neither part A) nor part B) of Theorem 4 are satisfied.

\subsection{Park, 2020, Theorem 4 conditions satisfied without varsortability} \label{subsec:park_A}

Consider the following ground-truth model
with unique causal order $A, B, C$:
\begin{align*}
A &:= N_A\\
B &:= \beta_{A\to B}A + N_B = 1A + N_B \\
C &:= \beta_{B\to C}B + N_C = \sqrt{\frac{2}{3}}B + N_C
\end{align*}
where $N_A,N_B,N_C$
are jointly independent zero-centred noise terms
with respective variances $\sigma_A^2 = 4, \sigma_B^2 = 2, \sigma_C^2 = 1$.
The marginal variances are
$\var{A} = 4 \ < \var{C} = 5 \ < \var{B} = 6$.
Our example resembles the examples in Section 3.1 of \cite{park2020identifiability}.
We can verify the three conditions for part A) of Theorem 4:

\begin{align*}
    (A1) \quad & \sigma_A^2 < \sigma_B^2 + \beta_{A \to B}^2\sigma_A^2, \\
    (A2) \quad & \sigma_B^2 < \sigma_C^2 + \beta_{B \to C}^2\sigma_B^2, \\
    (A3) \quad & \sigma_A^2 < \sigma_C^2 + \beta_{B \to C}^2\sigma_B^2 + \beta_{A \to B}^2\beta_{B \to C}^2\sigma_A^2
\end{align*}

Inserting the values from above, we obtain

\begin{minipage}[t]{.3\textwidth}
    \begin{align*}
        (A1) \quad 4 < 2 + 1\cdot4, 
    \end{align*}
    \end{minipage}
    \hfill
    \begin{minipage}[t]{.3\textwidth}
    \begin{align*}
        (A2) \quad 2 < 1 + \frac{2}{3}\cdot2, 
    \end{align*}
    \end{minipage}
    \hfill
    \begin{minipage}[t]{.3\textwidth}
    \begin{align*}
        (A3) \quad 4 < 1 + \frac{2}{3}\cdot2 + 1\cdot\frac{2}{3}\cdot4.
    \end{align*}
\end{minipage}\\[1em]

Our result verifies that identifiability is given as per Theorem 4 in \cite{park2020identifiability},
while the order of increasing marginal variances is not in complete agreement with the causal order and varsortability is not equal to $1$.

\subsection{Varsortability without Park, 2020, Theorem 4 conditions satisfied} \label{subsec:park_B}

Consider the following ground-truth model
with unique causal order $A, B, C$:
\begin{align*}
A &:= N_A\\
B &:= \beta_{A\to B}A + N_B = A + N_B \\
C &:= \beta_{A\to C}A + \beta_{B\to C}B+ N_C = \frac{1}{\sqrt{2}} A + \frac{1}{\sqrt{2}} B + N_C
\end{align*}
where $N_A,N_B,N_C$
are jointly independent zero-centred noise terms
with respective variances $\sigma_A^2 = 4, \sigma_B^2 = 3, \sigma_C^2 = 1$.
The marginal variances are
$\var{A} = 4 \ < \var{B} = 7 \ < \var{C} = 10.5$. We now verify, that for both case A) and B) in
Theorem 4 of \cite{park2020identifiability}
at least one of the inequality constraints is violated.

One of the three conditions in A) is
\begin{align*}
\sigma_B^2 < \sigma_C^2 + \beta_{B \to C}^2\sigma_B^2, 
\end{align*}
while for the above model we have
\begin{align*}
3 \nless 1 + \frac{1}{2}\cdot3. 
\end{align*}

One of the three conditions in B) is
    \begin{align*}
        \frac{\sigma_C^2}{\sigma_B^2} > (1-\beta_{B \to C}^2), 
    \end{align*}
while for the above model we have
\begin{align*}
\frac{1}{3} \ngtr (1-\frac{1}{2}). 
\end{align*}

For both criteria A) and B) in Theorem 4 at least one of the inequalities is not satisfied.
We thus verify that even if identifiability is not given as per the sufficient conditions in
Theorem 4, \cite{park2020identifiability},
varsortability may still render the causal order identifiable.

\section{Algorithms}\label{app:algorithms}

\paragraph{DirectLiNGAM} 
is a method for learning linear non-Gaussian acyclic models \citep{shimizu2011directlingam}. It recovers the causal order by iteratively selecting the node whose residuals are least dependent on any predecessor node.
In a strictly non-Gaussian setting, \textit{DirectLiNGAM}
is guaranteed to converge to the optimal solution asymptotically within a small fixed number of steps and returns a DAG.
We use the implementation provided by the authors\footnote{\url{https://github.com/cdt15/lingam}}.
We deliberately keep the default of a least-angle regression penalized by the Bayesian Informaion Criterion.
We find that this penalty strikes a good balance between SID and SHD performance.
Cross-validated least-angle regression performs better in terms of SID but poorer in terms of SHD.

\paragraph{PC}
\citep{spirtes1991algorithm}
is provably consistent in estimating the Markov equivalence class of the true data-generating graph
if the causal Markov and faithfulness assumptions hold.
The algorithm returns a completed partially directed acyclic graph (CPDAG).
For computational reasons, we refrain from computing the lower and upper bounds of the SID for comparing CPDAGS with the ground-truth DAG as proposed by \cite{peters2015structural}.
Instead, we adopt the approach by \cite{zheng2018dags} and resolve bidirectional edges favorably to obtain a DAG. We use the implementation in the \emph{Tetrad}\footnote{\url{https://github.com/cmu-phil/tetrad}} package \cite{ramsey2018tetrad}.

\paragraph{FGES} is an optimized version of the fast greedy equivalence search algorithm developed
by~\cite{meek1997graphical, chickering2002optimal}.
Under causal Markov and faithfulness assumptions, 
it is provably consistent for estimating the Markov equivalence class of the true data-generating graph. 
The algorithm returns a CPDAG, which we resolve favorably to obtain a DAG. We use the implementation in the 
\emph{Tetrad}\footnote{\url{https://github.com/cmu-phil/tetrad}} package \citep{ramsey2018tetrad}.

\paragraph{MSE-GDS} is a greedy DAG search procedure with a MSE score criterion.
We implement \textit{MSE-GDS} following other GDS procedures,
for example, as described by \citet[Section 4]{peters2014identifiability},
but use the MSE as score criterion instead of a likelihood- or BIC-based score criterion.
For simplicity and computational ease,
we consider a smaller search space and greedily forward-search over new edge insertions only
instead of greedily searching over all neighbouring DAGs obtainable by edge insertions, removals, and deletions.
For the linear setting,
linear regression is used to determine the edge weights
and the corresponding MSE-score for a given graph.
For the non-linear setting, support vector regression can be used instead.
The algorithm returns a DAG.

\paragraph{NOTEARS}
is a score-based method that finds both structure and parameters simultaneously by continuous optimization \citep{zheng2018dags}.
The optimization formulation is based on the mean squared error and includes a sparsity penalty parameter $\lambda$
and a differentiable acyclicity constraint:
\begin{align*}
    \argmin_{W \in \mathbb{R}^{d \times d}} \quad \MSE{W}{\X} + \lambda \|W\|_1 \quad \textrm{s.t.} \quad \tr(\exp(W \odot W)) - d = 0.
\end{align*}
The algorithm returns a DAG.
We use the implementation provided by the authors\footnote{\url{https://github.com/xunzheng/notears}}. 
Throughout all our experiments we use \textit{NOTEARS} over \textit{NOTEARS-L1} (setting $\lambda=0$),
following the findings of \citet[Tables 1 and 2, Figure 3]{zheng2018dags}, which
suggest regularization only for samples smaller than the $n=1000$ we use throughout.

\paragraph{GOLEM}
combines a soft version of the differentiable acyclicity constraint 
from \cite{zheng2018dags} with a MLE objective \citep{ng2020role}.
The authors propose a multivariate Gaussian MLE for equal (EV) or unequal (NV) noise variances and optimize
\begin{align*}
    \argmin_{W \in \mathbb{R}^{d \times d}} \quad \mathcal{\widetilde{L}}(W, \X) - \log(|\det(I-W)|) + \lambda_1 \|W\|_1 + \lambda_2(\tr(\exp(W \odot W))-d)
\end{align*}
where $\mathcal{\widetilde{L}}$ is either
\begin{align*}
    \mathcal{\widetilde{L}}_{EV}(W, \X) &= \frac{d}{2}\left(\mathcal{L}_{EV}(W, \X) + \log(n)\right) = \frac{d}{2}\log(n\MSE{W}{\X})\text{, or}\\
    \mathcal{\widetilde{L}}_{NV}(W, \X) &= \frac{1}{2}\left(\mathcal{L}_{NV}(W, \X) + d\log(n) \right) = \frac{1}{2}\sum_{j=1}^d \log(n\MSE{w_j}{j}).
\end{align*}

We use the implementation and hyperparameters provided by the authors\footnote{\url{https://github.com/ignavier/golem}}.
We train for $10^4$ episodes as we found that half of that suffices to ensure convergence.
Notably, we do not perform pretraining for our version of \textit{GOLEM-NV}.

\paragraph{sortnregress}
is implemented as shown in \cref{sec:sortnregress_code}. We find that
a least-angle regression penalized by the Bayesian Information Criterion strikes a good balance between SID and SHD performance.

\section{The~Subtle~Interplay~Between~Marginal~Variance~and~Gradient~Directions}\label{app:gradients}

We describe observations
about the gradients involved in the optimization procedures of \textit{NOTEARS} and \textit{GOLEM-EV/-NV}.
We present an instructive example in \cref{app:gradexample}
and provide some intuition about how the adjacency matrix changes throughout the optimization.
For convenience and reference we provide gradients of the individual terms involved in the respective objective functions (cf.\ \cref{app:gradients_individual}).
In \cref{app:residualvariance}
we argue why the nodes' residual variances for the first gradient steps
in an unconstrained optimization of MSE- or log-MSE-EV-based (\text{GOLEM-EV})
objective functions with acyclicity penalties
tend to follow the same ordering as the nodes' marginal variances.
We analyze gradient symmetry and asymmetry in \textit{GOLEM-EV/-NV}'s gradient descent optimization
under varsortability in \cref{app:gradients_golem}.
While the intuition for small step size gradient-based unconstrained optimization
partially carries over to the \textit{NOTEARS} optimization procedure,
here 
the interplay between varsortability and gradient directions
is intricate due to a constrained optimization that
is solved via the augmented Lagrangian method and dual descent with line-search
instead of gradient descent as used in \textit{GOLEM}~\citep{zheng2018dags}
(cf.\ \cref{app:gradientsnotears}).

The heuristic arguments presented here are preliminary and aim to provide intuition.
The optimization behaviour also heavily depends on the implementation of the optimization routine.
For example, the original implementation of \textit{NOTEARS} fixes the diagonal of $W$ at zero and leverages curvature information (L-BFGS-B),
while \textit{GOLEM} updates all entries of $W$ and employs learning rate optimizers.
Future research is required to determine how precisely continuous structure learning algorithms achieve state-of-the-art results on highly varsortable data and, given our observations, we expect explanations to be specific to individual algorithms and their distinct implementations.

\subsection{Example} \label{app:gradexample}

The following example considers the population limit and illustrates a few intuitions about gradient based optimization and varsortability.
Consider data is generated according to
\[
\begin{pmatrix}
 X \\ Y
\end{pmatrix}
=
\begin{pmatrix}
0 & \beta \\
0 & 0
\end{pmatrix}^\top
\begin{pmatrix}
 X \\ Y
\end{pmatrix}
+
\begin{pmatrix}
N_X \\ N_Y
\end{pmatrix}
\]
where $N_X$ and $N_Y$ are independently normally distributed with
standard deviations $\sigma_{N_X}$ and $\sigma_{N_Y}$.
Here, varsortability $\varv = 1$ and $1 = \operatorname{Var}{X} < \operatorname{Var}{Y} = 2$.

Initializing the weight matrix at the zero matrix,
the gradient of the population MSE is
\[
-2
\begin{pmatrix}
\operatorname{Var}(X) & 
\operatorname{Cov}(X, Y) \\
\operatorname{Cov}(Y, X) &
\operatorname{Var}(Y)
\end{pmatrix}
=
-2
\begin{pmatrix}
\sigma_{N_X}^2 & 
\beta\sigma_{N_X}^2 \\
\beta\sigma_{N_X}^2 &
\beta^2\sigma_{N_X}^2 + \sigma_{N_Y}^2
\end{pmatrix}
\]
(see also \cref{app:gradients_individual}).
The models for $X$ and $Y$ after a first gradient descent step of step size $\eta$ are
\begin{align*}
\widehat{X} &= 2\eta( \sigma_{N_X}^2 X + \beta\sigma_{N_X}^2 Y)
\\
\widehat{Y} &= 2\eta( \beta\sigma_{N_X}^2 X + (\beta^2\sigma_{N_X}^2 + \sigma_{N_Y}^2) Y)
\end{align*}
If the diagonal of the weight matrix is clamped to $0$ throughout the optimization,
the terms corresponding to self-loops ($2\eta\sigma_{N_X}^2X$ in $\widehat{X}$ and $(\beta^2\sigma_{N_X}^2 + \sigma_{N_Y}^2)Y$ in $\widehat{Y}$) are dropped above.
This is the case in the original implementation of NOTEARS, where the
unconstrained subproblem
is optimized via L-BFGS-B with identity bounds on the diagonal entries of $W$.

Below we visualize
$\operatorname{Var}(X - \widehat{X})$ (residual variance in $X$),
$\operatorname{Var}(Y - \widehat{Y})$ (residual variance in $Y$),
and the MSE $\operatorname{Var}(X - \widehat{X}) + \operatorname{Var}(Y - \widehat{Y})$,
for varying step sizes $\eta$ of the first gradient step
where we exemplary choose $\beta=\sigma_{N_X}=\sigma_{N_Y}=1$.

\begin{center}
\includegraphicsmaybe[keepaspectratio, width=.49\textwidth]{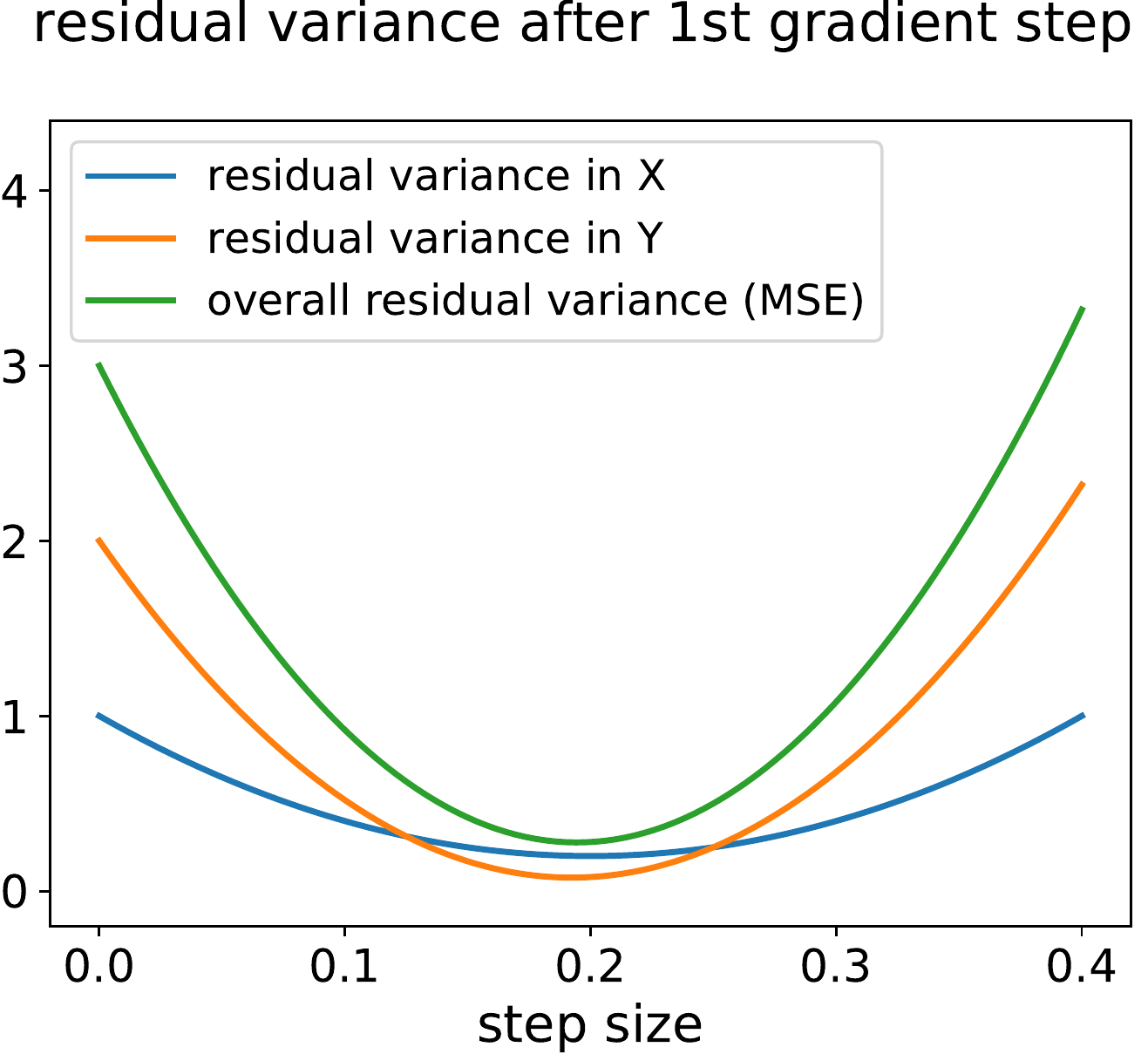}
\hfill
\includegraphicsmaybe[keepaspectratio, width=.49\textwidth]{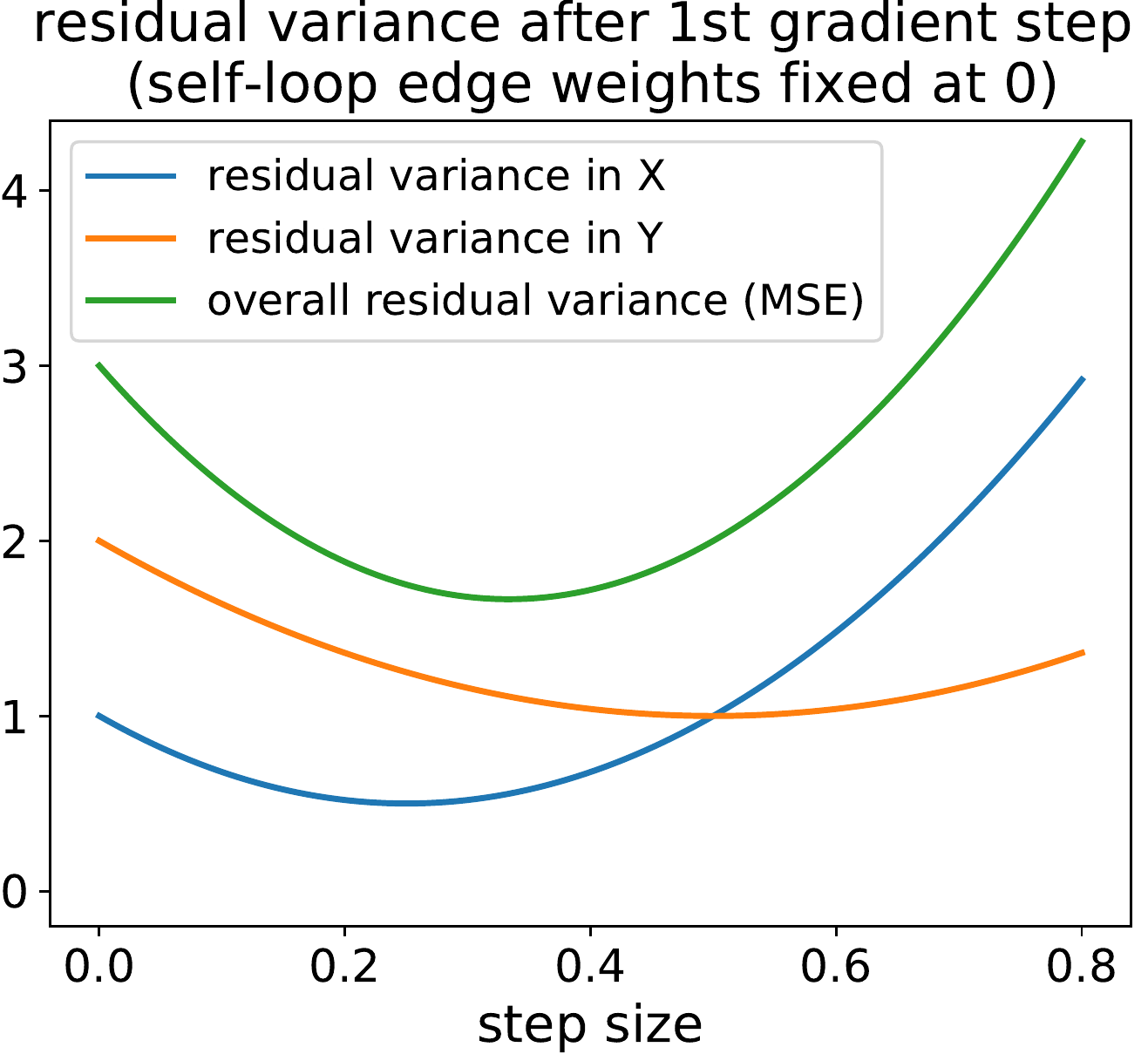}
\end{center}

Since the residual variances change continuously for increasing step sizes,
the residual variances follow the order of the marginal variances for small step sizes
(cf.\ also \cref{app:residualvariance}).
Since in GOLEM we solve an unconstrained optimization problem by gradient descent (with small step size and learning rate),
the order of residual variances tends to remain unchanged during the first optimization steps.
The order of the residual variances may swap relative to the order of marginal variances, though,
if line-search is employed to determine the step size that minimizes the MSE-objective.
This is the case in NOTEARS,
where the MSE is minimized by a dual descent routine with increasing weight on the acyclicity penalty term.
Here, the first symmetric update of the weight matrix occurs with a large step size that minimizes the MSE (minimum of the green curves in above plots).
The ordering of the resulting residual variances is less obvious.
In the above example, if the diagonal terms of the weight matrix are updated as well (left),
the residual variance order after the first gradient step is opposite to the marginal variance order.
If the diagonal entries are clamped at $0$ (as is the case in \textit{NOTEARS} and corresponding to the setting shown on the right),
the first gradient step in the above example leads to a scenario where the residual variance order follows the marginal variance order and where the resulting edge weight for the direction $X \gets Y$
overshoots the optimum, that is,
the blue curve's minimum is attained for a smaller step size than the green curve's minimum.
The intuition is as follows:
If we minimize the MSE the step size calibrates a trade-off between residual variances in the different nodes;
the high marginal variance nodes dominate the MSE such that the step size that minimizes the MSE
may result in ill-chosen edge weights for the edges incoming into low-variance nodes.
In the next optimization step,
the gradient of the MSE loss for the edge $X \to Y$ pushes towards increasing that edge weight,
while it pushes for decreasing the edge weight $X \gets Y$
(besides a gradient contribution from the acyclicity constraint).
As a result, the edge weights for $X\to Y$ and $X \gets Y$ are equal after the first step of \textit{NOTEARS},
but better calibrated for the direction from low- to high-variance nodes,
which here corresponds to the correct edge $X\to Y$.
In the subsequent optimization step, decreasing the edge weight $X\gets Y$ is
favored both by the MSE gradient and the acyclicity penalty, while for the correct edge
$X \to Y$ the MSE gradient pushes to further increasing the edge.
Intuitively, if one needs to cut one of the two edges to avoid cycles, it is ``cheaper'' in MSE to cut the wrong edge $X\gets Y$ from a high- to low-variance node.

\subsection{Stepwise Gradient Derivation} \label{app:gradients_individual}

\paragraph{MSE} For $\X \in \mathbb{R}^{n\times d}$, the gradient of $\MSE{W}{\X} = \frac{1}{n}\|\X - \X W\|_2^2$ is
\begin{align*}
\nabla_W \MSE{W}{\X} &= \frac{1}{n}\nabla_W \left(\operatorname{Tr}[\X^\top\X] - \operatorname{Tr} [W^\top\X^\top\X] - \operatorname{Tr}[\X^\top\X W] + \operatorname{Tr}[W^\top\X^\top\X W ] \right) \\
&= \frac{1}{n} \left(-\X^\top\X - \X^\top\X + \X^\top\X W + \X^\top\X W\right) \\
&= -\frac{2}{n} \left(\X^\top\X - \X^\top\X W \right) \\
&= -\frac{2}{n}\X^\top(\X - \X W) \\
&\propto \X^\top(\X - \X W)
\end{align*}
If $W$ is polynomial in $\X^\top\X$, $\nabla_W \MSE{W}{\X}$ is symmetric. 
$\nabla_W \MSE{\mathbf{0}_{d\times d}}{\X} = -\frac{2}{n} \X^\top \X $.

\paragraph{GOLEM-EV} The gradient of the unnormalized negative likelihood-part of the \textit{GOLEM-EV} objective denoted as $\mathcal{\widetilde{L}}_{EV}(W, \X)$ is 
\begin{align*}
\nabla_W \mathcal{\widetilde{L}}_{EV}(W,\X) &= \frac{d}{2}\nabla_W \log(n\MSE{W}{\X}) \\
&= \frac{d}{2}\frac{1}{\MSE{W}{\X}} \nabla_W \MSE{W}{\X} \\
&\propto \frac{1}{\MSE{W}{\X}}\X^\top(\X - \X W) 
\end{align*}
If $W$ is polynomial in $\X^\top\X$, $\nabla_W \mathcal{\widetilde{L}}_{EV}(W, \X)$ is symmetric.
$\nabla_W \mathcal{\widetilde{L}}_{EV}(\mathbf{0}_{d\times d}, \X) = -\frac{d}{\|\X \|_2^2} \X^\top \X $.

\paragraph{GOLEM-NV} The gradient of the unnormalized negative likelihood-part of the \textit{GOLEM-NV} objective denoted as $\mathcal{\widetilde{L}}_{NV}(W, \X)$ is
\begin{align*}
\nabla_W \mathcal{\widetilde{L}}_{NV}(W,\X) &= \frac{1}{2}\sum_{j=1}^d \nabla_W \log(n\MSE{w_j}{j}) \\
&= \left[
-\frac{1}{n\MSE{w_j}{j}} \X^\top(x_j - \X w_j)
\right]_{j=1,...,d} \\
&\propto \left[
\frac{\X^\top(x_j - \X w_j)}{\MSE{w_j}{j}}
\right]_{j=1,...,d}
\end{align*}
For the zero matrix, we have 
$\nabla_W \mathcal{\widetilde{L}}_{NV}(\mathbf{0}_{d\times d}, \X) = -\X^\top \X \operatorname{diag}\left(\|x_1\|_2^{-2}, ..., \|x_d\|_2^{-2} \right) $.

We focus on the gradients of $\operatorname{MSE}$,
$\mathcal{L}_{EV}$,
and
$\mathcal{L}_{NV}$
since l1 penalty,
acyclicity penalty $h$,
LogDet term,
and exact scaling of $\mathcal{\widetilde{L}}_{EV}$
and $\mathcal{\widetilde{L}}_{NV}$
play a subordinate role at the zero initialization,
where
the LogDet gradient has zero off-diagonals
and 
$\nabla_W h$ vanishes:

\paragraph{The LogDet in \textit{GOLEM-EV} and \textit{GOLEM-NV}}
\(
\operatorname{LogDet}(W)
=
\log(\operatorname{det}(I-W))
\)
has gradient
\[
\nabla_W \operatorname{LogDet}(W) = -(I-W)^{-\top}
\]
and vanishes when $W$ is the adjacency matrix of a DAG~\citep{ng2020role}.
If $W$ is symmetric, $\nabla_W \operatorname{LogDet}(W)$ is symmetric.
For the zero matrix, we have $\nabla_W \operatorname{LogDet}(\mathbf{0}_{d\times d}) = -I$.

\paragraph{Acyclicity Penalty/Constraint}
The function
\( h(W) =
\tr(\exp(W \odot W)) - d \)
has gradient
\( \nabla_W h(W) = \exp(W\odot W)^\top \odot 2W \).
The $h(W)\!\! = \!\!0$-level set characterizes adjacency matrices of DAGs~\citep{zheng2018dags}.
If $W$ is symmetric, $\nabla_W h(W)$ is symmetric.
For the zero matrix, we have 
$h(\mathbf{0}_{d\times d}) = 0$
and 
$\nabla_W h(\mathbf{0}_{d\times d}) = \mathbf{0}_{d\times d}$.

\subsection{Increasing Marginal and Residual Variances}\label{app:residualvariance}

We observe a strong positive correlation between the ordering by marginal variance and the ordering by residual variance after the first gradient step when minimizing a MSE- or likelihood-based objective function
via gradient descent with small step size (as in \textit{GOLEM-EV/-NV}).
For small step sizes and learning rates, marginal variance order and residual variance order are perfectly aligned for the first few optimization steps.
Here we argue for a MSE-based loss function why the residual variance follows the order of increasing marginal variance after the first optimisation step with sufficiently small step size.
Future work may investigate 
subsequent optimisation steps and the non-MSE terms of the objective functions.

Consider the data matrix $\X \in\mathbb{R}^{n\times d}$.
Without loss of generality, we assume the columns are zero-centred and ordered such that the sequence of diagonal entries $\operatorname{diag}(\X^\top\X)$ is weakly monotonically increasing.
The diagonal entries $\operatorname{diag}(\X^\top\X)$
correspond to ($n$-times) the marginal variances at step~$0$.
After the first gradient step with step size $\alpha$
in direction $-\nabla_W\MSE{\mathbf{0}_{d\times d}}{\X}=\frac{2}{n}\X^\top\X$
(cf.\ \cref{app:gradients_individual})
the vector of ($n$-times) the residual variances is
\begin{align*}
\R &= \diag{[\X-a\X\X^\top\X]^\top[\X-a\X\X^\top\X]} \\
&= \diag{\D}- 2a \diag{\D^2} + a^2 \diag{\D^3}
\end{align*}
where $\D=\X^\top\X$ and $a = \frac{2}{n}\alpha$.
For each coordinate $i$ the residual variance $\R_{i}$
is a continuous function in $a$ (and $\alpha$).
For $a=0$ and every $i\in[1,...,d-1]$
we have $\mathbf{R}_{i+1} - \mathbf{R}_i = \mathbf{D}_{i+1} - \mathbf{D}_i \geq 0$
with strict inequality if the variable pair $i,i+1$ is varsortable.
Due to continuity,
for any pair of variables with unequal marginal variances,
there exists a sufficiently small step size to ensure that the resulting residual variances follow the same order as the
marginal variances.

\subsection{Gradient Asymmetry} \label{app:gradients_golem}

We combine what we laid out in \cref{app:gradients_individual}.

\paragraph{The GOLEM-EV optimization problem is}
\[
\argmin_W \mathcal{\widetilde{L}}_{EV}(W,\X) - \operatorname{LogDet}(W) + \lambda_1\|W\|_1 + \lambda_2 h(W)
\]
with the following gradient of the objective function
\[
-\frac{d}{n\MSE{W}{\X}}\X^\top(\X - \X W) +  (I-W)^{-\top} + \lambda_1 W \oslash |W| + \lambda_2 \exp(W\odot W)^\top \odot 2W
\]
which at zero reduces to
\[
-\frac{d}{\|\X\|_2^2}\X^\top\X + I.
\]

\paragraph{The GOLEM-NV optimization problem is}
\[
\argmin_W \mathcal{\widetilde{L}}_{NV}(W,\X) - \operatorname{LogDet}(W) + \lambda_1\|W\|_1 + \lambda_2 h(W)
\]
with the following gradient of the objective function
\[
\left[ -\frac{1}{n\MSE{w_j}{j}} \X^\top(x_j - \X w_j) \right]_{j=1,...,d} +  (I-W)^{-\top} + \lambda_1 W \oslash |W| + \lambda_2 \exp(W\odot W)^\top \odot 2W
\]
which at zero reduces to
\[
-\X^\top\X \operatorname{diag}(\|x_1\|_2^{-2},...,\|x_d\|_2^{-2}) + I.
\]

The gradient in \textit{GOLEM-EV} is symmetric at $\mathbf{0}_{d\times d}$ at the first gradient descent step,
but not in general for later steps.
The gradient in \textit{GOLEM-NV} is in general not symmetric
and at $\mathbf{0}_{d\times d}$ (at the first gradient descent step)
the gradients for edges incoming into a node are inversely scaled by its marginal variance;
consequently, 
for weights $w_{i\to j}$ and $w_{j\to i}$ of opposing edges
the first gradient step is
larger magnitude for the direction with lower-variance end-note
and $w_{i\to j}$ is preferred over $w_{j\to i}$ if the variance of
$X_i$ is higher than that of $X_j$.
Under high-varsortability, the first \textit{GOLEM-NV} gradient step thus tends to favor
edges in anti-causal direction over those in causal direction.

\subsection{NOTEARS} \label{app:gradientsnotears}

The \textit{NOTEARS} optimization problem is
\( \argmin_W \frac{1}{2} \MSE{W}{\X} \text{ s.t. } h(W) = 0 \)
which is solved via the augmented Lagrangian method and dual descent~\citep{zheng2018dags}
(we omit the penalty term for the NOTEARS-l1 variant).
In the original implementation, the algorithm is initialized at $\mathbf{0}_{d\times d}$
and the diagonal of $W$ is not updated but fixed to zero
(this amounts to dual projected descent, where the adjacency matrix is projected onto the matrices with zero diagonal at each step avoiding self-loops per fiat).

The augmented Lagrangian
\[
\frac{1}{2}\MSE{W}{\X} + \frac{\rho}{2}h(W)^2  + \alpha h(W)
\]
has gradient
\[
-\frac{1}{n}\X^\top(\X - \X W)
+ (\rho h + \alpha)
\left( \exp(W\odot W)^\top \odot 2W \right)
\]
which at zero reduces to
\[
-\frac{1}{n} \X^\top \X
\]
The step size of the first gradient step in direction $\propto \X^\top\X$
is optimized by line-search to minmize the overall MSE.
As seen in the example in \cref{app:gradexample},
the residual variances may or may not follow the order of the marginal variances after this first step
due to the step size being larger than the small step size that would ensure agreement between the orders
(cf.\ \cref{app:residualvariance}).
Nonetheless,
the step size optimized by line-search aims to optimize the overall MSE
which tends to favor a better fit for edges incoming into nodes with high-marginal variance.
As a result, the first gradient step results in edge weights
that are better calibrated for edges incoming into high-marginal variance nodes 
than into low-marginal variance nodes.
In subsequent steps of the dual ascent procedure with increasing acyclicity penalty,
the reduction of overall MSE stands at odds with satisfying the DAG constraints;
it is then more costly in terms of MSE to change the weights for edges into high-marginal nodes
than into low-marginal nodes such that predominantly the edges into low-variance nodes 
tend to be removed to eventually satisfy the acyclicity constraint.
Under high varsortability, this amounts to a preference for causal edges.

\section{Standardization Is Not Enough and Regression Coefficients Tend to Increase Along the Causal Order}\label{app:standardization}

{Code to reproduce the calculations and results in this section is available at \url{https://github.com/Scriddie/Varsortability}.}

\subsection{Infinite Sample}

Here, we first discuss the three-variable case to complement the intuition provided in the main text.
Consider the following ground-truth linear additive acyclic models,
where the second model corresponds to a standardization of the first,
and the third model corresponds to a re-scaled version of the first following~\cite{mooij2020joint}:

Raw ground-truth model
\hspace{3.5em}
Standardized model
\hspace{3.5em}
Scale-harmonized model\\[-1.2em]
\begin{align*}
    A &:= N_A 
    &
    \qquad\qquad A_s &:= \nicefrac{A}{\sqrt{\var{A}}}
    &
    \qquad\qquad A_m &:= N_A \\
    B &:= \beta_{A\to B}A +N_B
    &
    B_s &:= \nicefrac{B}{\sqrt{\var{B}}}
    &
    B_m &:= \frac{\beta_{A\to B}}{\sqrt{\beta_{A\to B}^2 + 1}}A_m + N_B \\
    C &:= \beta_{B\to C}B +N_C
    &
    C_s &:= \nicefrac{C}{\sqrt{\var{C}}}
    &
    C_m &:= \frac{\beta_{B\to C}}{\sqrt{\beta_{B\to C}^2 + 1}}B_m + N_C
\end{align*}
where,
following common benchmark sampling schemes,
$N_A, N_B$, and $N_C$
are independent zero-centred noise terms that follow some distributions
with non-vanishing standard deviations $\sigma_A$, $\sigma_B$, and $\sigma_C$
sampled independently from $\operatorname{Unif}(.5, 2)$
and where
$\beta_{A\to B}$ and $\beta_{B\to C}$ are independently drawn
from $\operatorname{Unif}((-2, -.5)\cup (.5, 2))$. 
For any two nodes $X$ and $Y$, $\beta_{X \to Y}$ denotes an underlying model parameter, while $\hat{{\beta}}_{X\to Y}$
denotes the ordinary least-squares linear regression coefficient
when regressing $Y$ onto $X$ which is given as 
$\hat{\beta}_{X\to Y} = \frac{\covar{X}{Y}}{\var{X}}$.

Given observations from a variable triplet $(X,Y,Z)$,
the \emph{causal chain orientation task} is to 
infer whether the data generating causal chain is $X \to Y \to Z$,
that is, $(X,Y,Z) = (A,B,C)$
or $Z \gets Y \gets X$, that is, $(Z,X,Y) = (A,B,C)$.
While both graphs are Markov equivalent,
we can identify the correct orientation of the causal chain,
for all three considered scaling regimes,
with accuracy strictly greater than 50\%
by applying the following procedure:
\paragraph{Chain orientation rule:}
\begin{itemize}
 \item If $|\hat{\beta}_{X \to Y}| < |\hat{\beta}_{Y\to Z}|$ and $|\hat{\beta}_{Z\to Y}| > |\hat{\beta}_{Y\to X}|$,
 conclude $(X,Y,Z) = (A,B,C)$.\\[.5em]
 We conclude that $X\to Y \to Z$, if the regression coefficients are increasing in magnitude when regressing pairwise from ``left to right''.
 \item If $|\hat{\beta}_{X \to Y}| > |\hat{\beta}_{Y\to Z}|$ and $|\hat{\beta}_{Z\to Y}| < |\hat{\beta}_{Y\to X}|$,
 conclude $(X,Y,Z) = (C,B,A)$.\\[.5em]
 We conclude that $X\gets Y \gets Z$, if the regression coefficients are increasing in magnitude when regressing pairwise from ``right to left''.
 \item Otherwise, flip a coin to decide the orientation of the underlying causal chain.
\end{itemize}

For each data scale regime,
we can obtain the population regression coefficients and express those
in terms of the sampled model coefficients $\beta_{A\to B}, \beta_{B\to C}, \sigma_A, \sigma_B, \sigma_C$:
\begin{itemize}
    \item Raw ground-truth model
    \begin{itemize}
        \item ``left to right'': $\hat{\beta}_{A\to B}=\beta_{A\to B}\quad$ and $\quad\hat{\beta}_{B\to C}=\beta_{B\to C}$
        \item ``right to left'':
        $\hat{\beta}_{C\to B}=
        \frac{\beta_{B\to C} \left(\beta_{A\to B}^{2} \sigma_{A}^{2} + \sigma_{B}^{2}\right)}{\beta_{A\to B}^{2} \beta_{B\to C}^{2} \sigma_{A}^{2} + \beta_{B\to C}^{2} \sigma_{B}^{2} + \sigma_{C}^{2}}
        \quad$ 
        and 
        $\quad\hat{\beta}_{B\to A}=
        \frac{\beta_{A\to B} \sigma_{A}^{2}}{\beta_{A\to B}^{2} \sigma_{A}^{2} + \sigma_{B}^{2}}
        $
    \end{itemize}

    \item Standardized model
    \begin{itemize}
        \item ``left to right'':\\
        $\hat{\beta}_{A_s\to B_s}=
        \frac{\beta_{A\to B} \sigma_{A}^{2}}{\sqrt{\beta_{A\to B}^{2} \sigma_{A}^{2} + \sigma_{B}^{2}} \sqrt{\sigma_{A}^{2}}}
        \quad$ 
        and 
        $\quad\hat{\beta}_{B_s\to C_s}=
        \frac{\beta_{B\to C} \sqrt{\beta_{A\to B}^{2} \sigma_{A}^{2} + \sigma_{B}^{2}}}{\sqrt{\beta_{A\to B}^{2} \beta_{B\to C}^{2} \sigma_{A}^{2} + \beta_{B\to C}^{2} \sigma_{B}^{2} + \sigma_{C}^{2}}}
        $
        \item ``right to left'':\\
        $\hat{\beta}_{C_s\to B_s}=
        \frac{\beta_{B\to C} \sqrt{\beta_{A\to B}^{2} \sigma_{A}^{2} + \sigma_{B}^{2}}}{\sqrt{\beta_{A\to B}^{2} \beta_{B\to C}^{2} \sigma_{A}^{2} + \beta_{B\to C}^{2} \sigma_{B}^{2} + \sigma_{C}^{2}}}
        \quad$ 
        and 
        $\quad\hat{\beta}_{B_s\to A_s}=
        \frac{\beta_{A\to B} \sigma_{A}^{2}}{\sqrt{\beta_{A\to B}^{2} \sigma_{A}^{2} + \sigma_{B}^{2}} \sqrt{\sigma_{A}^{2}}}$
    \end{itemize}
    
    \item Scale-harmonized model
    \begin{itemize}
        \item Regression coefficients ``from left to right'':\\
        $\hat{\beta}_{A_m\to B_m}=
        \frac{\beta_{A\to B}}{\sqrt{\beta_{A\to B}^{2} + 1}}
        \quad$ 
        and 
        $\quad\hat{\beta}_{B_m\to C_m}=
        \frac{\beta_{B\to C}}{\sqrt{\beta_{B\to C}^{2} + 1}}
        $
        \item Regression coefficients ``from right to left'':\\
        $\hat{\beta}_{C_m\to B_m}=
        \frac{\beta_{B\to C} \left(\beta_{B\to C}^{2} + 1\right)^{1.5} \left(\beta_{A\to B}^{2} \sigma_{A}^{2} + \sigma_{B}^{2} \left(\beta_{A\to B}^{2} + 1\right)\right)}{\beta_{A\to B}^{2} \beta_{B\to C}^{2} \sigma_{A}^{2} \left(\beta_{B\to C}^{2} + 1\right) + \beta_{B\to C}^{2} \sigma_{B}^{2} \left(\beta_{A\to B}^{2} + 1\right) \left(\beta_{B\to C}^{2} + 1\right) + \sigma_{C}^{2} \left(\beta_{A\to B}^{2} + 1\right) \left(\beta_{B\to C}^{2} + 1\right)^{2}}
        \quad$ and 
        $\quad\hat{\beta}_{B_m\to A_m}=
        \frac{\beta_{A\to B} \sigma_{A}^{2} \sqrt{\beta_{A\to B}^{2} + 1}}{\beta_{A\to B}^{2} \sigma_{A}^{2} + \sigma_{B}^{2} \left(\beta_{A\to B}^{2} + 1\right)}
        $
    \end{itemize}
\end{itemize}

We obtain the following probabilities by Monte Carlo approximation, resampling the $5$ model parameters $100,000$ times:
\begin{table}[H]
\small
\begin{longtable}{llr}
    \caption{Chain orientation results in the population limit.}\\
    \label{tab:chain_pop_tab}
    \small
    \textbf{Weight distribution} & \textbf{Chain orientation rule cases} \\
    \midrule
    $\operatorname{Unif}((-2, .5) \cup (.5, 2))$ 
    & $P\left[|\hat{\beta}_{A\to B}| < |\hat{\beta}_{B\to C}| \text{ and } |\hat{\beta}_{C\to B}| > |\hat{\beta}_{B\to A}|\right]$
    & 29.376\%
    \\
    & $P\left[|\hat{\beta}_{A\to B}| > |\hat{\beta}_{B\to C}| \text{ and } |\hat{\beta}_{C\to B}| < |\hat{\beta}_{B\to A}|\right]$
    & 5.486\%
    \\
    & $P\left[\textbf{``orientation rule correct on raw data''}\right]$
    & \textbf{61.945\%}
    \\
    \midrule

    & $P\left[|\hat{\beta}_{A_s\to B_s}| < |\hat{\beta}_{B_s\to C_s}| \text{ and } |\hat{\beta}_{C_s\to B_s}| > |\hat{\beta}_{B_s\to A_s}|\right]$
    & 73.181\%
    \\
    & $P\left[|\hat{\beta}_{A_s\to B_s}| > |\hat{\beta}_{B_s\to C_s}| \text{ and } |\hat{\beta}_{C_s\to B_s}| < |\hat{\beta}_{B_s\to A_s}|\right]$
    & 26.819\%
    \\
    & $P\left[\textbf{``orientation rule correct on standardized data''}\right]$
    & \textbf{73.181\%}
    \\
    \midrule

    & $P\left[|\hat{\beta}_{A_m\to B_m}| < |\hat{\beta}_{B_m\to C_m}| \text{ and } |\hat{\beta}_{C_m\to B_m}| > |\hat{\beta}_{B_m\to A_m}|\right]$
    & 31.631\%
    \\
    & $P\left[|\hat{\beta}_{A_m\to B_m}| > |\hat{\beta}_{B_m\to C_m}| \text{ and } |\hat{\beta}_{C_m\to B_m}| < |\hat{\beta}_{B_m\to A_m}|\right]$
    & 17.318\%
    \\
    & $P\left[\textbf{``orientation rule correct on scale-harmonized data''}\right]$
    & \textbf{57.1565\%}
    \\
    \midrule
    \midrule

    $\operatorname{Unif}((-.9, -.5) \cup (.5, .9))$ 
    & $P\left[|\hat{\beta}_{A\to B}| < |\hat{\beta}_{B\to C}| \text{ and } |\hat{\beta}_{C\to B}| > |\hat{\beta}_{B\to A}|\right]$
    & 31.033\%
    \\
    & $P\left[|\hat{\beta}_{A\to B}| > |\hat{\beta}_{B\to C}| \text{ and } |\hat{\beta}_{C\to B}| < |\hat{\beta}_{B\to A}|\right]$
    & 18.124\%
    \\
    & $P\left[\textbf{``orientation rule correct on raw data''}\right]$
    & \textbf{56.454\%}
    \\
    \midrule

    & $P\left[|\hat{\beta}_{A_s\to B_s}| < |\hat{\beta}_{B_s\to C_s}| \text{ and } |\hat{\beta}_{C_s\to B_s}| > |\hat{\beta}_{B_s\to A_s}|\right]$
    & 62.231\%
    \\
    & $P\left[|\hat{\beta}_{A_s\to B_s}| > |\hat{\beta}_{B_s\to C_s}| \text{ and } |\hat{\beta}_{C_s\to B_s}| < |\hat{\beta}_{B_s\to A_s}|\right]$
    & 37.769\%
    \\
    & $P\left[\textbf{``orientation rule correct on standardized data''}\right]$
    & \textbf{62.231\%}
    \\
    \midrule

    & $P\left[|\hat{\beta}_{A_m\to B_m}| < |\hat{\beta}_{B_m\to C_m}| \text{ and } |\hat{\beta}_{C_m\to B_m}| > |\hat{\beta}_{B_m\to A_m}|\right]$
    & 30.025\%
    \\
    & $P\left[|\hat{\beta}_{A_m\to B_m}| > |\hat{\beta}_{B_m\to C_m}| \text{ and } |\hat{\beta}_{C_m\to B_m}| < |\hat{\beta}_{B_m\to A_m}|\right]$
    & 20.607\%
    \\
    & $P\left[\textbf{``orientation rule correct on scale-harmonized data''}\right]$
    & \textbf{54.709\%}
    \\
    \midrule
    \midrule

    $\operatorname{Unif}((-.9, -.1) \cup (.1, .9))$ 
    & $P\left[|\hat{\beta}_{A\to B}| < |\hat{\beta}_{B\to C}| \text{ and } |\hat{\beta}_{C\to B}| > |\hat{\beta}_{B\to A}|\right]$
    & 32.480\%
    \\
    & $P\left[|\hat{\beta}_{A\to B}| > |\hat{\beta}_{B\to C}| \text{ and } |\hat{\beta}_{C\to B}| < |\hat{\beta}_{B\to A}|\right]$
    & 24.012\%
    \\
    & $P\left[\textbf{``orientation rule correct on raw data''}\right]$
    & \textbf{54.234\%}
    \\
    \midrule

    & $P\left[|\hat{\beta}_{A_s\to B_s}| < |\hat{\beta}_{B_s\to C_s}| \text{ and } |\hat{\beta}_{C_s\to B_s}| > |\hat{\beta}_{B_s\to A_s}|\right]$
    & 55.790\%
    \\
    & $P\left[|\hat{\beta}_{A_s\to B_s}| > |\hat{\beta}_{B_s\to C_s}| \text{ and } |\hat{\beta}_{C_s\to B_s}| < |\hat{\beta}_{B_s\to A_s}|\right]$
    & 44.210\%
    \\
    & $P\left[\textbf{``orientation rule correct on standardized data''}\right]$
    & \textbf{55.790\%}
    \\
    \midrule

    & $P\left[|\hat{\beta}_{A_m\to B_m}| < |\hat{\beta}_{B_m\to C_m}| \text{ and } |\hat{\beta}_{C_m\to B_m}| > |\hat{\beta}_{B_m\to A_m}|\right]$
    & 31.867\%
    \\
    & $P\left[|\hat{\beta}_{A_m\to B_m}| > |\hat{\beta}_{B_m\to C_m}| \text{ and } |\hat{\beta}_{C_m\to B_m}| < |\hat{\beta}_{B_m\to A_m}|\right]$
    & 25.136\%
    \\
    & $P\left[\textbf{``orientation rule correct on scale-harmonized data''}\right]$
    & \textbf{53.3655\%}
\end{longtable}
\end{table}
We draw edge weights independently from the uniform distribution indicated in the first column of \cref{tab:chain_pop_tab} and
noise standard-deviations $\sigma_A, \sigma_B, \sigma_C$ are
drawn independently from $\operatorname{Unif}(.5,2)$
in all cases.
A 99\% confidence interval for the orientation accuracy
under random guessing is $(49.593\%, 50.407\%)$.
The orientation rule achieves above chance accuracy
in all regimes.

\subsection{Finite Sample}
Given observations from $(X_1,...,X_d)$
generated by a linear ANM with either
$X_1\to X_2 \to ...\to X_d$
or
$X_d \to X_{d-1} \to ... \to X_1$,
we can decide the directionality 
by identifying the direction in which the absolute values of the regression coefficients
tend to increase.
More precisely,
we compare the sequences of absolute regression coefficients\\
\hspace*{3em}
\emph{``left-to-right regression coefficients''}
$|\hat{\beta}_{X_1\to X_2}|, ..., |\hat{\beta}_{X_{d-1}\to X_d}|$\\
to\\
\hspace*{3em}
\emph{``right-to-left regression coefficients''}
$|\hat{\beta}_{X_d\to X_{d-1}}|, ..., |\hat{\beta}_{X_{2}\to X_1}|.$\\[.5em]
We infer $X_1\to ... \to X_d$ if the former is in better agreement with an
ascending sorting than the latter and infer $X_d \to ... \to X_1$
otherwise.

In the main text, we discussed the case for standardized data where the regression coefficients for any two nodes $X_i$ and $X_j$ are given as $|\cor(X_i, X_j)|$.
We expect the sequence of absolute regression coefficients
to increase along the causal order because 
the correlation between consecutive nodes tends to be higher further downstream
as parent nodes contribute more to a nodes marginal variance relative to its noise term.

On the raw data scale, the sequences of regression coefficients are\\
\hspace*{.5em}
\emph{``left-to-right''}
$
\frac{\sqrt{\var{X_2}}}{\sqrt{\var{X_1}}}
|\cor(X_1, X_2)|, ...,
\frac{\sqrt{\var{X_{d}}}}{\sqrt{\var{X_{d-1}}}}
|\cor(X_{d-1}, X_d)|\quad$ and\\
\hspace*{.5em}
\emph{``right-to-left''}
$
\frac{\sqrt{\var{X_{d-1}}}}{\sqrt{\var{X_{d}}}}
|\cor(X_{d-1}, X_d)|, ...,
\frac{\sqrt{\var{X_1}}}{\sqrt{\var{X_2}}}
|\cor(X_1, X_2)|.
$\\[.5em]

On both raw and standardized data, we find that the direction in which absolute regression coefficients tend to increase most corresponds to the causal direction in more than 50\% of cases.
To quantify ``increasingness'' of sequences of absolute regression coefficients
we count the number of correctly ordered pairs of regression coefficients,
that is, how often a regression coefficient is smaller in magnitude
than regression coefficients later in the sequence
and substract the number of discordant pairs.
The decision rule then predicts the direction in which the sequence of regression
coefficients is more increasing according to this criterion.

We apply this orientation rule to simulated data (sample size $1000$)
for varying chain lengths and edge distributions,
and when applied to raw observational data, standardized observational data,
and data when the parameters were scale-harmonized as per \cite{mooij2020joint}.
The table below establishes,
that for iid distributed parameters of the underlying data generating process,
the orientation of a causal chain can be identified with probability strictly greater than 50\%.

\begin{table}[h]
    \caption{Empirical Chain Orientation Results}
    \small
    \begin{tabular}{cc|ccc|ccc}
    & & \multicolumn{3}{c}{accuracy by variance-sorting} & \multicolumn{3}{c}{accuracy by coefficient-sorting} \\
    d & edge range & raw & standardized & harmonized & raw & standardized & harmonized \\
    \midrule
    3 & $\pm (0.5, 2.0)$ &97.50\% & {\color{gray}50.05\%} & 84.70\% & 62.58\% & 73.03\% & 57.30\% \\
    & $\pm (0.5, 0.9)$ &80.38\% & {\color{gray}50.05\%} & 69.62\% & 57.15\% & 62.38\% & 55.65\% \\
    & $\pm (0.1, 0.9)$ &65.65\% & {\color{gray}50.30\%} & 60.08\% & 54.17\% & 55.88\% & 53.45\% \\
    \midrule
    5 & $\pm (0.5, 2.0)$ &98.67\% & {\color{gray}50.15\%} & 82.17\% & 78.60\% & 86.58\% & 64.20\% \\
    & $\pm (0.5, 0.9)$ &77.65\% & {\color{gray}49.27\%} & 66.30\% & 61.83\% & 68.65\% & 57.50\% \\
    & $\pm (0.1, 0.9)$ &63.08\% & {\color{gray}50.38\%} & 57.65\% & 58.17\% & 57.33\% & 56.35\% \\
    \midrule
    10 & $\pm (0.5, 2.0)$ &99.38\% & {\color{gray}50.02\%} & 79.30\% & 93.72\% & 96.97\% & 69.08\% \\
    & $\pm (0.5, 0.9)$ &73.75\% & {\color{gray}50.25\%} & 62.00\% & 64.97\% & 70.70\% & 58.50\% \\
    & $\pm (0.1, 0.9)$ &62.55\% & {\color{gray}51.23\%} & 58.25\% & 55.85\% & 56.05\% & 54.40\% \\
    \end{tabular}
    \label{tab:chain_finite}
\end{table}

A 99\% confidence interval for the orientation accuracy
under random guessing is $(47.975\%, 52.025\%)$ 
($1000$ repetitions for each of the four noise types).
Thus, variance-sorting on the standardized data is
the only setting in which no above-chance orientation accuracy is achieved.
This is expected, as variance sorting amounts to a random sorting once nodes are standardized.

\section{Empirical Evaluation of Varsortability} \label{appendix:varsortability}
We empirically estimate expected varsortability for our experimental set-up and a non-linear version of our experimental set-up by calculating the fraction of directed paths that are correctly sorted by marginal variance
in the randomly sampled ANMs.

\subsection{Varsortability in Linear Additive Noise Models} \label{appendix:varsortability_linear}
Consistent with our theoretical results,
varsortability is close to 1
across all graph and noise types in our experimental set-up,
cf.~\cref{tab:varsortability_linear}.
Varsortability is higher in denser than in sparser graphs.

\begin{table}[H]
    \small
    \caption{Empirical varsortability in our experimental linear ANM set-up. Average varsortability is high in all settings. Our parameter choices are common in the literature. We sample 1000 observations of ten 50-node graphs for each combination of graph and noise type.}
    \label{tab:varsortability_linear}
    \centering
    \begin{tabular}{llrrr}
        \toprule
             &        & \multicolumn{3}{l}{varsortability} \\
             &        &            min & mean &  max \\
        graph & noise &                &      &      \\
        \midrule
        ER-1 & Gauss-EV &           0.94 & 0.97 & 0.99 \\
             & exponential &           0.94 & 0.97 & 0.99 \\
             & gumbel &           0.94 & 0.97 & 1.00 \\
        ER-2 & Gauss-EV &           0.97 & 0.99 & 1.00 \\
             & exponential &           0.97 & 0.99 & 1.00 \\
             & gumbel &           0.98 & 0.99 & 0.99 \\
        ER-4 & Gauss-EV &           0.98 & 0.99 & 0.99 \\
             & exponential &           0.98 & 0.99 & 0.99 \\
             & gumbel &           0.98 & 0.99 & 0.99 \\
        SF-4 & Gauss-EV &           0.98 & 1.00 & 1.00 \\
             & exponential &           0.98 & 1.00 & 1.00 \\
             & gumbel &           0.98 & 1.00 & 1.00 \\
        \bottomrule
    \end{tabular}
\end{table}

\subsection{Varsortability in Non-Linear Additive Noise Models}\label{appendix:varsortability_nonlinear}

\cref{tab:varsortability_nonlinear} shows
varsortabilities for a non-linear version of our experimental set-up as used by \cite{zheng2020learning}. While the fluctuations in \cref{tab:varsortability_nonlinear} are greater
than in \cref{tab:varsortability_linear}, all settings exhibit high varsortability on average. 
Our findings indicate that varsortability is a concern for linear and non-linear ANMs.

\begin{table}[H]
    \small
    \caption{Empirical varsortability in non-linear ANM. Average varsortability is high in all settings. Our parameter choices are common in the literature. We sample 1000 observations of ten 20-node graphs for each combination of graph and ANM-type.}
    \label{tab:varsortability_nonlinear}
    \centering 
    \begin{tabular}{llrrr}
        \toprule
                &                   & \multicolumn{3}{l}{varsortability} \\
                &                   &            min & mean &  max \\
        graph & ANM-type &                &      &      \\
        \midrule
        ER-1 & Additive GP &           0.81 & 0.91 & 1.00 \\
                & GP &           0.72 & 0.86 & 0.96 \\
                & MLP &           0.55 & 0.79 & 0.96 \\
                & Multi Index Model &           0.62 & 0.82 & 1.00 \\
        ER-2 & Additive GP &           0.79 & 0.91 & 0.98 \\
                & GP &           0.82 & 0.89 & 0.97 \\
                & MLP &           0.46 & 0.71 & 0.87 \\
                & Multi Index Model &           0.65 & 0.79 & 0.89 \\
        ER-4 & Additive GP &           0.90 & 0.95 & 0.98 \\
                & GP &           0.74 & 0.88 & 0.93 \\
                & MLP &           0.59 & 0.72 & 0.85 \\
                & Multi Index Model &           0.57 & 0.73 & 0.85 \\
        SF-4 & Additive GP &           0.95 & 0.97 & 0.99 \\
                & GP &           0.88 & 0.94 & 0.97 \\
                & MLP &           0.75 & 0.83 & 0.93 \\
                & Multi Index Model &           0.77 & 0.84 & 0.97 \\
        \bottomrule
    \end{tabular}
\end{table}

\subsection{Causal Order and Marginal Variance} \label{app:causal_order_vsb}
We observe strong empirical evidence in \cref{fig:causal_order_vsb} that marginal variance tends to increase quickly along the causal order, even if the settings are not guaranteed to yield high expected varsortability between a pair of root cause and effect (for example, if all edges are chosen in a small-magnitude range).
This indicates that high levels of varsortability can scarcely be avoided on larger graphs.
\begin{figure}[H]
    \centering
    \includegraphicsmaybe[scale=0.5]{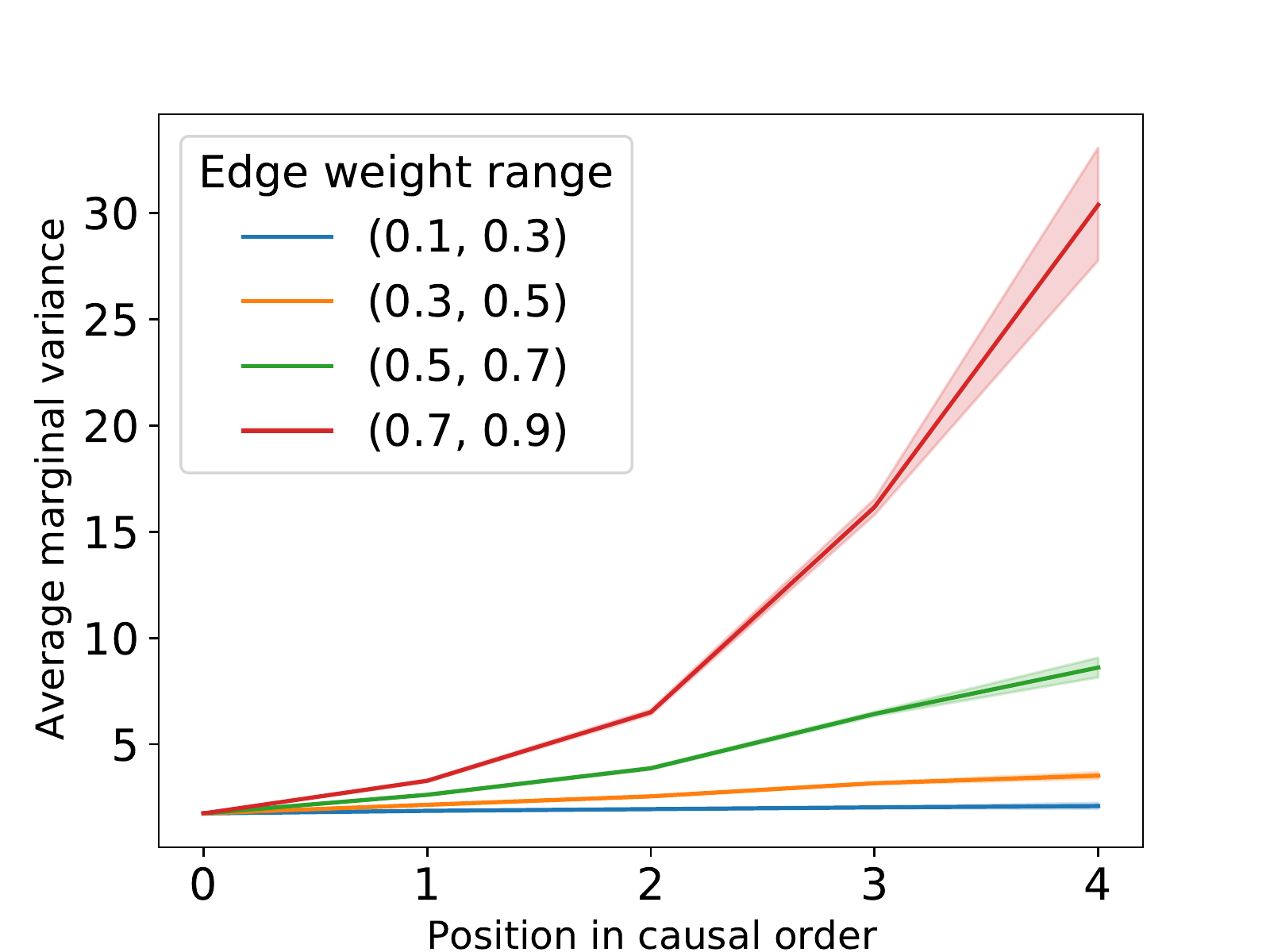}
    \caption{Average marginal variance along the causal order for $1000$ observations of $1000$ simulated $30$-node ER-2 graphs with Gaussian noise standard deviations sampled uniformly in $(0.5, 2)$ for each edge weight range.
    Edge weights are drawn independently and uniformly from the union of negative and positive of the indicated edge range, that is, for example, the edge weights for the red curve are drawn from $\operatorname{Unif}((-.9,-.7)\cup(.7, .9))$.}
    \label{fig:causal_order_vsb}
\end{figure}

\subsection{Varsortability Algorithm}\label{algo:varsortability}

{The implementation is also available at \url{https://github.com/Scriddie/Varsortability}.}

\begin{lstlisting}[language=python, basicstyle=\footnotesize]
import numpy as np

def varsortability(X, W, tol=1e-9):
    """ Takes n x d data and a d x d adjaceny matrix,
    where the i,j-th entry corresponds to the edge weight for i->j,
    and returns a value indicating how well the variance order
    reflects the causal order. """
    E = W != 0
    Ek = E.copy()
    var = np.var(X, axis=0, keepdims=True)

    n_paths = 0
    n_correctly_ordered_paths = 0

    for _ in range(E.shape[0] - 1):
        n_paths += Ek.sum()
        n_correctly_ordered_paths += (Ek * var / var.T > 1 + tol).sum()
        n_correctly_ordered_paths += 1/2*(
            (Ek * var / var.T <= 1 + tol) *
            (Ek * var / var.T >  1 - tol)).sum()
        Ek = Ek.dot(E)

    return n_correctly_ordered_paths / n_paths

if __name__ == "__main__":
    W = np.array([[0, 1, 0], [0, 0, 2], [0, 0, 0]])
    X = np.random.randn(1000, 3).dot(np.linalg.inv(np.eye(3) - W))
    print("Varsortability:", varsortability(X, W))
    
    X_std = (X - np.mean(X, axis=0))/np.std(X, axis=0)
    print("Varsortability standardized:", varsortability(X_std, W))
\end{lstlisting}

\section{\textit{sortnregress}: A Diagnostic Tool to Reveal Varsortability}

In \cref{sec:baselines} we introduce \textit{sortnregress} as a simple baseline method.
In the following subsections,
we provide Python code that implements \textit{sortnregress} thereby establishing its ease
and illustrate how its DAG recovery performance reflects varying degrees of varsortability.

\subsection{Implementation of Sortnregress} \label{sec:sortnregress_code}

{The implementation is also available at \url{https://github.com/Scriddie/Varsortability}.}

\begin{lstlisting}[language=python, basicstyle=\footnotesize]
import numpy as np
from sklearn.linear_model import LinearRegression, LassoLarsIC

def sortnregress(X):
    """ Take n x d data, order nodes by marginal variance and 
    regresses each node onto those with lower variance, using
    edge coefficients as structure estimates. """
    LR = LinearRegression()
    LL = LassoLarsIC(criterion='bic')

    d = X.shape[1]
    W = np.zeros((d, d))
    increasing = np.argsort(np.var(X, axis=0))

    for k in range(1, d):
        covariates = increasing[:k]
        target = increasing[k]

        LR.fit(X[:, covariates], X[:, target].ravel())
        weight = np.abs(LR.coef_)
        LL.fit(X[:, covariates] * weight, X[:, target].ravel())
        W[covariates, target] = LL.coef_ * weight

    return W
\end{lstlisting}

\subsection{Varsortabiltiy and Score Attainable by Variance Ordering} \label{app:varsortability_sortnregress}
In \cref{fig:vsb_performance} we observe that \textit{sortnregress} improves linearly with varsortability. For a varsortability of 0.93 as in our experimental settings (cf.\ \cref{sec:bench}), it recovers the structure near-perfectly. \textit{randomregress} uses a random ordering but is otherwise identical to \textit{sortnregress}.
The different ranges of varsortability can be classified as follows (n=30):
\begin{itemize}
    \item < 0.33: \textit{sortnregress} performs significantly worse than \textit{randomregress} (p<1e-4)
    \item 0.33–0.66: no significant difference between \textit{sortnregress} and \textit{randomregress} (p=0.40)
    \item > 0.66: \textit{sortnregress} performs significantly better than \textit{randomregress} (p<1e-4)
\end{itemize}
\begin{figure}[H]
    \centering
    \begin{subfigure}{.49\linewidth}
        \centering
        \includegraphicsmaybe[width=\linewidth]
        {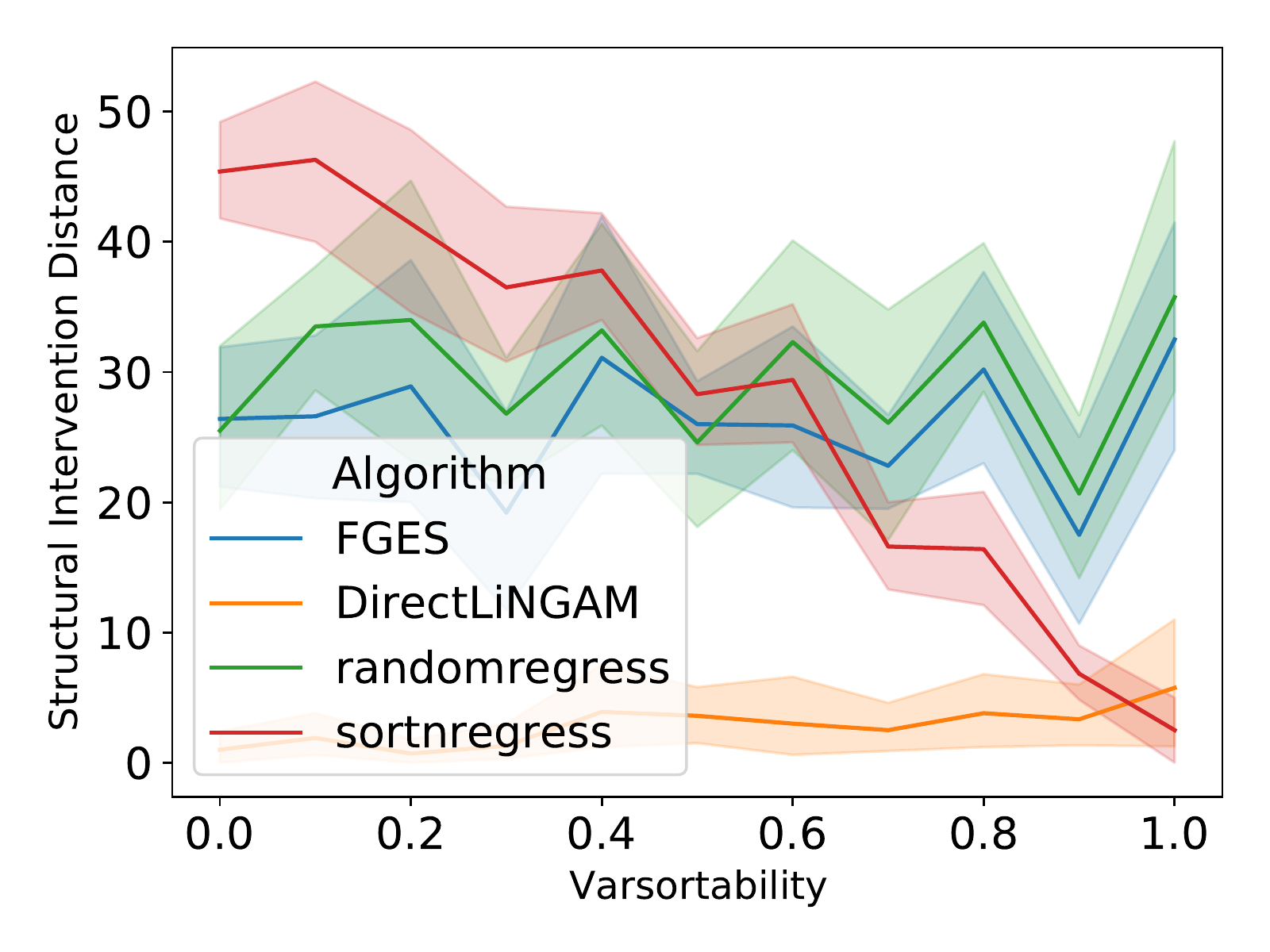}
    \end{subfigure}
    \hfil
    \begin{subfigure}{.49\linewidth}
        \centering
        \includegraphicsmaybe[width=\linewidth]
        {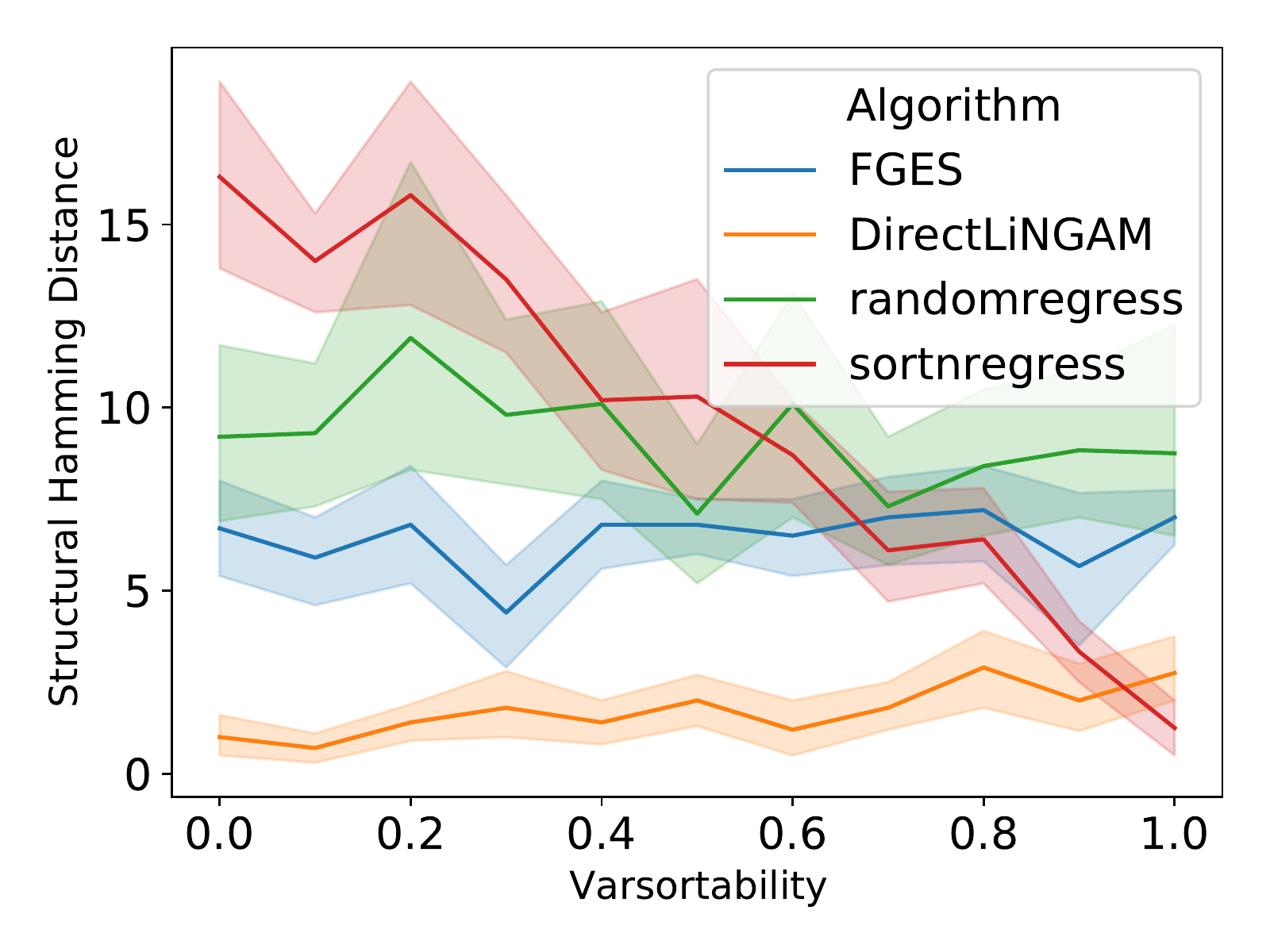}
    \end{subfigure}
    \caption{Relationship between varsortability and score attainable through ordering by variance. Results shown for 10 simulated 10-node ER-1 graphs in each of 10 equally spaced varsortability bins. Note that for standard simulation settings most models have high varsortability.
    We use edge weights in $(-0.5, -0.1)\cup(0.1, 0.5)$, Gumbel noise with standard deviations in $(0.5, 2)$, and still need to discard many models with high varsortability to
    obtain $10$ instances per varsortability bin.}
    \label{fig:vsb_performance}
\end{figure}

\section{Evaluation on Real-World Data}\label{sec:realworld}\label{app:sachs}
We analyze a dataset on protein signaling networks obtained by \cite{sachs2005causal}. 
We evaluate our algorithms on ten bootstrap samples of the observational part of the dataset consisting of 853 observations, 11 nodes, and 17 edges.
Our results show that there is no dominating algorithm. 
On average, most algorithms achieve performances similar to those of \textit{randomregress} or the empty graph.
Note that the results in terms of SHD are susceptible to thresholding choices and the empty graph baseline outperforms a majority of the algorithms.
Our results are in line with previous reports \citep{lachapelle2019gradient,ng2020role}.
We observe scale-sensitivity of the continuous learning algorithms and \textit{sortnregress}. However, in contrast to our simulation study in \cref{sec:simulations}, the effect is small and inconsistent. The results do not show the patterns observed under high varsortability, which is consistent with the measured mean varsortability of $0.57$ with a standard deviation of $0.01$ across our bootstrapped samples.

\begin{figure}[H]
\begin{minipage}[b]{0.5\textwidth}
    \centering
    \includegraphicsmaybe[width=\linewidth]  %
    {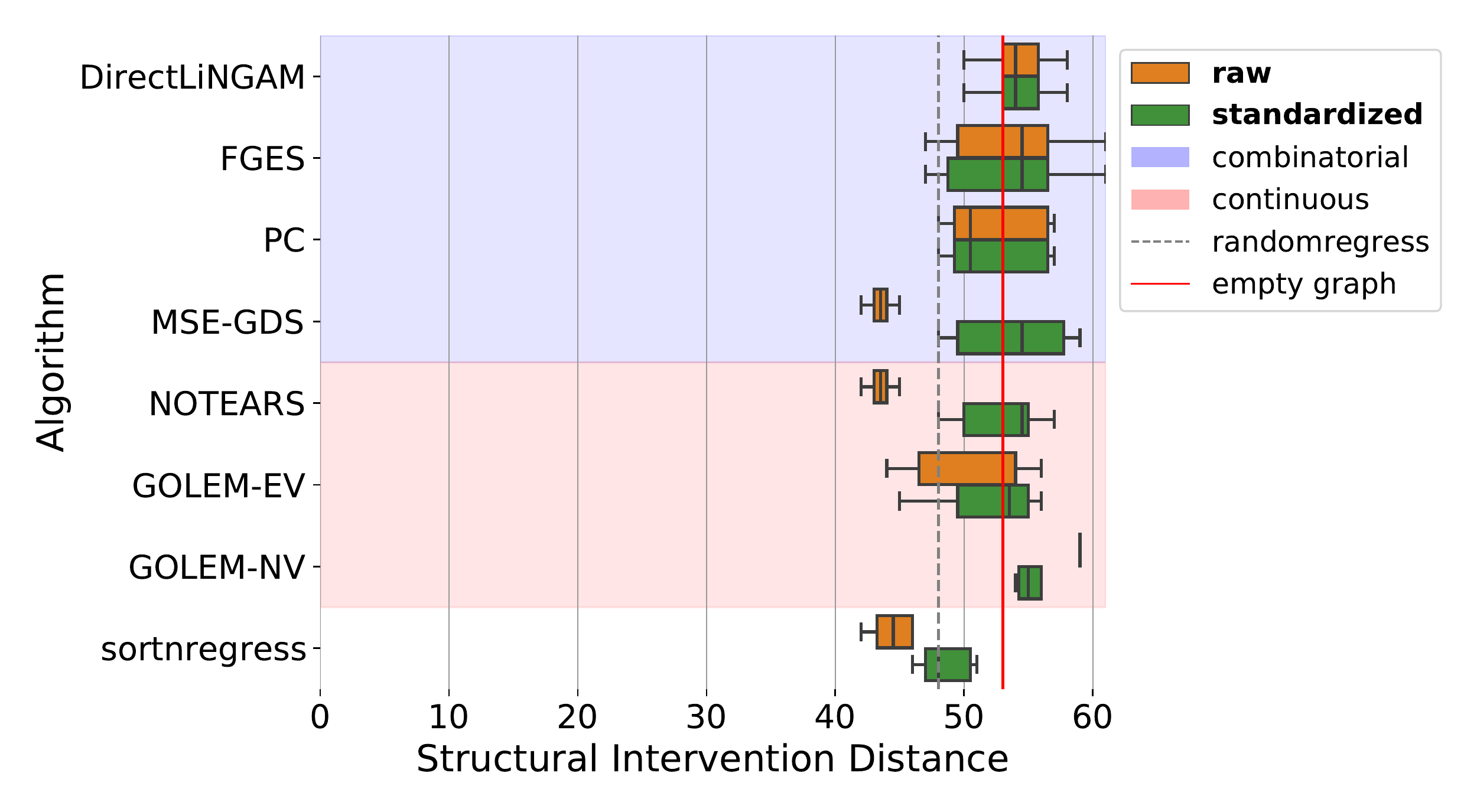}
\end{minipage}
\begin{minipage}[b]{0.5\textwidth}
    \centering
    \includegraphicsmaybe[width=\linewidth]  %
    {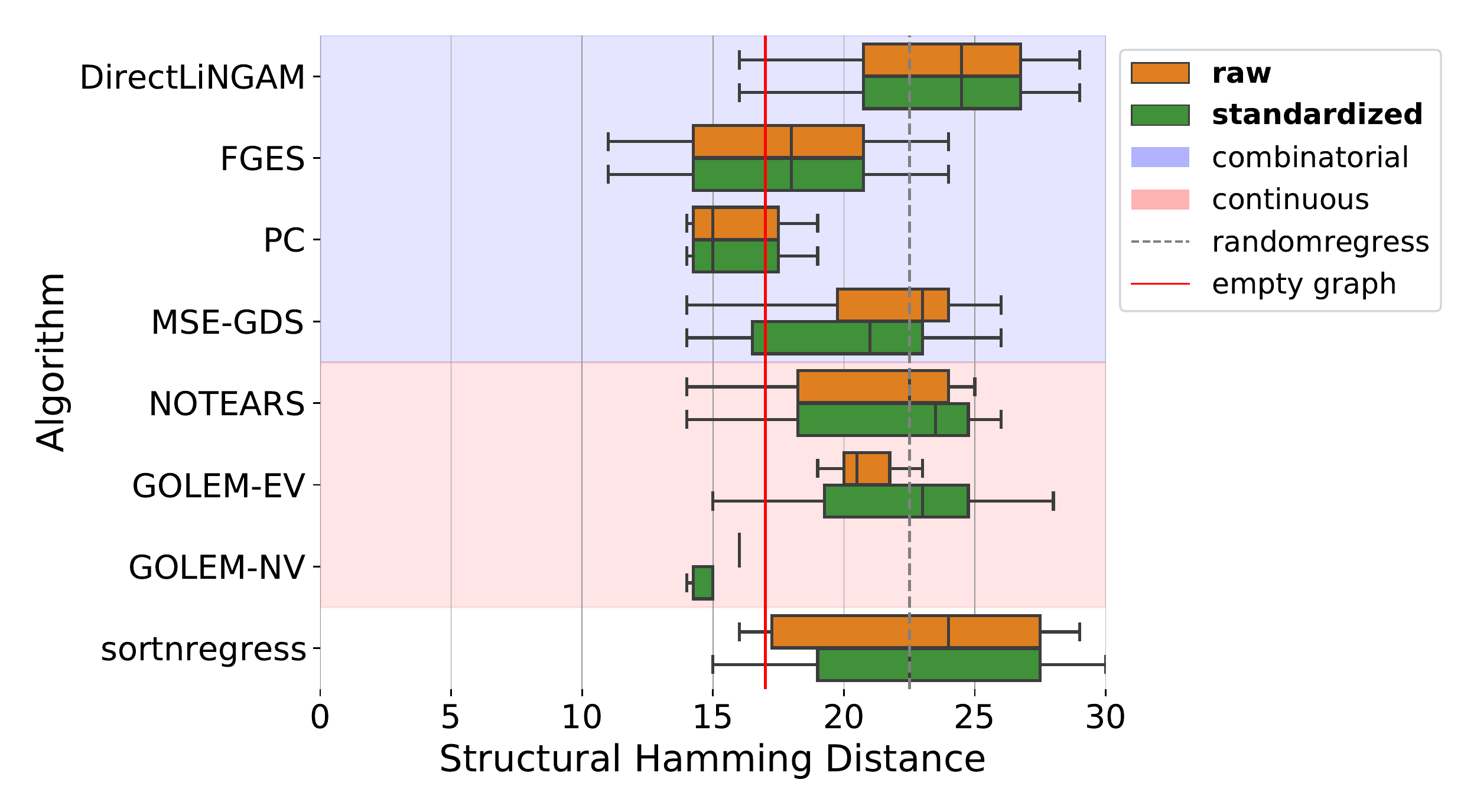}
\end{minipage}
    \caption{SID (left) and SHD (right) performance of combinatorial and continuous methods on real-world data.
    }
    \label{fig:sachs_sid}
    \label{fig:sachs_shd}
\end{figure}

\section{Model Selection in Continuous Optimization}\label{sec:landscape}

We illustrate the optimization landscape for the Gaussian MLE under Gaussian noise. This corresponds to the loss of \textit{GOLEM-NV} as stated in \cref*{app:algorithms} with a sparsity penalty of zero. We compare vanilla MLE to MLE with Lasso regularization for raw and standardized data. In \cref{fig:loss_landscape} we show the loss landscape in terms of SID and SHD difference to the true structure and highlight global optima. In the case of tied scores between the true structure and an alternative structure we select the true structure.
For MLE with Lasso regularization using a penalty of $0.1$,
the optimal loss is achieved by the true structure more frequently under standardization
(red dots accumulate in the bottom left corner).
Our result indicates that the Lasso sparsity penalty is influenced by the data scale
and is better calibrated on standardized data.
It is not unexpected that penalization is scale dependent,
a problem that is, for example, discussed in applications of Ridge regression.
\begin{figure}[!htp]
    \centering
    \includegraphicsmaybe[width=.8\textwidth]
    {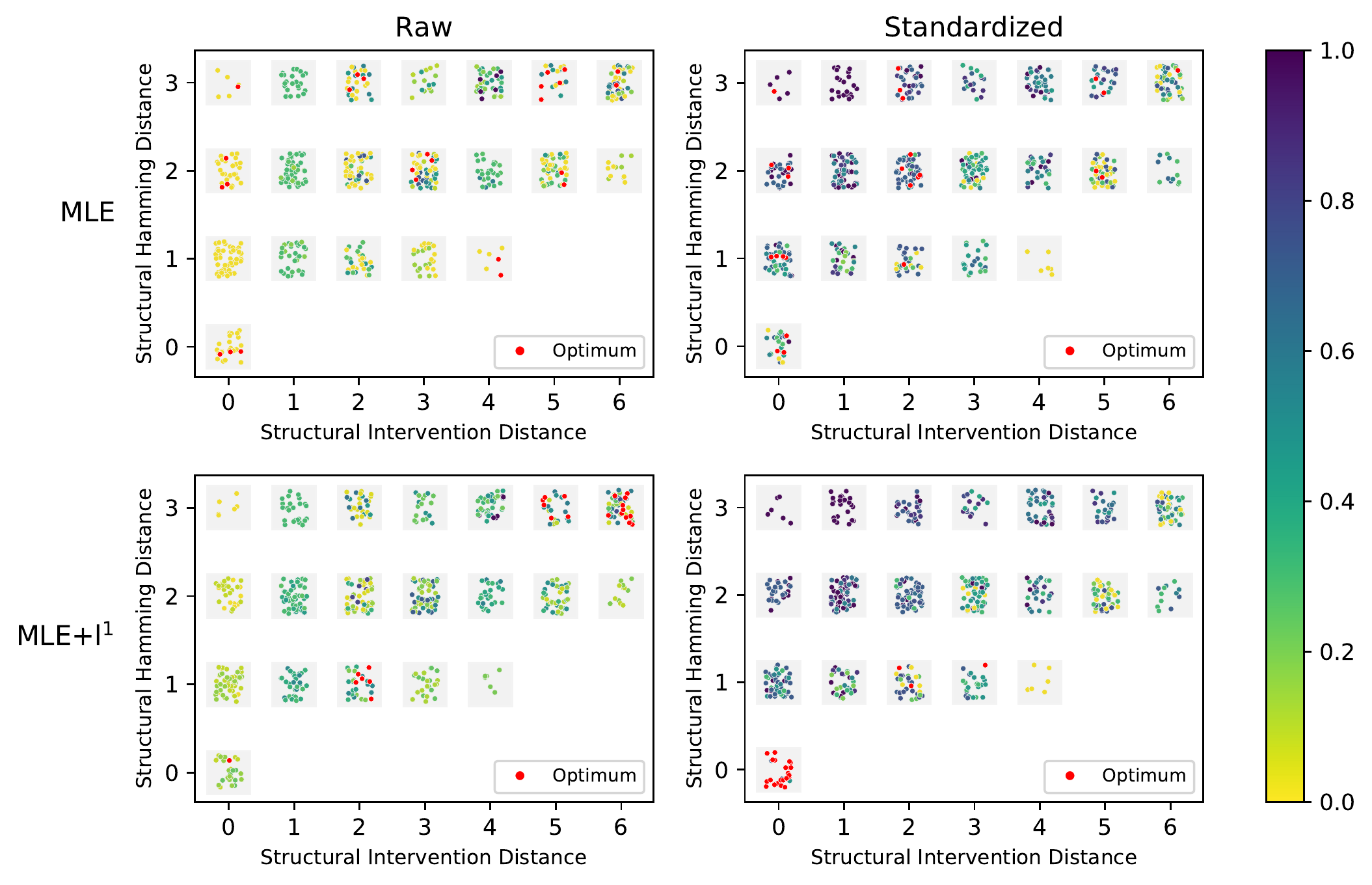}
    \caption{Standardized loss landscape for all 25 candidate graphs relative to each of the 25 possible 3-node ground-truth structures (a total of $25\times 25$ candidate-true graph pairs).
    The loss is scaled to $[0,1]$, see colorbar.}
    \label{fig:loss_landscape}
\end{figure}

\section{Detailed Results} \label{appendix:detailed_results}

We provide a comprehensive overview over our empirical DAG/MEC recovery results for different evaluation metrics, graph types, and graph sizes.

\subsection{MEC Recovery}
An analysis of MEC recovery allows us to distinguish whether any drops in performance are within the expectations of identifiability. 
We evaluate the discovery of the MEC of the ground-truth DAG in a Gaussian setting with non-equal noise variances where only the ground-truth MEC but not the ground-truth DAG are identifiable. 
Since evaluating the SID between Markov equivalence classes is computationally expensive and prohibitively so for large graphs, we restrict ourselves to the setting here.
When comparing MEC, we choose the upper limit of SID differences in \cref{fig:main_norm_sid} in the main text.
In \cref{fig:MEC_SID_lower} we show that the relative performances are similar for the lower SID limit. 
\begin{figure}[H]
    \centering
    \includegraphicsmaybe[width=3.5in]{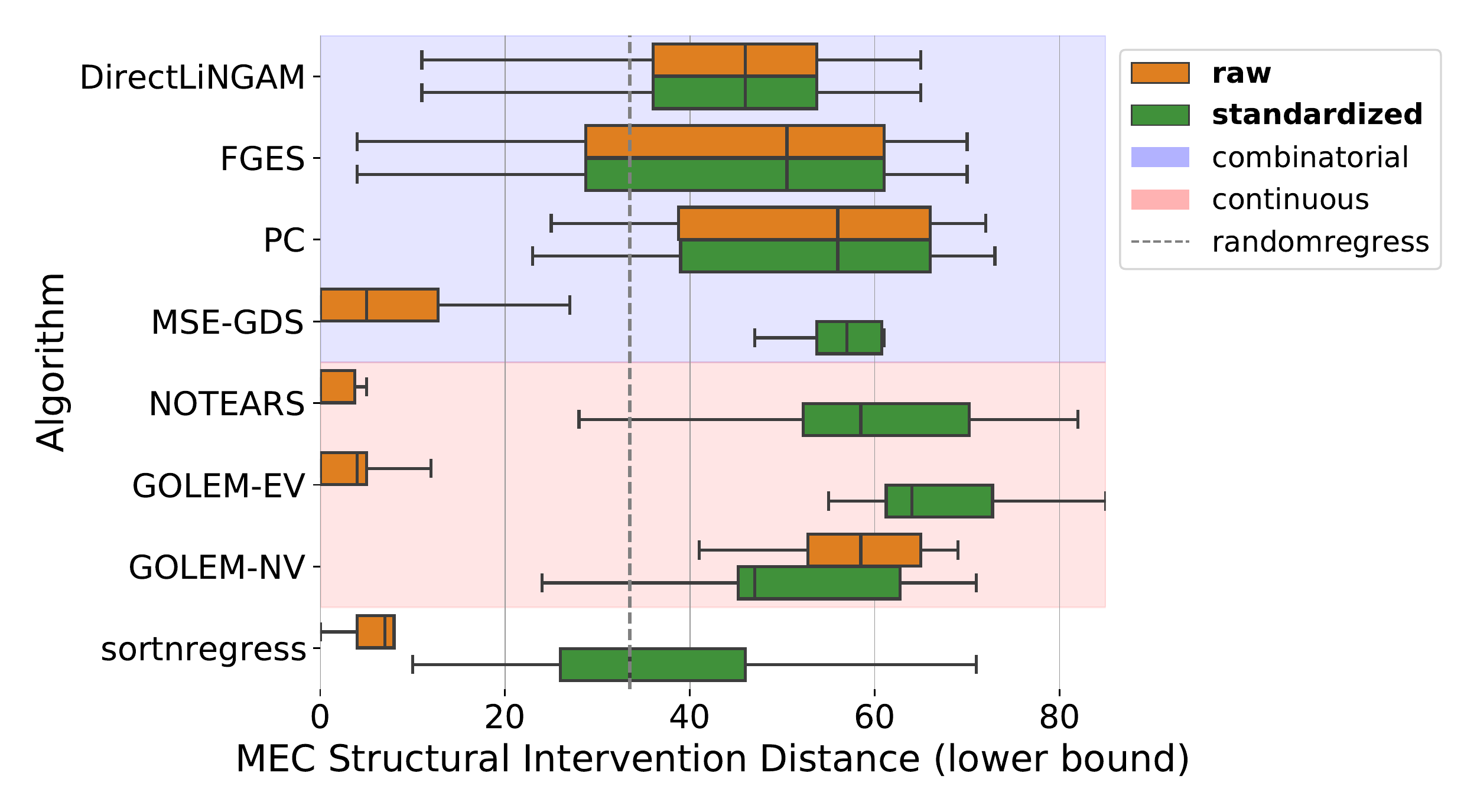}
    \caption{Lower bound of SID in MEC recovery for 10 node ER-2 graphs with non-equal Gaussian noise.}
    \label{fig:MEC_SID_lower}
\end{figure} 
We conclude that the drop in performance extends from the recovery of the DAG to the recovery of the MEC and therefore goes beyond the difficulty of identifying the correct DAG within a MEC.

\subsection{Results Across Thresholding Regimes} \label{app:thresholding}

To ensure the effects we observe constitute a general phenomenon, we evaluate algorithm performance for different thresholding regimes. 
This is especially critical on standardized data. By re-scaling the data, standardization may impact the correct edge weights between nodes, potentially pushing them outside the thresholding range.
Following \cite{zheng2018dags,ng2020role}, we perform thresholding for the continuous structure learning algorithms and prune edges with an edge weight in the recovered adjacency matrix of less than $0.3$.
If the returned graph is not acyclic,
we iteratively remove the edge with the smallest magnitude weight
until all cycles are broken.
We find that the qualitative performance differences
between raw and standardized data
are robust to a wide range of threshold choices.

\cref{subfig:SID_thresholding_0001} and \cref{subfig:SID_thresholding_03} show SID performance for different thresholds. Even though the thresholds are orders of magnitude apart, a comparison reveals that the relative performances are nearly identical.

We observe that SHD performance is also robust across different thresholding regimes. \cref{subfig:SHD_thresholding_favorable} shows performance using \emph{favorable} thresholding. In this regime, the threshold leading to the most favorable SHD performance is applied to each instance individually. \cref{subfig:SHD_thresholding_03} shows performance for a fixed threshold of 0.3. A comparison reveals nearly identical relative performances in both cases.

Overall, we observe that the effect of varsortability is present even for the most favorable threshold in case of SHD, and for a wide range of thresholds in case of SID, where computation of a favorable threshold is computationally infeasible.

\begin{figure}[H]
    \centering
    \begin{subfigure}[b]{.49\linewidth}
        \centering
        \includegraphicsmaybe[width=\linewidth]
        {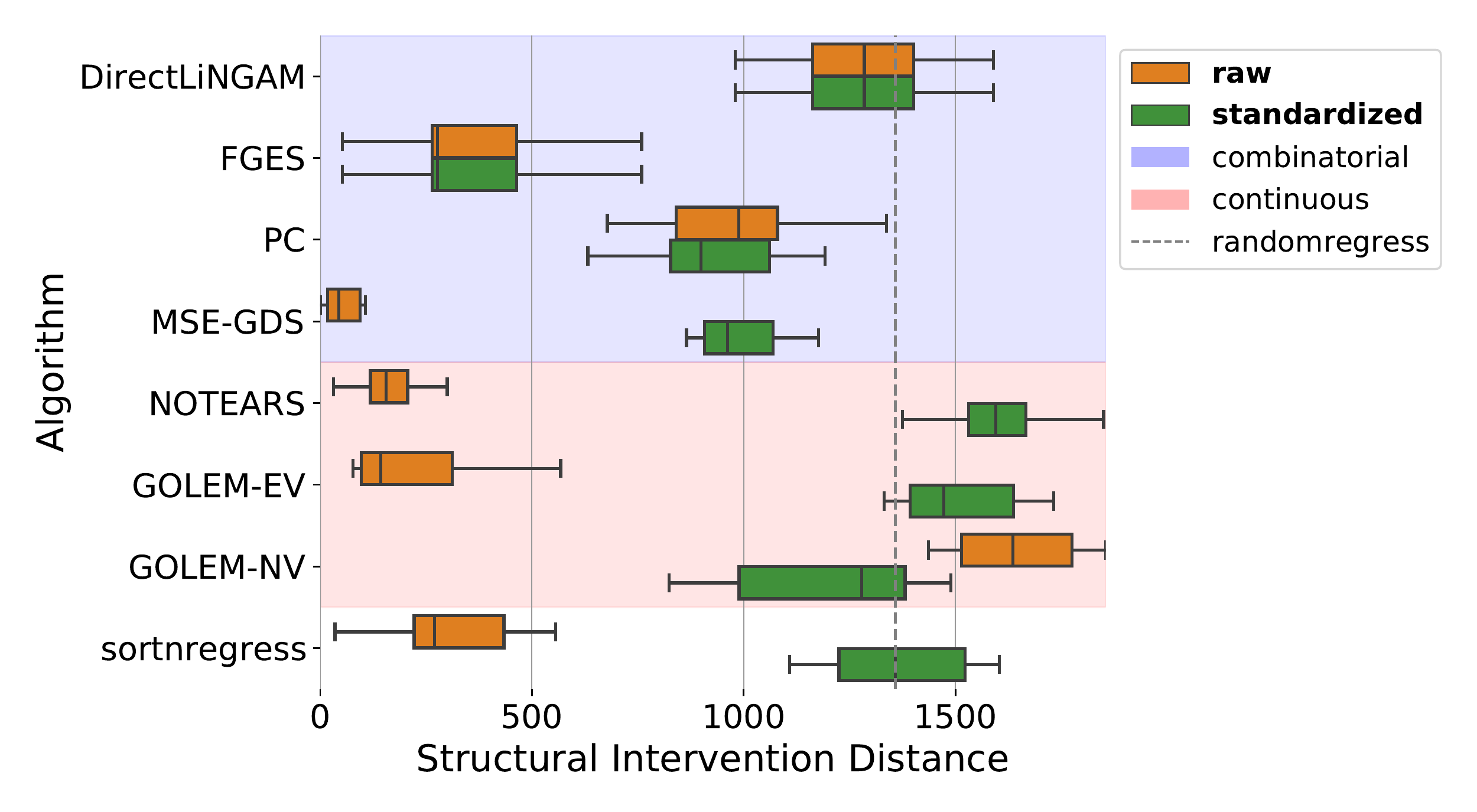}
        \caption{SID, threshold=0.001}
        \label{subfig:SID_thresholding_0001}
    \end{subfigure}
    \hfil
    \begin{subfigure}[b]{.49\linewidth}
        \centering
        \includegraphicsmaybe[width=\linewidth]
        {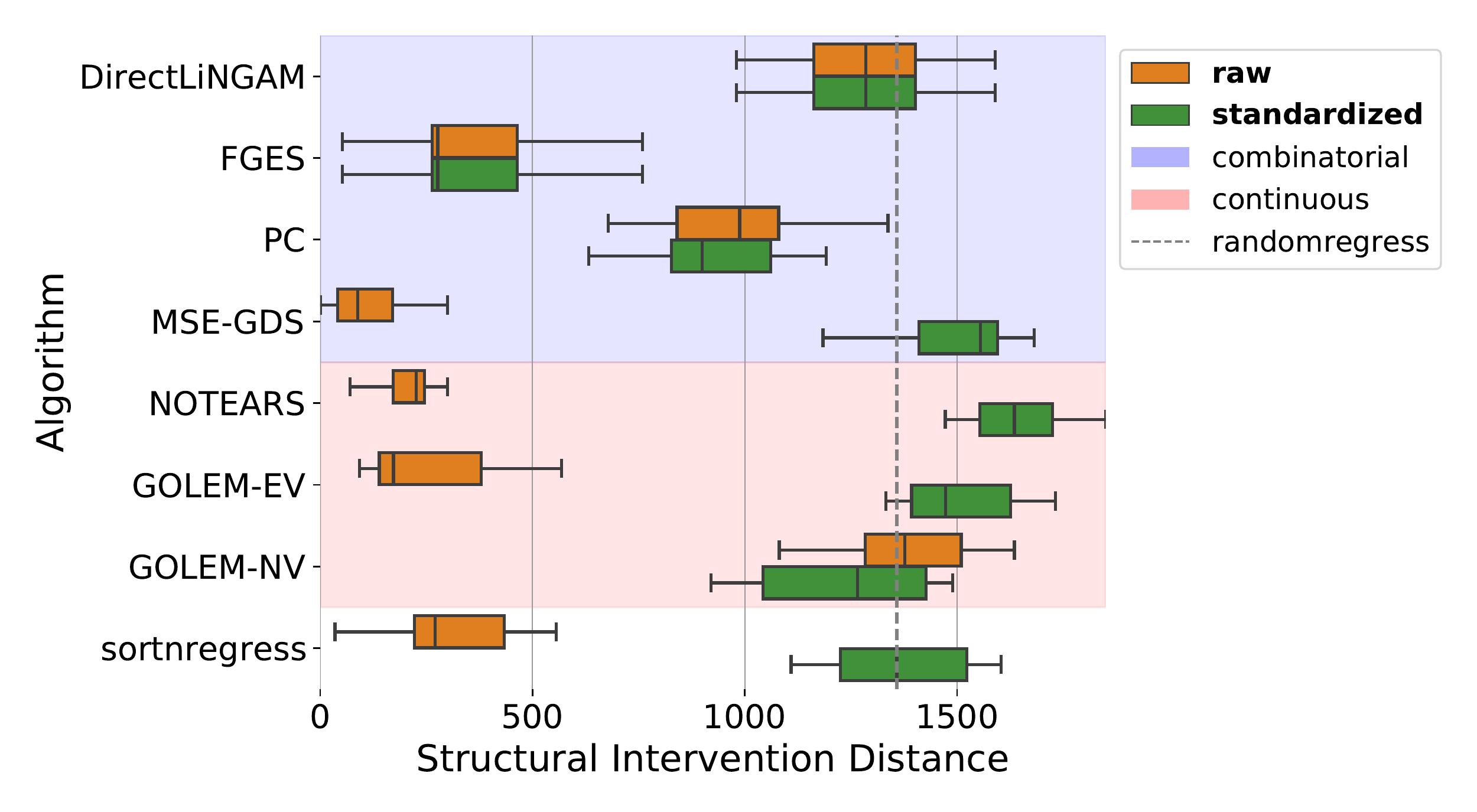}
        \caption{SID, threshold=0.3}
        \label{subfig:SID_thresholding_03}
    \end{subfigure}
    \hfil %
    \begin{subfigure}[b]{.49\linewidth}
        \centering
        \includegraphicsmaybe[width=\linewidth]
        {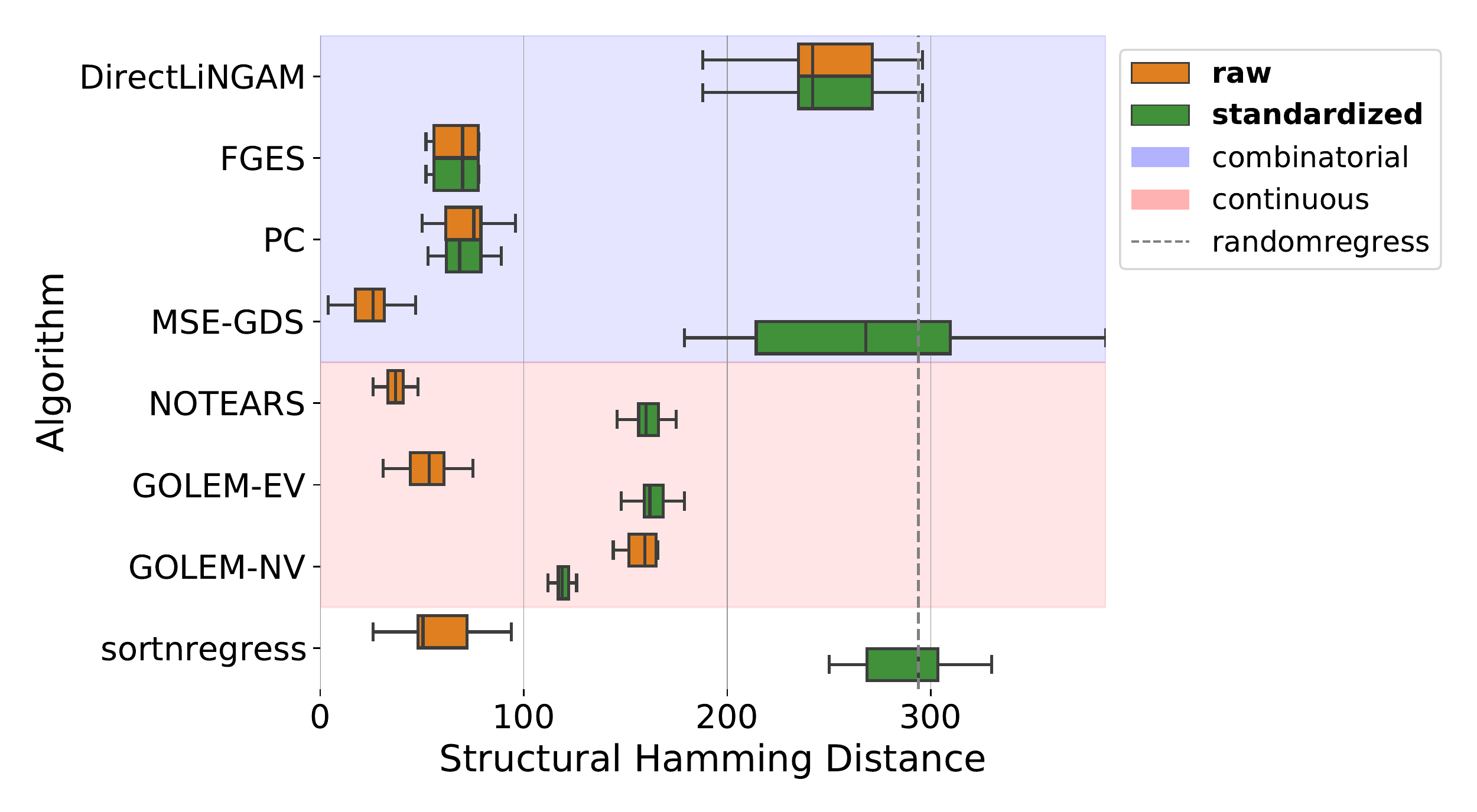}
        \caption{SHD, favorable thresholding}
        \label{subfig:SHD_thresholding_favorable}
    \end{subfigure}
    \hfil
    \begin{subfigure}[b]{.49\linewidth}
        \centering
        \includegraphicsmaybe[width=\linewidth]
        {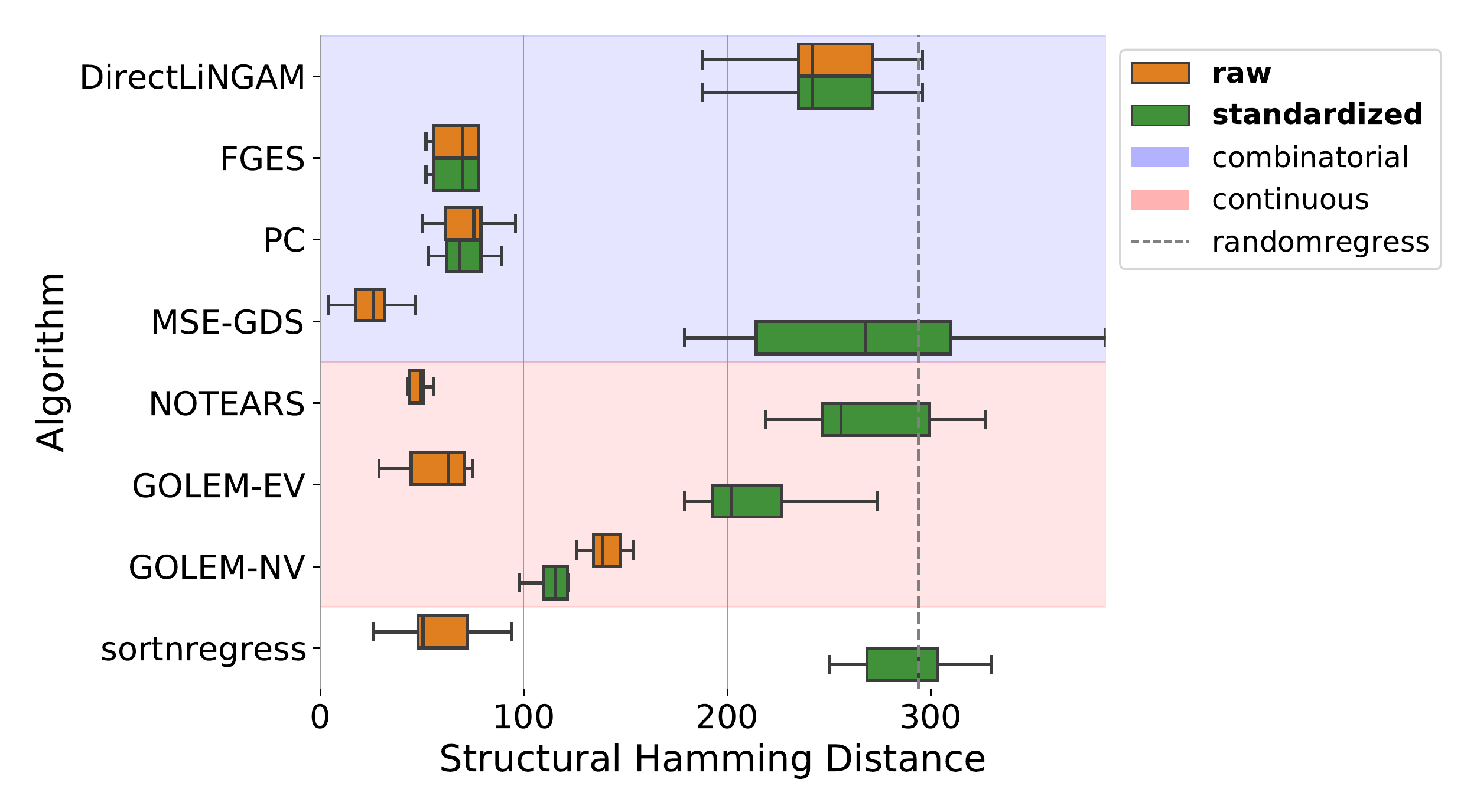}
        \caption{SHD, threshold=0.3}
        \label{subfig:SHD_thresholding_03}
    \end{subfigure}
    \hfil
    \caption{Results for different thresholding regimes. Gaussian-NV noise, ER-2 graph, 50 nodes.}
    \label{fig:SID_thresholding}
\end{figure}

\subsection{Results Across Noise Distributions and Graph Types}

\cref{fig:SID_all_settings} and \cref{fig:SHD_all_settings}
show algorithm comparisons in terms of SID and SHD, respectively.
The differences in performance on raw versus standardized data are
qualitatively similar regardless of the noise distribution.
We showcase results for different graph types in the non-Gaussian setting.
\textit{DirectLiNGAM}
performs well only in the non-Gaussian cases,
as is expected based on its underlying identifiability assumptions.

\begin{figure}[H]
    \centering
    \begin{subfigure}[b]{.49\linewidth}
        \centering
        \includegraphicsmaybe[scale=0.27]
        {figures/plots/supplementary/boxplots/boxplot_standard_0.3_sid_gauss_0.5,2_50_ER-2_selection.pdf}
        \caption{SID, Gaussian-NV noise, ER-2 graph, 50 nodes}
        \label{subfig:sid_gaussNV}
    \end{subfigure}
    \begin{subfigure}[b]{.49\linewidth}
        \centering
        \includegraphicsmaybe[scale=0.27]
        {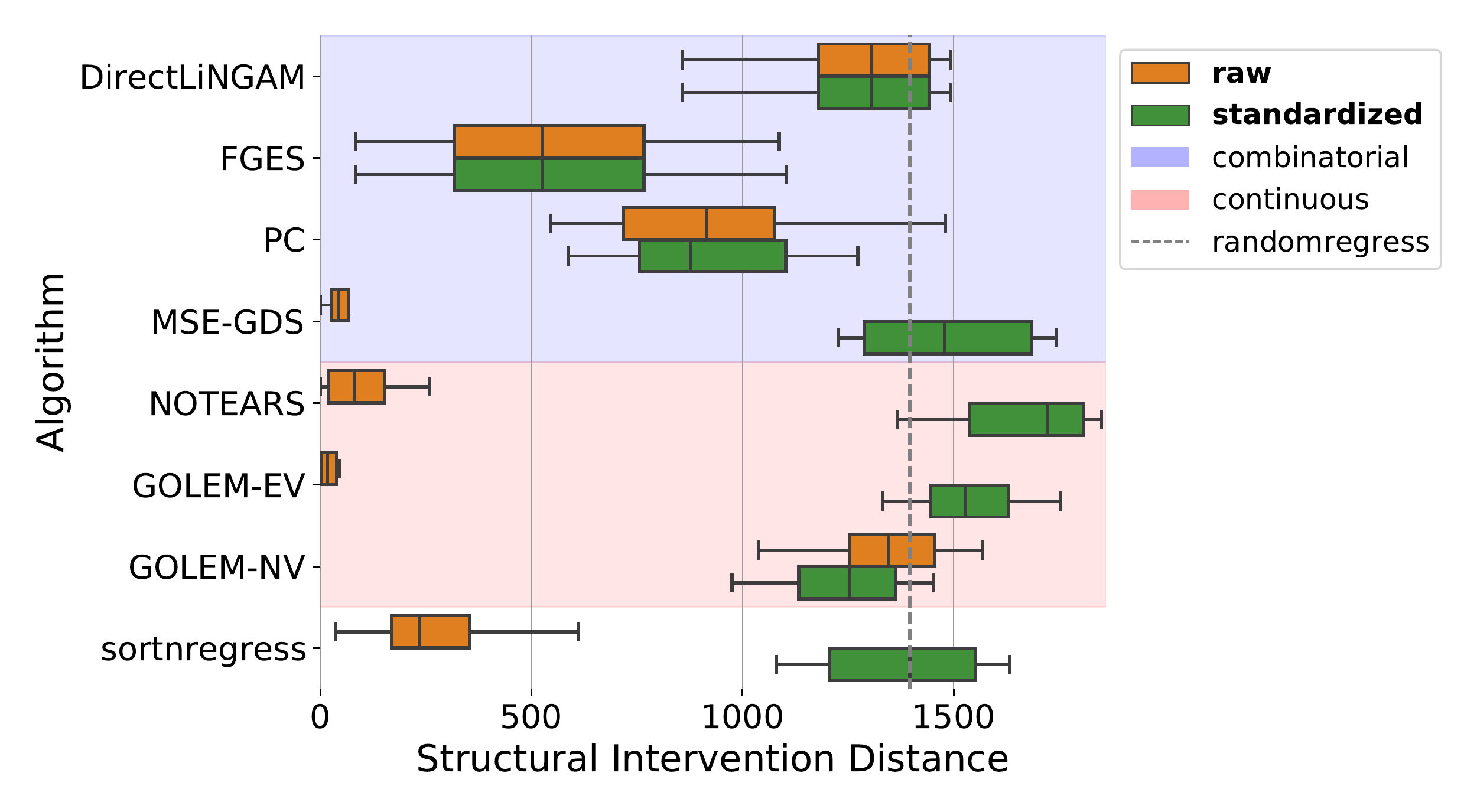}
        \caption{SID, Gaussian-EV noise, ER-2 graph, 50 nodes}
        \label{subfig:sid_gaussEV}
    \end{subfigure}
    \begin{subfigure}[b]{0.49\linewidth}
        \centering
        \includegraphicsmaybe[scale=0.27]
        {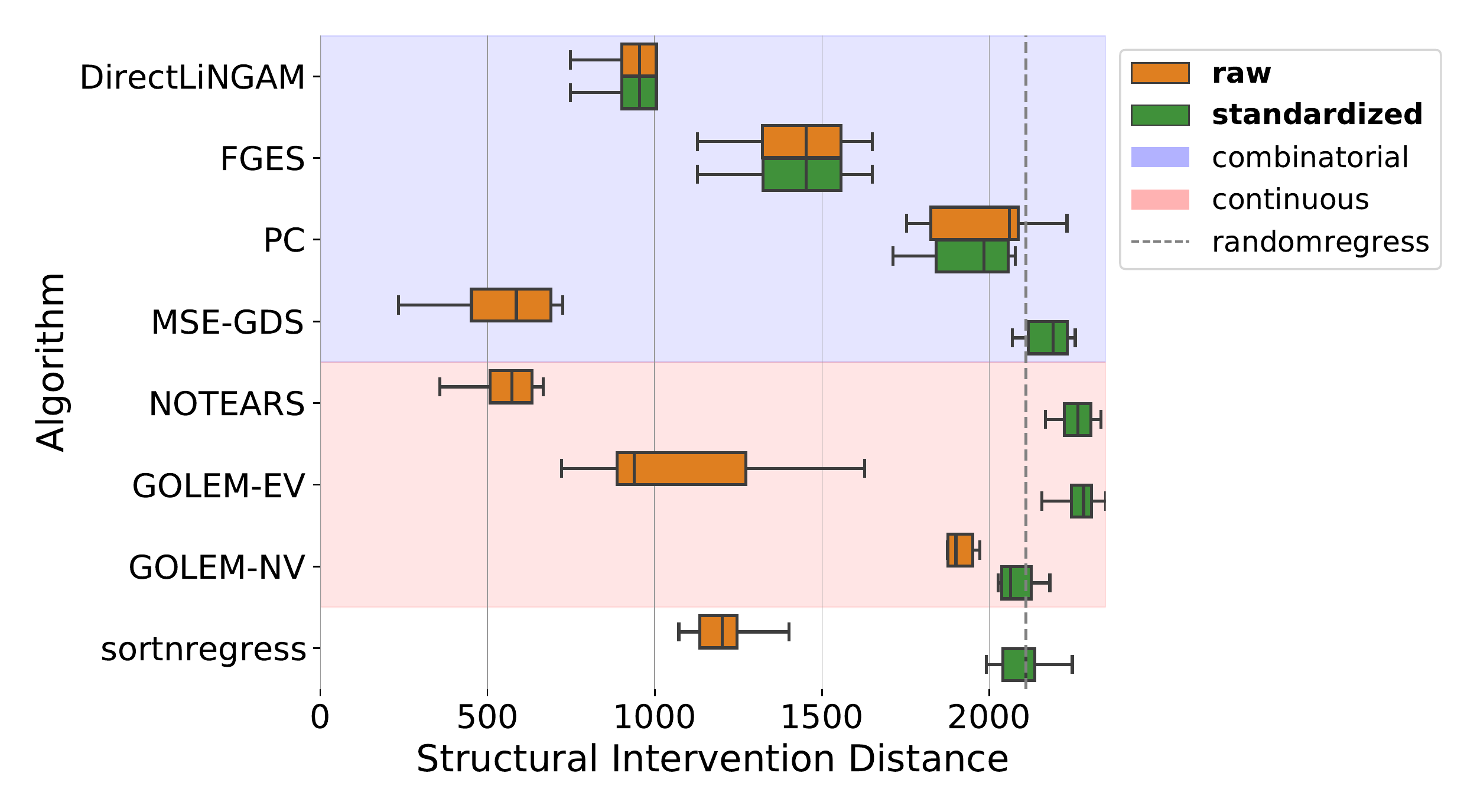}
        \caption{SID, Exponential noise, ER-4 graph, 50 nodes}
        \label{subfig:SID_exp}
    \end{subfigure}
    \hfill
    \begin{subfigure}[b]{0.49\linewidth}
        \centering
        \includegraphicsmaybe[scale=0.27]
        {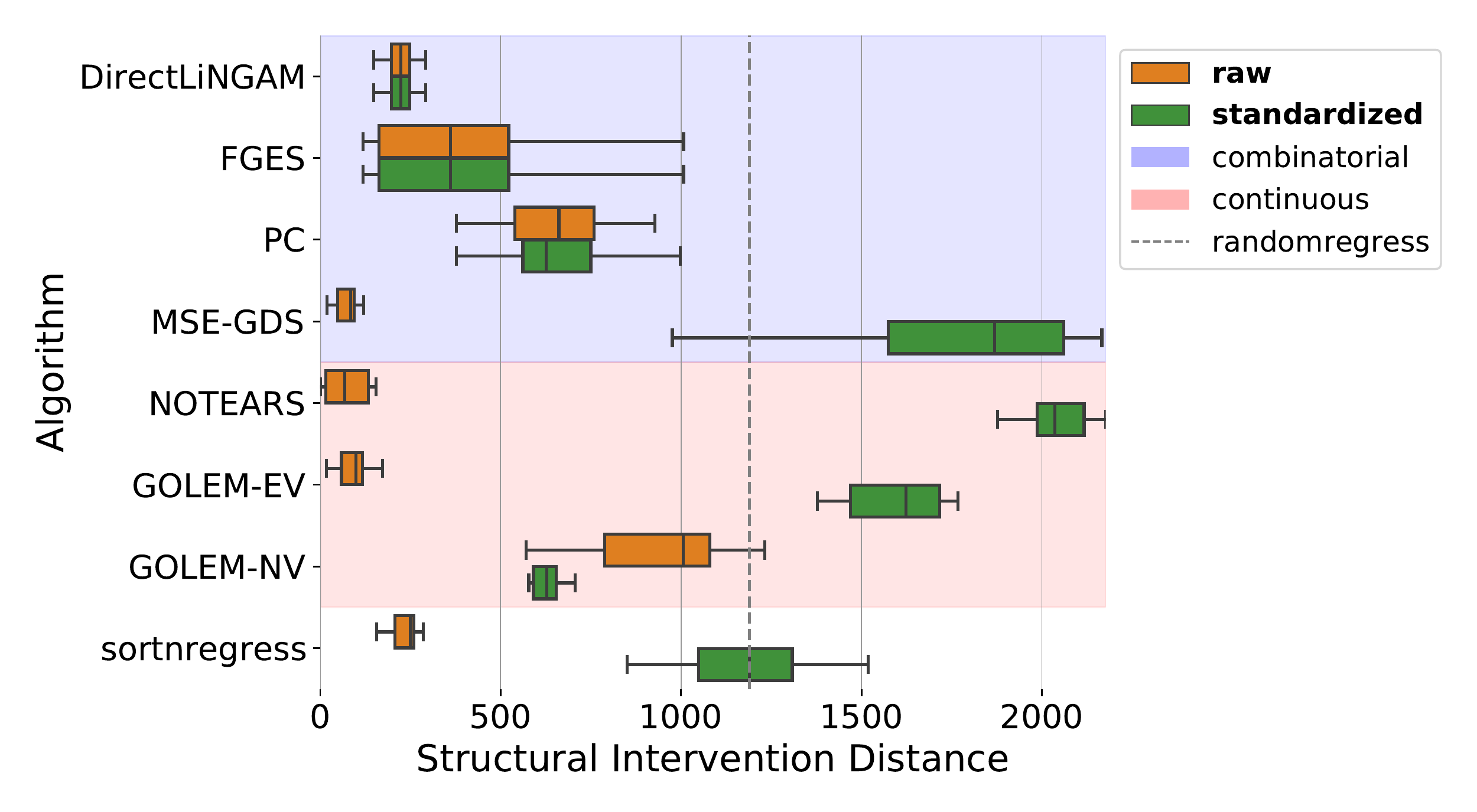}
        \caption{SID, Gumbel noise, SF-4 graph, 50 nodes}
        \label{subfig:SID_gum}
    \end{subfigure}
    \hfill
    \caption{SID results across noise types and for different graph types with 50 nodes}
    \label{fig:SID_all_settings}
\end{figure}

\begin{figure}[H]
    \centering
    \centering
    \begin{subfigure}[b]{.49\linewidth}
        \centering
        \includegraphicsmaybe[scale=0.27]
        {figures/plots/supplementary/boxplots/boxplot_standard_0.3_shd_gauss_0.5,2_50_ER-2_selection.pdf}
        \caption{SHD, Gaussian-NV noise, ER-2, 50 nodes}
        \label{fig:std_gauss_NV}
    \end{subfigure}
    \hfill
    \begin{subfigure}[b]{.49\linewidth}
        \centering
        \includegraphicsmaybe[scale=0.27]
        {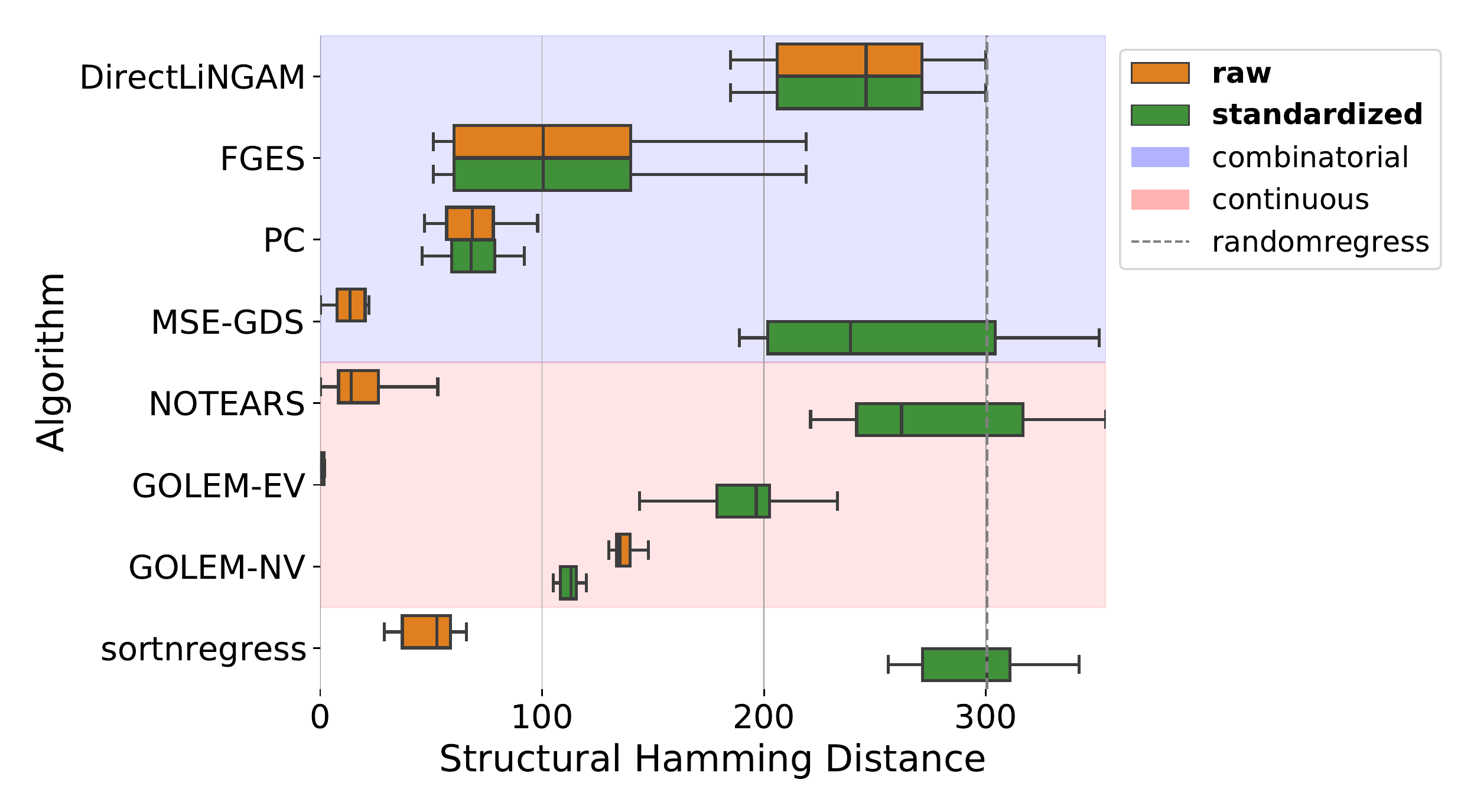}
        \caption{SHD, Gaussian-EV noise, ER-2, 50 nodes}
        \label{fig:std_gauss_EV}
    \end{subfigure}
    \hfill
    \begin{subfigure}[b]{.49\linewidth}
        \centering
        \includegraphicsmaybe[scale=0.27]
        {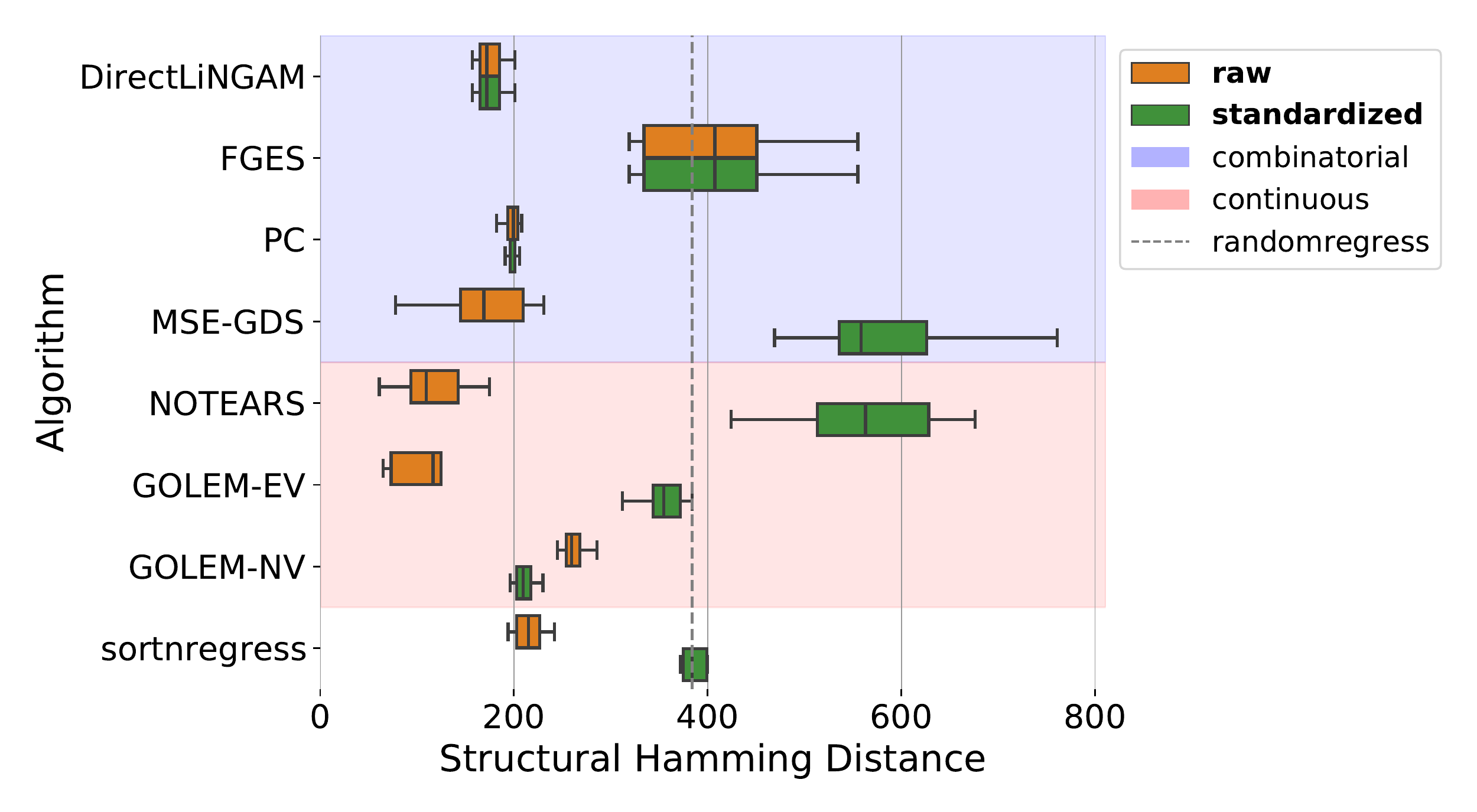}
        \caption{SHD, Exponential Noise, ER-4, 50 nodes}
    \end{subfigure}
    \hfill
    \begin{subfigure}[b]{.49\linewidth}
        \centering
        \includegraphicsmaybe[scale=0.27]
        {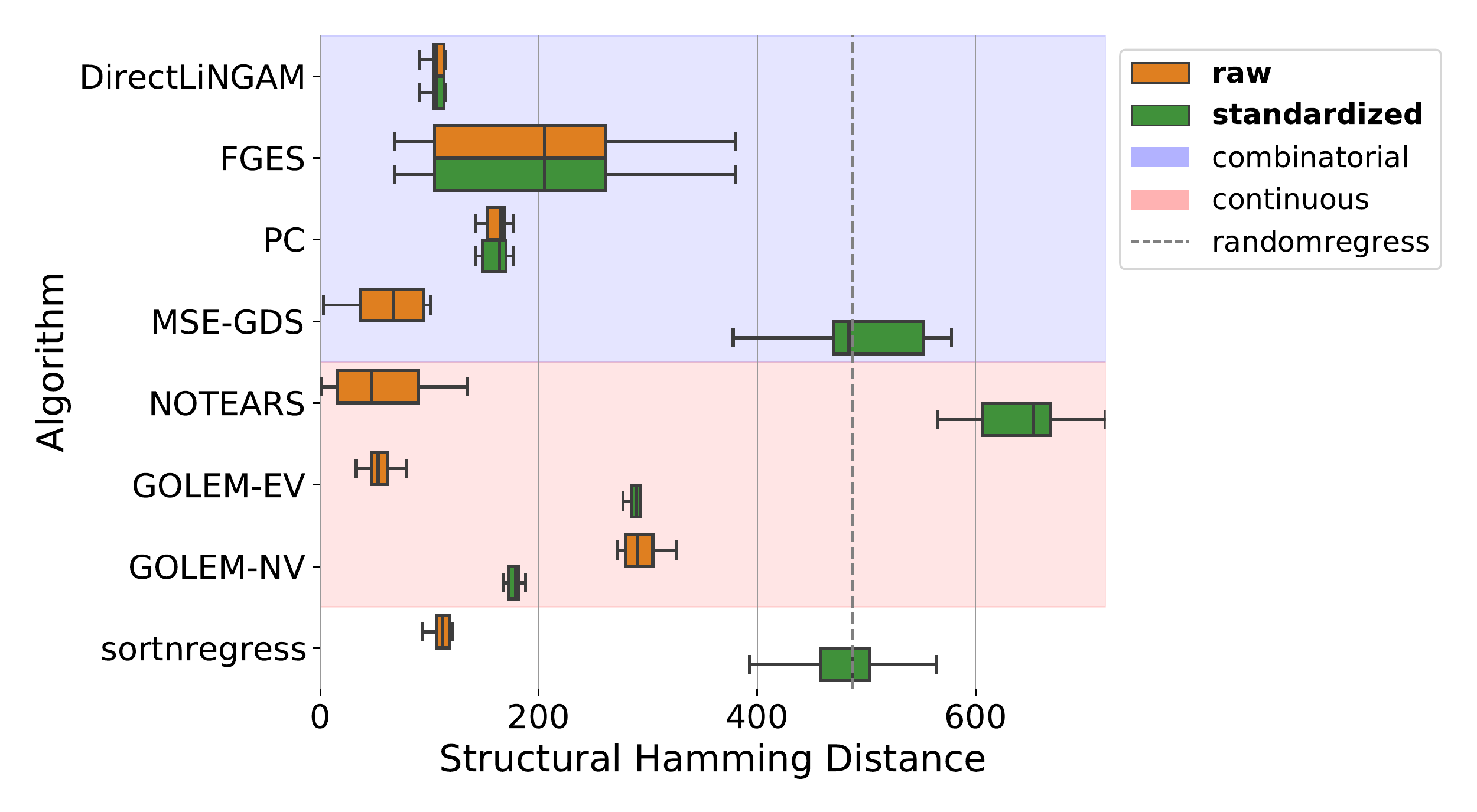}
        \caption{SHD, Gumbel Noise, SF-4, 50 nodes}
    \end{subfigure}
    \hfill
    \caption{SHD results across noise types and for different graph types with 50 nodes}
    \label{fig:SHD_all_settings}
\end{figure}

\clearpage
\subsection{Results Across Noise Distributions, Graph Types, and Graph Sizes}
The following experimental results largely follow earlier settings and results by \cite{zheng2018dags, ng2020role}.

\begin{figure}[H]
    \centering
    \begin{subfigure}[b]{.49\linewidth}
        \centering
        \includegraphicsmaybe[width=\linewidth]
        {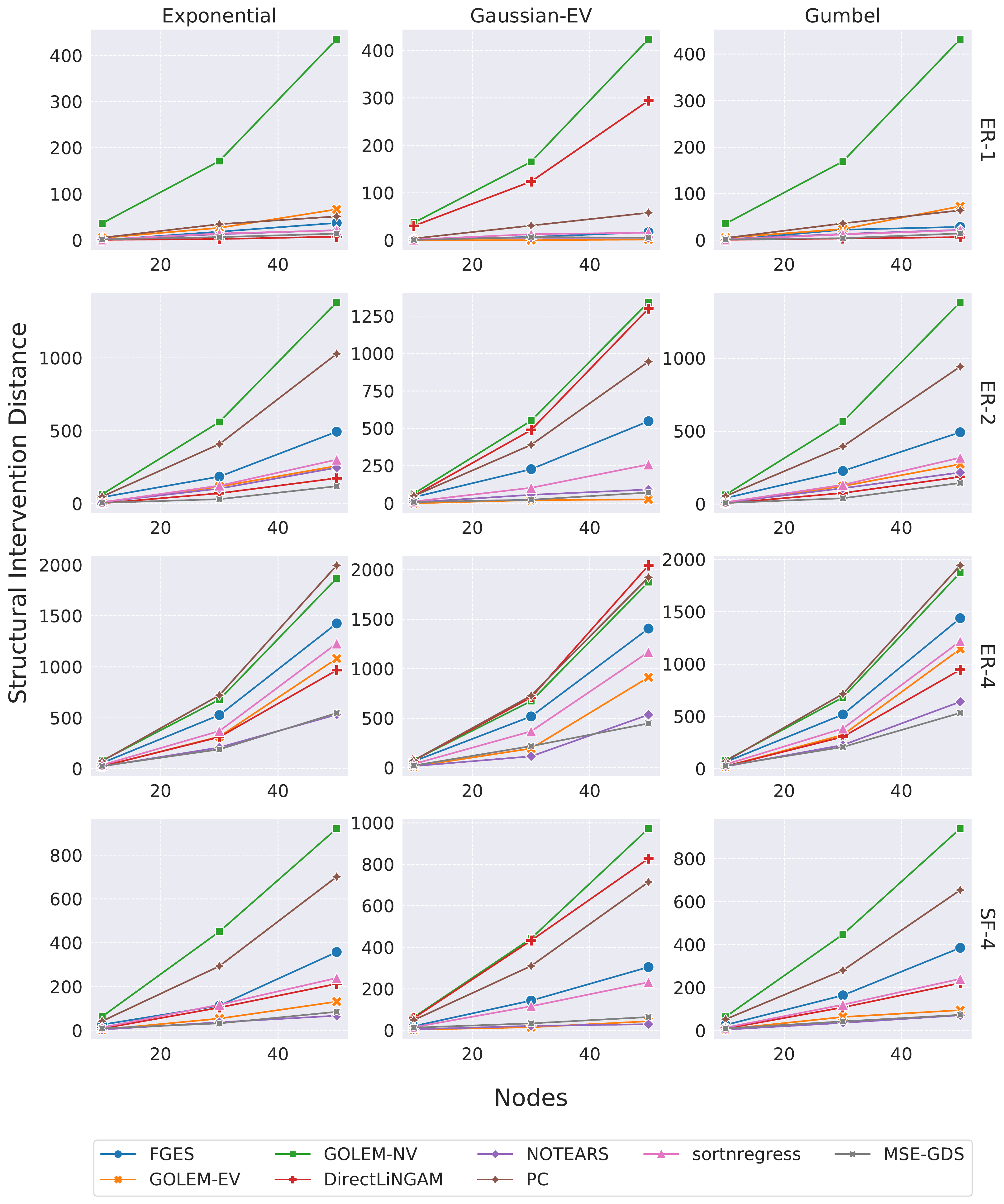}
        \caption{Raw data}
        \label{subfig:overview_sid_raw}
    \end{subfigure}
    \hfill
    \begin{subfigure}[b]{.49\linewidth}
        \centering
        \includegraphicsmaybe[width=\linewidth]
        {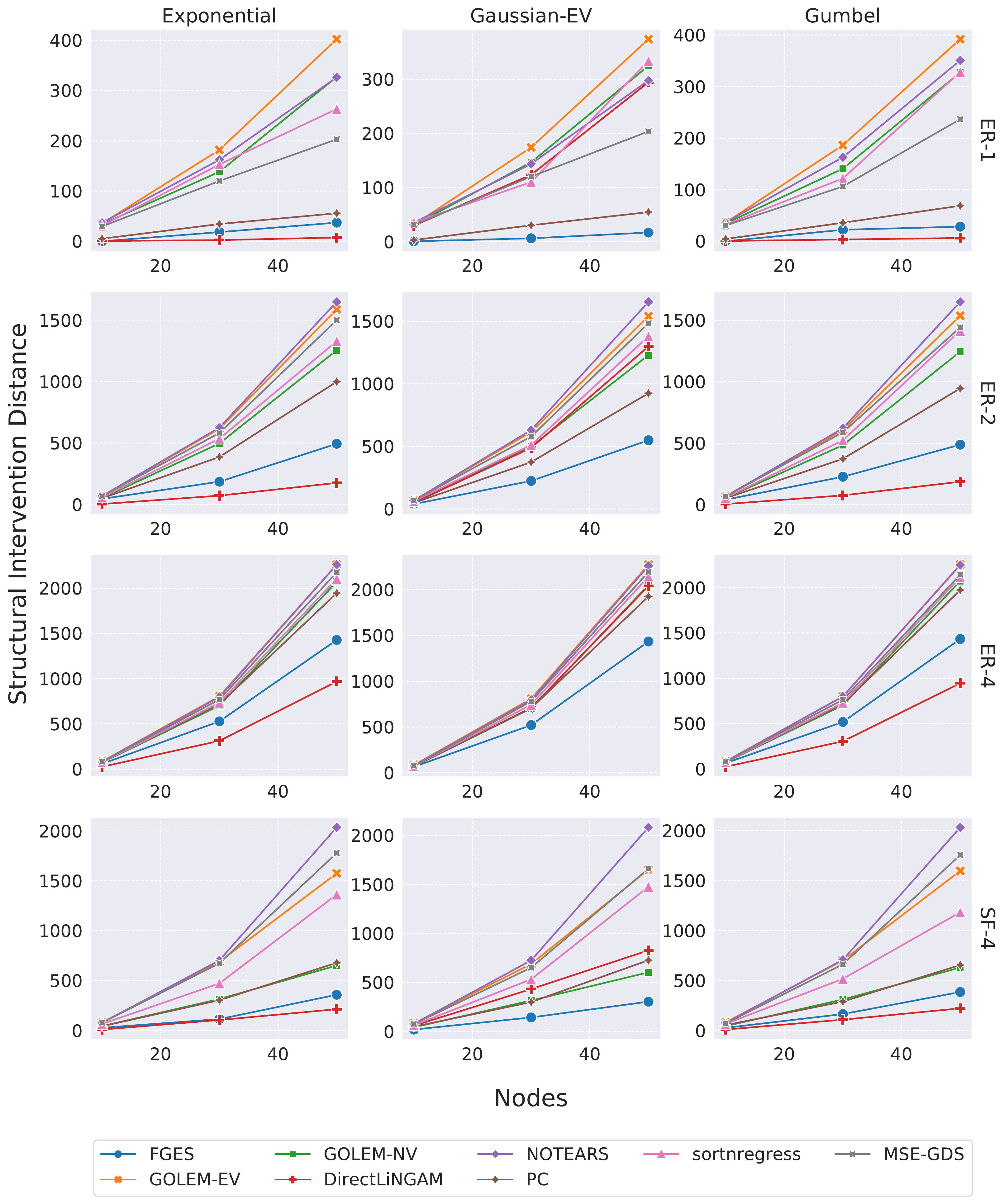}
        \caption{Standardized data}
        \label{subfig:overview_sid_normalized}
    \end{subfigure}
    \hfill
    \label{}
    \caption{SID results across noise types, graph types, and graph sizes.}
\end{figure}

\begin{figure}[H]
    \centering
    \begin{subfigure}[b]{.49\linewidth}
        \centering
        \includegraphicsmaybe[width=\linewidth]
        {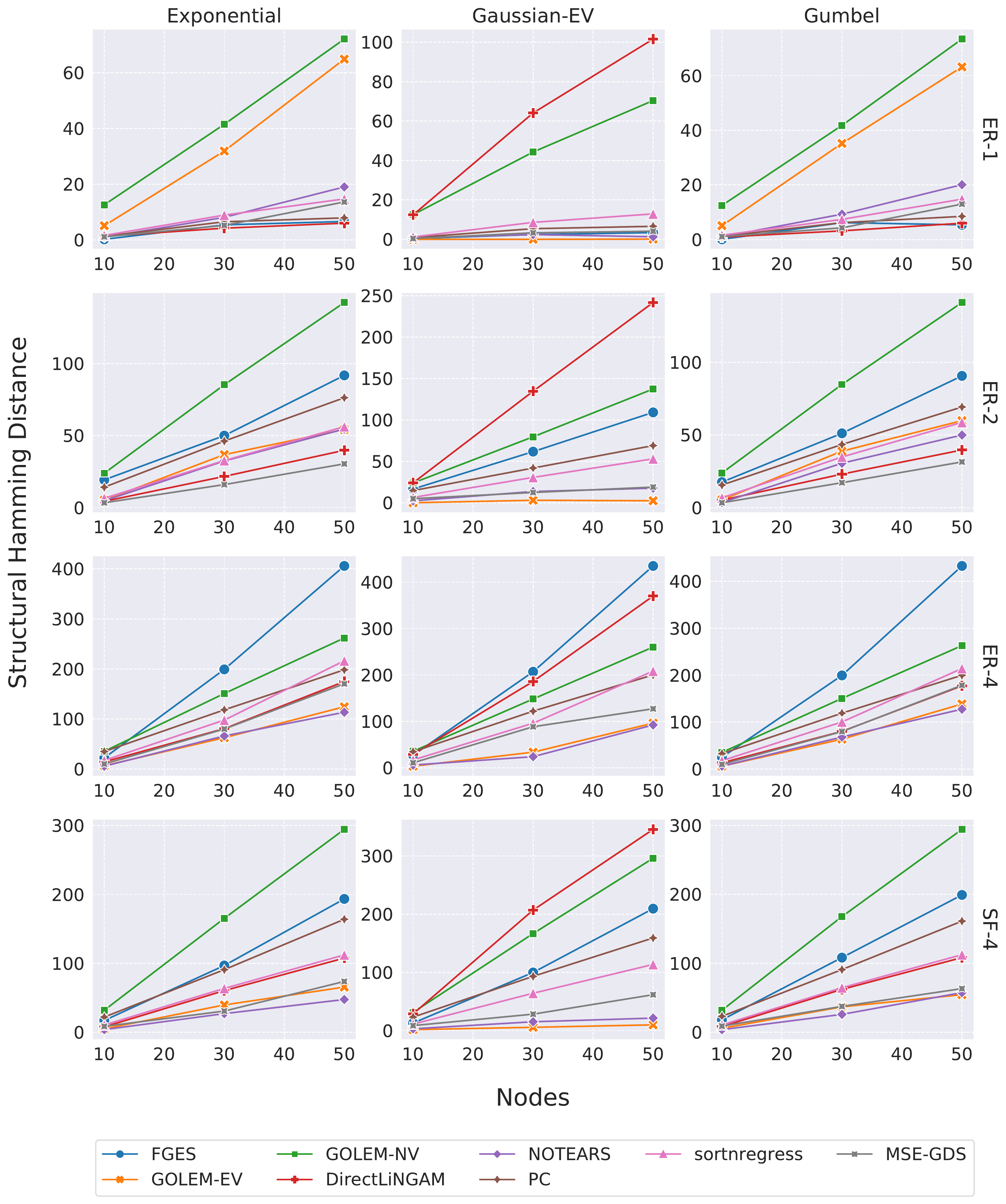}
        \caption{Raw data}
        \label{subfig:overview_shd_raw}
    \end{subfigure}
    \hfill
    \begin{subfigure}[b]{.49\linewidth}
        \centering
        \includegraphicsmaybe[width=\linewidth]
        {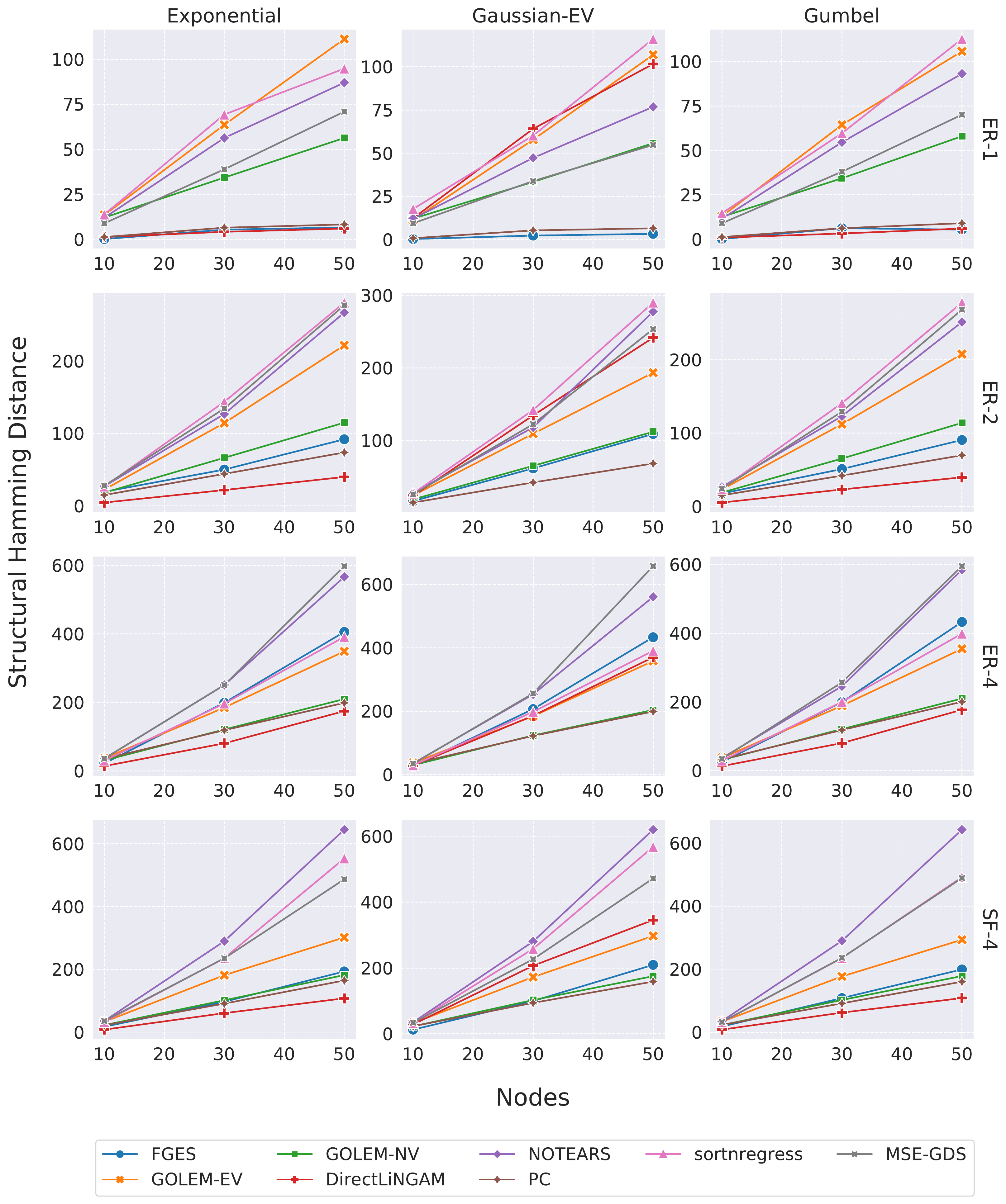}
        \caption{Standardized data}
        \label{subfig:overview_shd_normalized}
    \end{subfigure}
    \hfill
    \label{}
    \caption{SHD results across noise types, graph types, and graph sizes.}
\end{figure}

\end{document}